\documentclass{article}

\PassOptionsToPackage{sort&compress,numbers}{natbib}

    \usepackage[final]{neurips_2022}

\usepackage{amsmath,amsfonts,bm}

\def\eqref#1{equation~\ref{#1}}

\def\1{\bm{1}}

\def\vc{{\bm{c}}}

\DeclareMathAlphabet{\mathsfit}{\encodingdefault}{\sfdefault}{m}{sl}
\SetMathAlphabet{\mathsfit}{bold}{\encodingdefault}{\sfdefault}{bx}{n}

\newcommand{\E}{\mathbb{E}}

\newcommand{\Ltwo}{\mathcal{L}_{2}}
\newcommand{\R}{\mathbb{R}}

\newcommand{\KL}{D_{\mathrm{KL}}}

\DeclareMathOperator*{\argmax}{arg\,max}

\newcommand{\lo}{\alpha}
\newcommand{\hi}{\omega}
\newcommand{\zhi}{z_{\hi}}
\newcommand{\zlo}{z_{\lo}}
\usepackage[utf8]{inputenc}
\usepackage{graphicx}
\usepackage{nicefrac}
\usepackage{xcolor}
\usepackage{nameref}
\usepackage{array}
\usepackage{lineno}
\usepackage[T1]{fontenc}    
\usepackage{url}            
\usepackage{amsfonts}       
\usepackage{amsmath}
\usepackage{microtype}      
\usepackage{comment}
\usepackage{accents}
\usepackage{xspace}
\usepackage[subrefformat=parens,labelformat=parens]{subcaption}
\usepackage{booktabs}
\usepackage{bibunits}
\usepackage{siunitx}
\usepackage{appendix}
\usepackage{pifont}
\usepackage{hyperref}       
\usepackage{xspace}
\usepackage{multirow}
\usepackage{wrapfig}
\usepackage{calc}
\usepackage{lipsum}
\usepackage{selectp}

\newif\ifphantomgraphics
\phantomgraphicstrue

\newcommand{\phantomgraphics}[2][]{%
  \ifphantomgraphics
  \leavevmode\phantom{\includegraphics[#1]{#2}}%
  \else
  \includegraphics[#1]{#2}%
  \fi
}

\usepackage{listings}
\usepackage{textcomp}

\definecolor{codegreen}{rgb}{0,0.6,0}
\definecolor{codegray}{rgb}{0.5,0.5,0.5}
\definecolor{codepurple}{rgb}{0.58,0,0.82}
\definecolor{backcolour}{rgb}{0.95,0.95,0.92}

\lstdefinestyle{mystyle}{
    backgroundcolor=\color{backcolour},   
    commentstyle=\color{codegreen},
    keywordstyle=\color{magenta},
    numberstyle=\tiny\color{codegray}\ttfamily,
    stringstyle=\color{codepurple},
    basicstyle=\fontsize{7.5pt}{9pt}\ttfamily,
    breakatwhitespace=false,         
    breaklines=true,                 
    captionpos=b,                    
    keepspaces=true,                 
    numbers=left,                    
    numbersep=5pt,                  
    showspaces=false,                
    showstringspaces=false,
    showtabs=false,                  
    tabsize=2,
    upquote=true
}
\usepackage[capitalise]{cleveref}
\crefname{equation}{}{}

\crefname{lstlisting}{listing}{listings}
\Crefname{lstlisting}{Listing}{Listings}

\crefname{supp}{Supplementary Material}{Supplementary Materials}

\newcommand{\rulesep}{\unskip\ \vrule\ }

\newcommand{\bi}{\begin{itemize}}
\newcommand{\ei}{\end{itemize}}
\newcommand{\be}{\begin{enumerate}}
\newcommand{\ee}{\end{enumerate}}
\newcommand{\bb}{\begin{block}}
\newcommand{\eb}{\end{block}}

\newcommand{\OurAlg}{Multiagent Offline Hierarchical Behavior Analyzer\xspace}
\newcommand{\OurAlgAcronym}{MOHBA\xspace}

\title{Beyond Rewards: a Hierarchical Perspective on\\Offline Multiagent Behavioral Analysis}

\author{%
  Shayegan~Omidshafiei\\
  \texttt{somidshafiei@google.com} \\
  \And
  Andrei~Kapishnikov\\
  \texttt{kapishnikov@google.com} \\
  \And
  Yannick~Assogba\\
  \texttt{yassogba@google.com} \\
  \And
  Lucas~Dixon\\
  \texttt{ldixon@google.com} \\
  \And
  Been~Kim\\
  \texttt{beenkim@google.com} \\
  \AND
  {\normalfont Google Research}
}

\begin{document}

\maketitle

\begin{abstract}
    Each year, expert-level performance is attained in increasingly-complex multiagent domains, where notable examples include Go, Poker, and StarCraft~II. This rapid progression is accompanied by a commensurate need to better understand how such agents attain this performance, to enable their safe deployment, identify limitations, and reveal potential means of improving them. In this paper we take a step back from performance-focused multiagent learning, and instead turn our attention towards agent behavior analysis. We introduce a model-agnostic method for discovery of behavior clusters in multiagent domains, using variational inference to learn a hierarchy of behaviors at the joint and local agent levels. Our framework makes no assumption about agents' underlying learning algorithms, does not require access to their latent states or policies, and is trained using only offline observational data. We illustrate the effectiveness of our method for enabling the coupled understanding of behaviors at the joint and local agent level, detection of behavior changepoints throughout training, discovery of core behavioral concepts, demonstrate the approach's scalability to a high-dimensional multiagent MuJoCo control domain, and also illustrate that the approach can disentangle previously-trained policies in OpenAI's hide-and-seek domain.
\end{abstract}

\section{Introduction}

Multiagent approaches have driven numerous advances in artificial intelligence research, with seminal examples including TD-gammon~\citep{tesauro1995temporal}, DeepBlue~\citep{campbell2002deep}, AlphaGo~\citep{silver2016mastering}, 
AlphaZero~\citep{silver2018general}, 
Libratus~\citep{brown2018superhuman}, AlphaStar~\citep{vinyals2019grandmaster}, OpenAI Five~\citep{berner2019dota}, and Pluribus~\citep{brown2019superhuman}.
During training, many of these approaches seek to push the performance of agents as measured by a reward signal, or derivatives thereof.

Despite this, post-hoc methods that seek to \emph{understand} agent interactions often use less reward-centric techniques.
Instead, insights are drawn from behavioral analysis to identify unique or interesting agent strategies.
Examples include clustering-based analysis of neuron activations and trajectories in capture-the-flag~\citep{jaderberg2019human}, inspection of trajectories in a hide-and-seek domain to detect interesting behaviors such as agents that learn to exploit the underlying physics engine~\citep{Baker2020Emergent}, monitoring of statistics such as pass ranges and frequencies in humanoid football~\citep{liu2021motor}, and analysis of AlphaZero's acquisition of chess knowledge~\citep{AZ2021}. 
Crucially, such insights are often drawn via manual analysis and detection of behavioral clusters, or use of statistics associated with certain behaviors as defined by humans experts (e.g., various skills and relevant metrics in humanoid football).

\begin{figure}[t]
    \newcommand{\myFigHeight}{5.0cm}
    \captionsetup[subfigure]{aboveskip=-2pt,belowskip=-2pt}
    \centering
    \begin{minipage}{0.45\textwidth} 
        \centering
        \begin{subfigure}[t]{\textwidth}
            \vspace{-10pt}
            \centering
            \includegraphics[width=\textwidth]{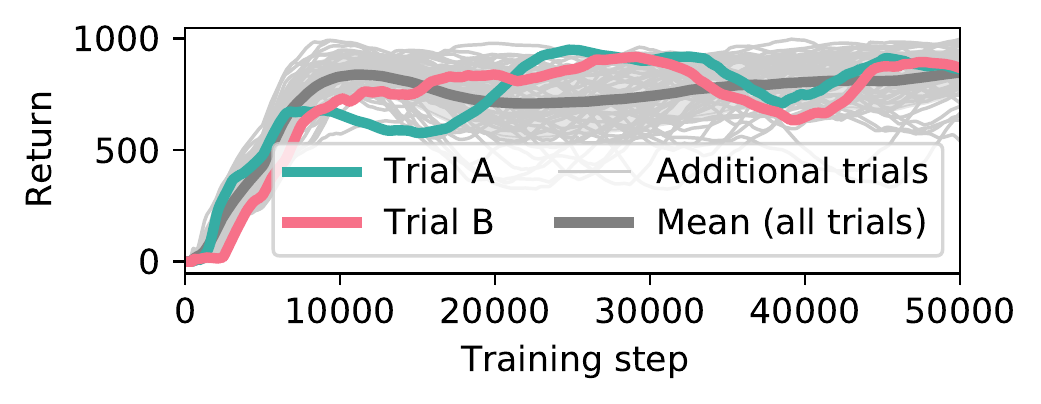}
            \caption{}
            \label{fig:multimodal_reward_env_returns}
        \end{subfigure}\\[-5pt]
        \begin{subfigure}[t]{\textwidth}
            \centering
            \includegraphics[height=\myFigHeight/2]{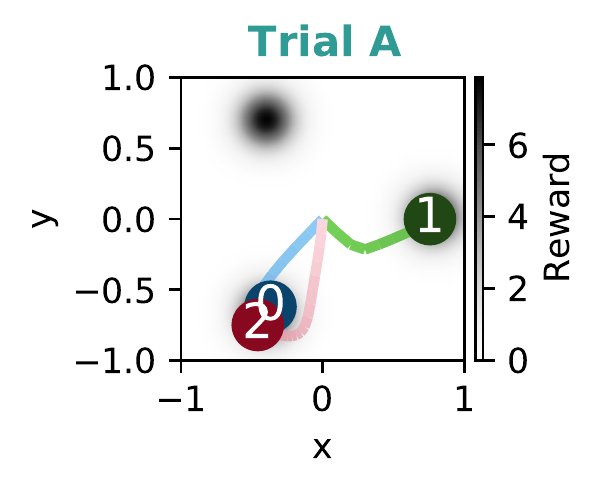}%
            \includegraphics[height=\myFigHeight/2]{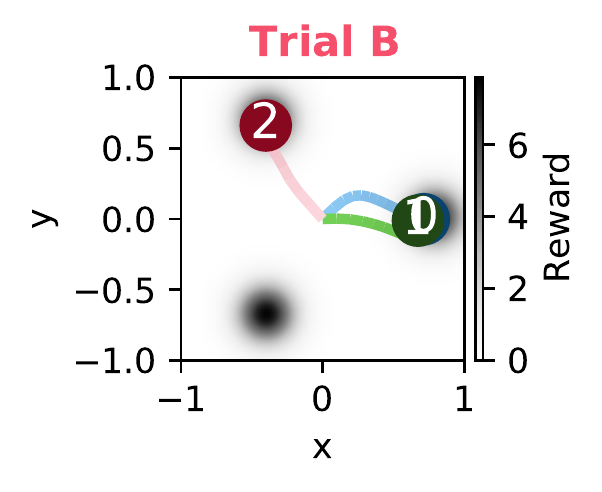}
            \vspace{-10pt}
            \caption{}
            \label{fig:multimodal_reward_env_trials}
        \end{subfigure}%
    \end{minipage}%
    \begin{minipage}{0.5\textwidth} 
        \begin{subfigure}[t]{\textwidth}
            \centering
            \vspace{-20pt}
            \includegraphics[height=\myFigHeight]{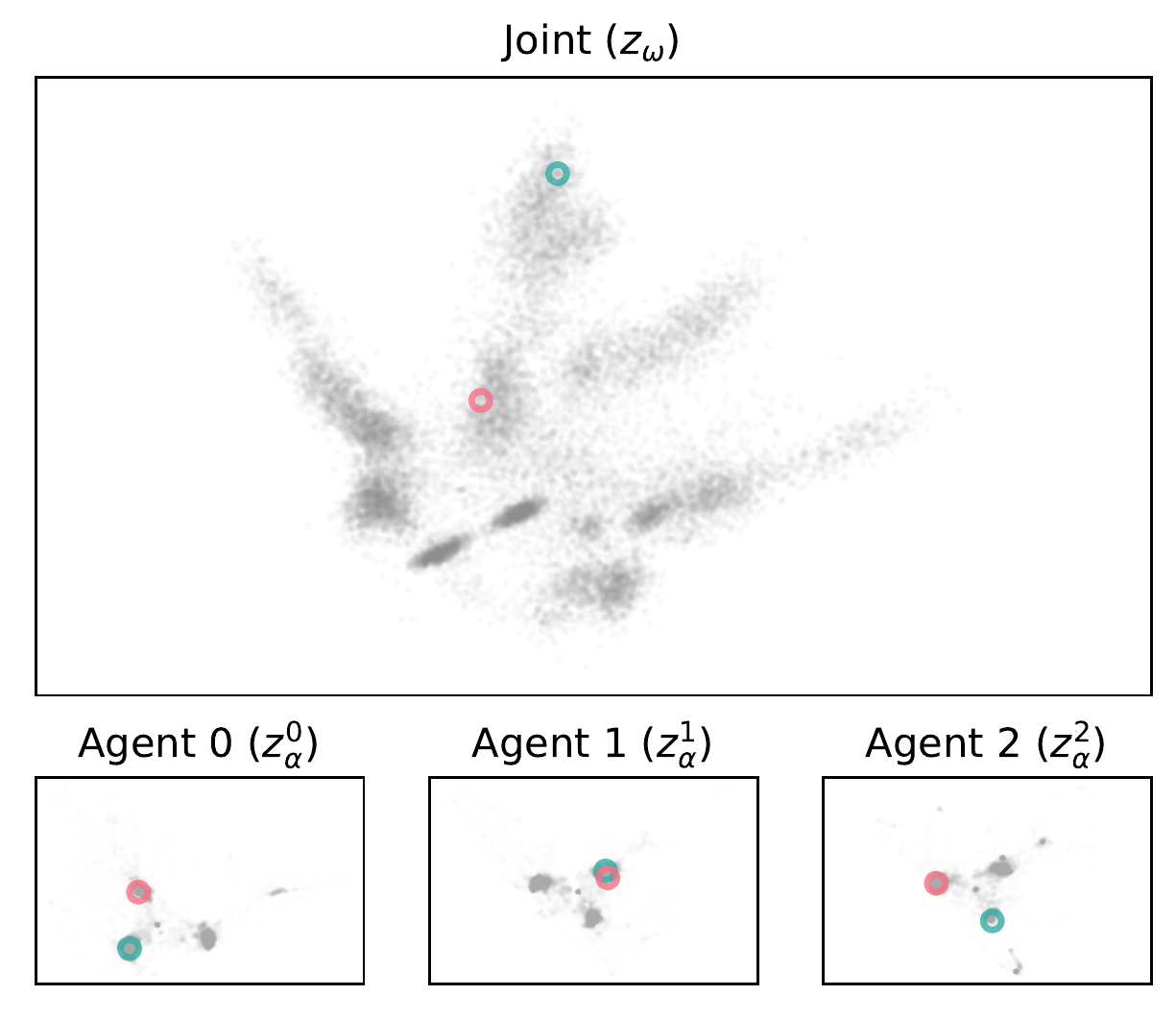}
            \caption{}
            \label{fig:multimodal_reward_env_latent_trials}
        \end{subfigure}%
    \end{minipage}
    \caption{
        Reward alone is not enough to understand underlying behaviors in a 3-agent hill-climbing domain.
        Agents here start at the origin, each receiving rewards by navigating to any of 3 equidistant hills. 
        \subref{fig:multimodal_reward_env_returns} visualizes the total returns of agents throughout training, over 50 independent trials.
        Two trials (A and B) with similar final returns are highlighted.
        \subref{fig:multimodal_reward_env_trials} visualizes the actual converged behaviors of the agents at the end of Trials A and B, which are distinct despite their similar returns. 
        Visualizing these same trajectories in the behavior space learned by our approach immediately reveals differences in the joint behavior of agents in the top panel of \subref{fig:multimodal_reward_env_latent_trials}, where the two color markers correspond to the trajectories from each trial.
        Simultaneously analyzing the agent-wise latent spaces in the bottom 3 panels of \subref{fig:multimodal_reward_env_latent_trials} highlight that agent 1 behaves the same way in both trials, in contrast to agents 0 and 2.
    }
    \vspace{-13pt}
    \label{fig:multimodal_reward_env_intro}
\end{figure}

As evident above, understanding emergent multiagent behaviors is enriched by techniques beyond pure reward-based analysis, as behavioral signifiers are not always discernible via rewards. 
\Cref{fig:multimodal_reward_env_returns} provides intuition on this notion, illustrating returns (sum of rewards) throughout training for a multiagent hill-climbing domain (later described in detail).
We highlight two independent training trials (A and B) with similar final returns.
Despite similar returns, comparing the trajectories generated via the agents' deterministic policies following each trial's training~(\cref{fig:multimodal_reward_env_trials}) reveals entirely different behaviors.
Analogous examples are evident in the above works (e.g., Fig.~1 of \citet{Baker2020Emergent}, where substantial behavior changes occur in multiagent hide-and-seek despite a smooth reward curve).

This paper formalizes the problem of offline multiagent behavior analysis.
Our proposed algorithm, \OurAlg (\OurAlgAcronym), learns a hierarchical latent space that simultaneously reveals behavior clusters at the joint level (i.e., interactions \emph{between} agents) and local level (i.e., behaviors of \emph{individual} agents).
Our method is agnostic of the underlying algorithm used to generate agents' behaviors, requires no access or control of the underlying environment, does not assume availability of a reward signal, and does not require access to agents' models or internal states.
Our experiments investigate the structure of the learned behavior space, which goes beyond prior works on latent-clustering by identifying relationships between individual agent and joint behaviors.
We illustrate that clusters identified by \OurAlgAcronym are useful for highlighting similarities and differences in behaviors throughout training.
We also quantitatively analyze the completeness of discovered behavior clusters by adopting a modified version of the concept-discovery framework of~\citet{yeh2020completeness} to identify interesting behavior concepts in our multiagent setting.
We then test the scalability of our approach by using it for behavioral analysis of several high-dimensional multiagent MuJoCo environments~\citep{de2020deep}.
Finally, we evaluate the approach on the open-sourced OpenAI hide-and-seek policy checkpoints~\citep{Baker2020Emergent}, confirming that the behavioral clusters detected by \OurAlgAcronym closely match those of the human-expert annotated labels provided in their policy checkpoints.

\section{Related Work}

Significant research has been conducted in single-agent skill discovery, which seeks to learn reusable policies useful for downstream tasks~\citep{csimcsek2008skill,eysenbach2018diversity,hausman2018learning,kipf2019compile,sharma2019dynamics,hasenclever2020comic,lynch2020learning,singh2020parrot,shankar2020learning,campos2020explore,pertsch2021accelerating,baumli2021relative,villecroze2022bayesian,zhu2022bottom}.
Related approaches discover motor primitives to express longer-horizon policies~\citep{shankar2019discovering}, including use of offline reinforcement learning (RL) for learning useful behaviors~\citep{ajay2020opal}.
Option discovery methods learn temporally-abstracted actions (i.e., options~\citep{sutton1999between}), chained together to form cohesive skills~\citep{srinivas2016option,henderson2018optiongan,achiam2018variational,barreto2019option,manoharan2020option,kim2021unsupervised}.
In contrast to our work, these approaches focus on maximizing performance in single-agent settings.
There also exists a related line of work for learning diverse policies in RL settings~\citep{lupu2021trajectory,parker2020effective,zhou2021continuously,pugh2016quality,tang2021discovering,song2019diversity,hong2018diversity,gangwani2018learning,cohen2019diverse};
despite their focus on policy diversity, several of these approaches make much stronger assumptions than ours (e.g., access to the policies generating agent trajectories, or unique identifiers of the policy that generated each trajectory). 
By contrast, our work assumes no access to any of the underlying raw policies, or even labels of which policies agent trajectories were obtained from.
Recent works have also focused on hierarchical skill learning in multiagent reinforcement learning (MARL).
\citet{lee2019learning} use the mutual information maximization objective introduced in~\citet{eysenbach2018diversity} to learn multiagent policies.
\citet{yang2020hierarchical} use a bi-level policy to learn agent skills: a high-level policy first generates latent vectors for each agent, and a low-level policy conditions behaviors on said vector to perform the task.
In \citet{wang2020rode}, distinct `roles' are learned for agents to enable decomposition of tasks.
\citet{mao2020neighborhood} investigate use of consistent latent cognition variables in agent neighborhoods to induce increased cooperation.
Many MARL approaches have noted the emergence of interesting behaviors in multiagent systems in specific domains of interest~\citep{johanson2022,cao2018emergent,noukhovitch2021emergent,liu2018emergent,bansal2018emergent}.
In contrast to our focus, these approaches use RL to maximize agent performance, rather than understand arbitrary behaviors via offline analysis.

RL interpretability methods primarily focus on single-agent settings and either modify the RL algorithm itself to increase transparency, or conduct post-hoc explainability~\citep{heuillet2021explainability}.
These approaches represent agent policies as programming languages~\citep{verma2018programmatically}, extract visual summaries of behaviors using `interestingness' statistics such as uncertainty in selected actions or the value of state-transitions~\citep{sequeira2020interestingness}, or combine agent neuron activations with gradient information to construct behavioral embeddings~\citep{zahavy2016graying}.
Behavioral clusters in our work share similarities with concept-based explanation approaches in non-RL domains.
Detecting `concepts' in pre-trained models has been explored in vision~\citep{ghorbani2019towards,bau2020units,dissect2021}, discrete games~\citep{AZ2021}, and language~\citep{Rogers2020,hewitt-manning-2019-structural}. 
In vision, clustering-based approaches describe discovered concepts using examples~\citep{ghorbani2019towards} or use generative modeling to create new data to describe concepts~\citep{dissect2021}. 
\citet{ghorbani2019towards} uses a vision-specific method (i.e., superpixels) to sub-divide input before conducing clustering;
while sub-division is less natural in RL, our LSTM and VAE baselines serve as an RL-adopted counterpart of such works.

Works using latent-clustering and analysis are also related to ours.
These include approaches using multi-level variational autoencoders~\citep{bouchacourt2018multi,vahdat2020nvae,shen2020towards,child2020very} to learn compositional latent spaces, although not in decision-making domains such as ours.
Hierarchical latent approaches have been used in single-agent RL~\citep{haarnoja2018latent,rao2021learning}.
Behavior analysis has also been conducted by embedding agent neuron activations into a low-dimensional space, using a (non-hierarchical) variational approach~\citep{jaderberg2019human}.
The representation power of agents' internal states has also been gauged by predicting future events~\citep{liu2021motor}. 
Overall, the key difference between the above works and ours is that our method combined hierarchical learning with behavior analysis and applies it to the multiagent setting. 

\section{Offline Analysis of Multiagent Behaviors}\label{sec:offline_analysis}
This section introduces the offline multiagent behavioral analysis problem and our proposed algorithm.

\begin{figure}[t]
    \centering
    \begin{subfigure}[b]{0.3\textwidth}
        \centering
        \vspace{-12pt}
        \includegraphics[width=\textwidth,page=1]{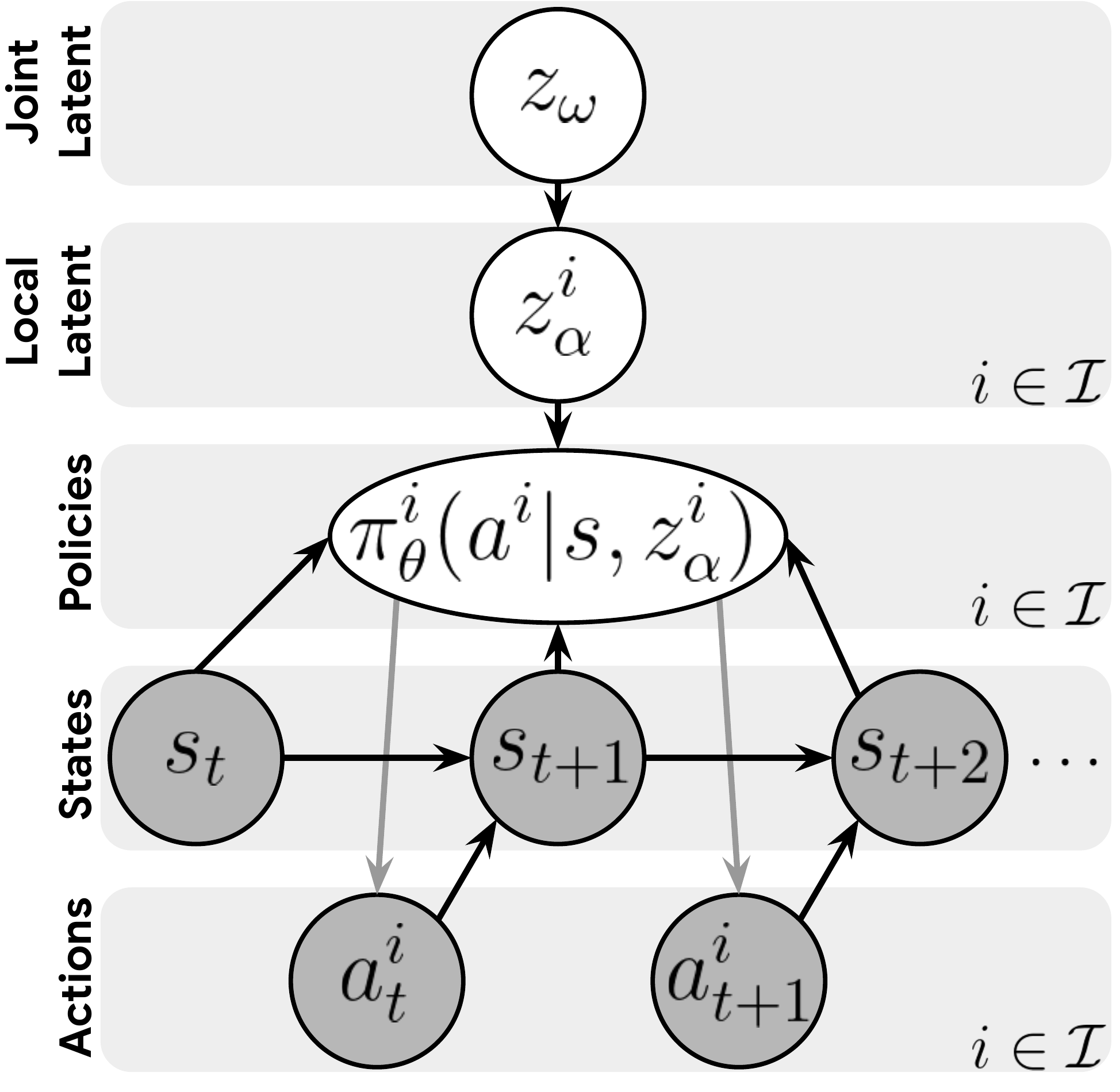}
        \caption{}
        \label{fig:plate_notation}
    \end{subfigure}%
    \hspace{30pt}
    \begin{subfigure}[b]{0.5\textwidth}
        \centering
        \vspace{-12pt}
        \includegraphics[width=\textwidth]{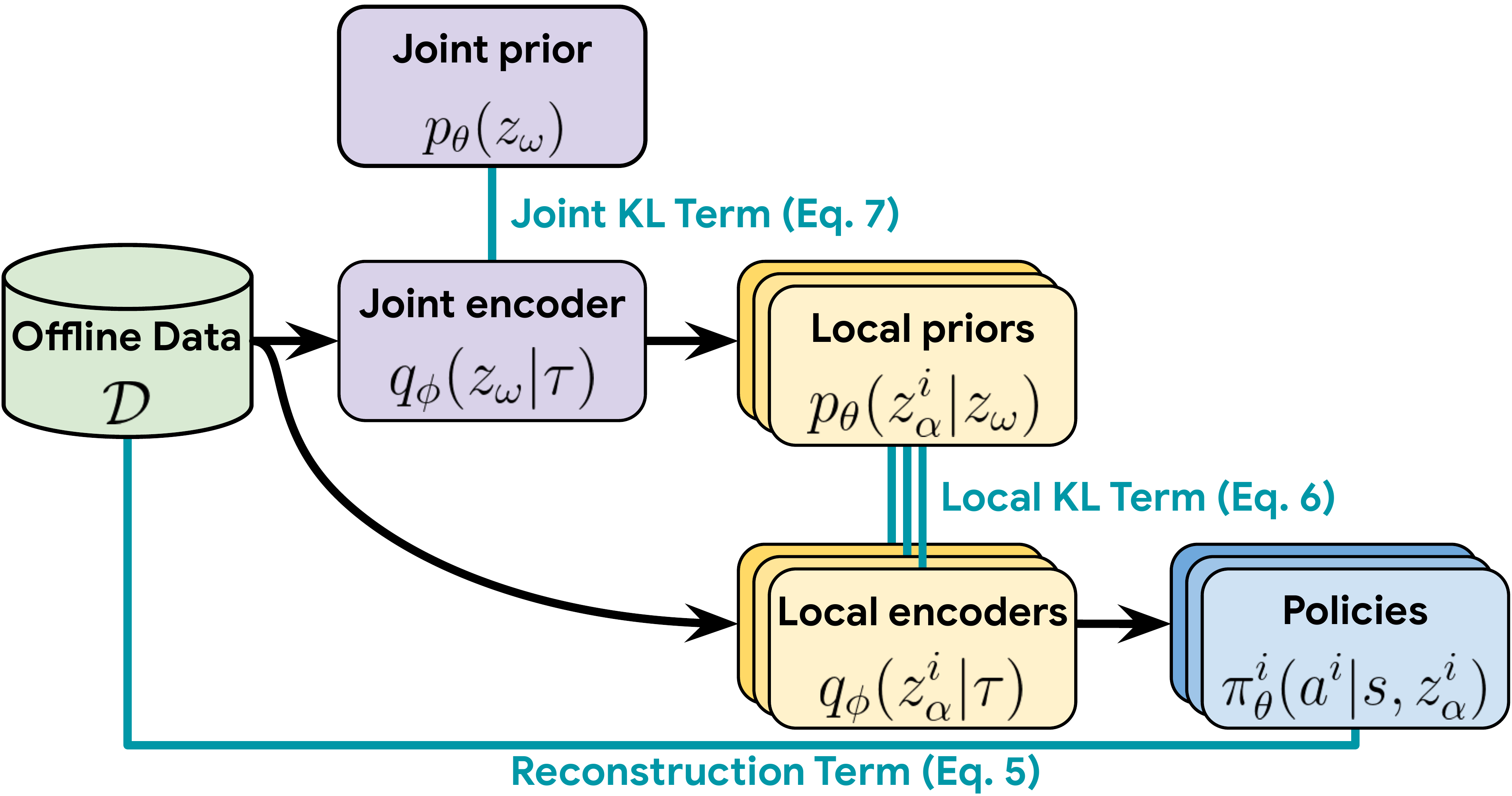}
        \caption{}
        \label{fig:network_arch}
    \end{subfigure}
    \vspace{-6pt}
    \caption{
        Approach overview.
        \subref{fig:plate_notation} Graphical model of the latent-conditioned trajectory generation process that \OurAlgAcronym uses to learn multiagent behavior clusters.
        The joint behavior latent parameter $\zhi$ informs local (agent-wise) behavior parameters $\zlo^i$, which affects their behavioral policies.
        Given a state-action trajectory dataset, our approach learns these joint and local behavior spaces.
        \subref{fig:network_arch}~Architecture of the \OurAlgAcronym model with variational lower bound terms \cref{eq:loss_reconstruction,eq:loss_kl_local,eq:loss_kl_joint} indicated.
    }
    \vspace{-13pt}
\end{figure}

\paragraph{Preliminaries.}
We first formalize the problem of offline multiagent behavioral analysis.
Consider a rewardless multiagent Markov Decision Process (MA-MDP), defined by tuple $(\mathcal{I}, \mathcal{S}, \mathcal{A}, \mathcal{T})$, where $\mathcal{I} = \{1,...,N\}$ is the set of $N$ agents, $\mathcal{S}$ is the state space, $\mathcal{A}$ is the action space, and $\mathcal{T}$ denotes the state transition probability function.
By not relying on the presence of rewards, behaviors generated even without reliance on a reward function (e.g., human interactions, or agents using curiosity-based exploration) can be considered.
We use the term `local' for elements associated with individual agents, and `joint' for those associated with the entire system.
At each timestep $t \in \{0,\ldots,T-1\}$, the agents execute joint action $a_t \in \mathcal{A}$ in state $s_t$ using joint policy $\pi(a_t|s_t)$, causing the state to transition to $s_{t+1}$ with probability $p(s_{t+1}|s_t,a_t) = \mathcal{T}(s_{t+1},a_{t},s_{t})$.
As standard in multiagent frameworks~\citep{oliehoek2016concise}, we assume the joint action space factorizes as $\mathcal{A} = \times_i \mathcal{A}^i$, such that $a_t = (a_t^1, \ldots, a_t^N)$, where $i \in \mathcal{I}$ and $a_t^i \in \mathcal{A}^i$.
Similarly, $\pi^i(a_t^i|s_t)$ is the local policy for agent $i$, and $\pi = \prod_i \pi^i$ is the joint policy.

Let $\tau = \left(s_0, (a_0^i)_{i \in \mathcal{I}}, \ldots, s_{T-1},(a_{T-1}^i)_{i \in \mathcal{I}}, s_T \right)$ denote a trajectory induced by this process, and $\mathcal{D} = \{\tau_1, \ldots, \tau_K\}$ denote a dataset of $K$ such trajectories.
This dataset may consist of trajectories from multiple training runs, including variations over agent algorithms, hyperparameters, random seeds, or other factors influencing emergent behaviors.
Given dataset $\mathcal{D}$, the offline multiagent analysis problem seeks to uncover potential clusters of agent behaviors.

\paragraph{Approach.}
Our approach, called the \OurAlg (\OurAlgAcronym), uses offline trajectory data to discover behaviors exhibited by agents at the local and joint level.

We first use \cref{fig:plate_notation} to build intuition before discussing technical details.
Let the agent interactions exhibited in a trajectory $\tau$ be encoded by a latent variable, $\zhi \in \R^{D_{\omega}}$, capturing their joint behavior.
For example, $\zhi$ may encode (at a high level) whether agents were cooperating or competing in a given trajectory;
conditioned on joint signal $\zhi$, each agent then exhibits its own local behavior.
Let local latent vectors $\zlo = (\zlo^1, \ldots, \zlo^N)$ encode individual agent behaviors, where $\zlo^i \in \R^{D_{\alpha}^i}$.
Conditioned on the local behavior vector $\zlo^i$, each agent then executes actions using a behavior-conditioned policy $\pi^i(a^i|s, \zlo^i)$.
Given trajectory dataset $\mathcal{D}$, we seek to learn the latent-conditioned policies and distributions over latent vectors, such that we can reconstruct \emph{any} behaviors exhibited by the agents in $\mathcal{D}$.
Thus, latent vectors $\zhi$ and $\zlo$ will encode the agents' behavioral spaces and, ideally, identify behavioral clusters in the dataset.
Given this framework, the joint policy is decomposed,
\begin{align}
    \pi(a_t|s_t) &= 
    \int_{\zlo,\zhi}\pi(a_t|s_t,\zlo)p(\zlo|\zhi) p(\zhi)d_{\zlo}d_{\zhi} \label{eq:policy_decomp_intermed}\\
    &= \int_{\zlo,\zhi}\prod_{i=1}^N \pi^i(a_t^i|s_t,\zlo^i) p(\zlo^i|\zhi)p(\zhi)d_{\zlo}d_{\zhi} \label{eq:policy_decomp}\, ,
\end{align}
where in \cref{eq:policy_decomp_intermed} we have assumed that each agent's latent-conditioned policy is conditionally-independent of the high-level latent behavior $\zhi$ given its low-level latent $\zlo^i$ (see \cref{sec:appendix_conditional_indep} for discussion).
Next, given initial state distribution $p(s_0)$ and latent behavior spaces, the probability of a trajectory $\tau$ under joint policy $\pi(\cdot)$ is as follows:
\begin{align}
    p^{\pi}(\tau) 
    &= p(s_0)\prod_{t=0}^{T-1}p(s_{t+1}|s_t,a_t)\pi(a_t|s_t)\label{eq:traj_proba}\\
     &= \int_{\zlo,\zhi} p(s_0)\prod_{t=0}^{T-1}p(s_{t+1}|s_t,a_t)\prod_{i=1}^{N}\pi^i(a_t^i|s_t,\zlo^i) p(\zlo^i|\zhi)p(\zhi) d_{\zlo}d_{\zhi}\, . \label{eq:conditional_tau}
\end{align}

We seek to learn the distributions over variables $\zlo$ and $\zhi$, alongside the latent-conditioned policies $\pi^i(a^i|s_t,\zlo^i)$, which maximize trajectory probabilities~\cref{eq:conditional_tau}.
In \cref{sec:appendix_elbo_derivation}, we derive the following variational lower bound, enabling approximation of these components using parametric models:
\begin{align}
    J_{lb} = &\E_{\tau \sim \mathcal{D}, \zlo \sim q_\phi(\zlo|\tau)}\Big[\sum_{t,i} \log \pi^i_\theta(a_t^i|s_t,\zlo^i)\Big] \label{eq:loss_reconstruction}\\
    - \beta \bigg[&\E_{\tau \sim \mathcal{D}, \zhi \sim q_\phi(\zhi|\tau)} \Big[\sum_i \KL(q_\phi(\zlo^i|\tau) || p_\theta(\zlo^i|\zhi))\Big]   \label{eq:loss_kl_local}\\
    +  &\E_{\tau \sim \mathcal{D}}\left[ \KL(q_\phi(\zhi|\tau)||p_\theta(\zhi))\right] \bigg] \,, \label{eq:loss_kl_joint}
\end{align}
where $q_\phi(\zlo^i|\tau)$ and $p_\theta(\zlo^i|\zhi)$ are, respectively, learned encoder (posterior) and prior distributions over the local behavior latents $\zlo^i$;
likewise, $q_\phi(\zhi|\tau)$ and $p_\theta(\zhi)$ are, respectively, learned encoder and prior distributions over the joint behavior latent $\zhi$;
$\beta$ is a KL-weighting term as in $\beta$-VAEs~\citep{higgins2017beta}.

\Cref{fig:network_arch} illustrates \OurAlgAcronym's model architecture, which is informed by the three bound components~\cref{eq:loss_reconstruction,eq:loss_kl_local,eq:loss_kl_joint}.
During training, each trajectory $\tau \sim \mathcal{D}$ is simultaneously passed through the joint and local encoders, which respectively produce parameters for distributions over $\zhi$ and $\zlo$ (e.g., parameters of Gaussian distributions).
Samples of low-level latent vectors $\zlo$ are passed to the reconstructed agent policies $\pi_\theta^i(a_t^i|s_t,\zlo^i)$, which are trained via the reconstruction component~\cref{eq:loss_reconstruction}. 
The local KL-divergence component~\cref{eq:loss_kl_local} induces the local encoder distribution (which is conditioned directly on $\tau$) to be similar to the local prior distribution (which is conditioned only on samples $\zhi$), thus enabling meaningful correlations between the encoded local and joint latent space, as later shown.
Finally, the joint KL-divergence component~\cref{eq:loss_kl_joint} is akin to that in a standard variational autoencoder~\citep{kingma2014auto,rezende2014stochastic}.
Overall, \OurAlgAcronym enables learning of a hierarchical behavioral space (at the joint and local agent levels, $\zhi$ and $\zlo$, respectively) that exposes interesting behavioral clusters.

\section{Experiments}

We showcase various use-cases for \OurAlgAcronym in a range of domains including continuous coordination games, multiagent MuJoCo~\citep{de2020deep}, and OpenAI hide-and-seek~\citep{Baker2020Emergent}.
\Cref{sec:appendix_exp_details} provides data generation, networks, computation, and hyperparameter details.
\Cref{sec:model_code} provides pseudocode.

\paragraph{Data generation.} Multiagent trajectory data is generated for each domain via the Acme RL library~\citep{hoffman2020acme}, using the TD3 algorithm~\citep{fujimoto2018addressing} in a decentralized MARL fashion, with datasets managed using RLDS~\citep{ramos2021rlds}.
Trajectories are collected at constant intervals throughout training, which also enables analysis of behavioral emergence.
We conduct a wide sweep over random seeds for data generation, yielding a diverse trajectory dataset (see \cref{appendix:additional_dataset_details} for dataset details and statistics).

\paragraph{\OurAlgAcronym setup.} To analyze the above data using \OurAlgAcronym, we use a GMM (Gaussian Mixture Model) for the joint prior, a bidirectional LSTM (long short-term memory network) with GMM head for the joint encoder, an MLP (multi-layer perceptron) with Gaussian head for the local priors, a bidirectional LSTM with Gaussian head for the local encoder, and an MLP for reconstructed policies. 
GMMs are used for the joint prior and encoder as they produce discernible joint behavior clusters~\citep{jiangijcai2017}, whereas the conditioning of the local prior on $\zhi$ yields such clusters at the local level with a standard Gaussian head.
We use parameter-sharing across local priors, local encoders, and reconstructed policies, as common in multiagent setups~\citep{christianos2021scaling}, with a unique one-hot vector identifier appended to agent-specific network inputs to enable heterogeneity in model outputs.

\paragraph{Independent analysis of joint ($\zhi$) and local ($\zlo$) behaviors.}
\begin{figure}[t]
    \newcommand{\myFigVSpace}{3pt}
    \captionsetup[subfigure]{aboveskip=-3pt,belowskip=-3pt}
    \centering
    \begin{subfigure}[b]{0.32\textwidth}
        \centering
        \includegraphics[width=\textwidth]{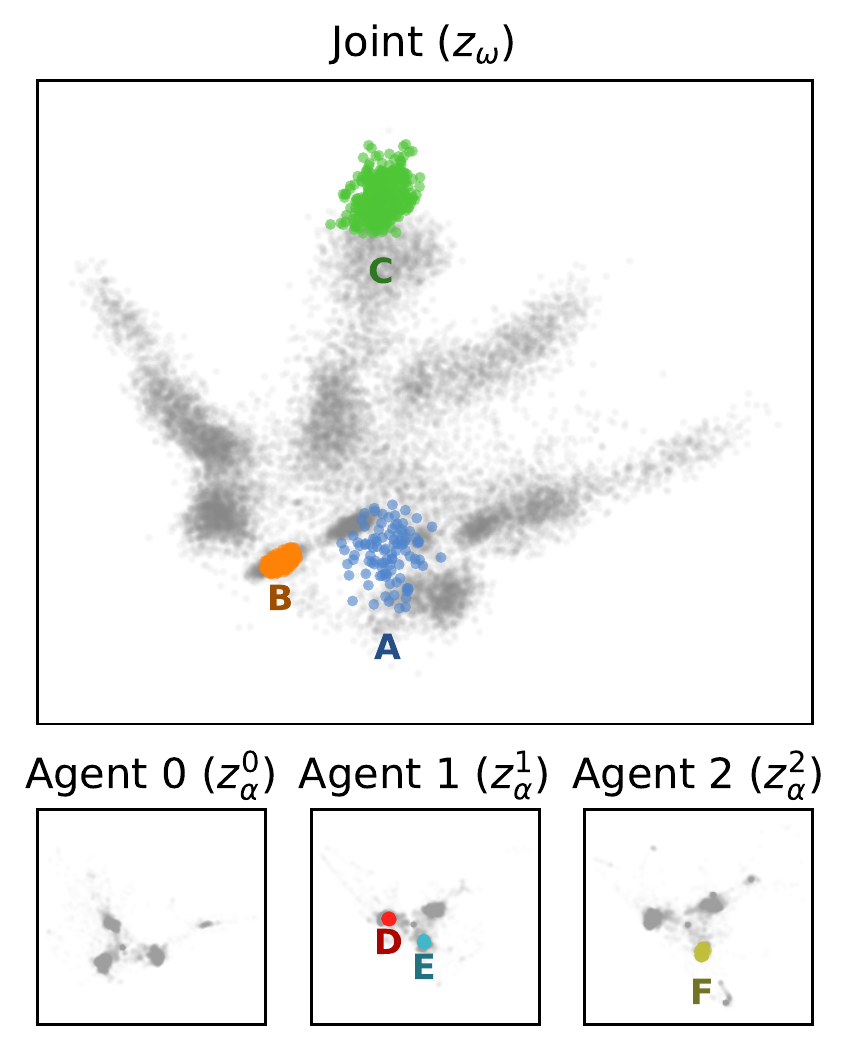}
        \caption{}
        \label{fig:multimodal_reward_env_clusters_latent}
    \end{subfigure}%
    \includegraphics[width=0.2\textwidth]{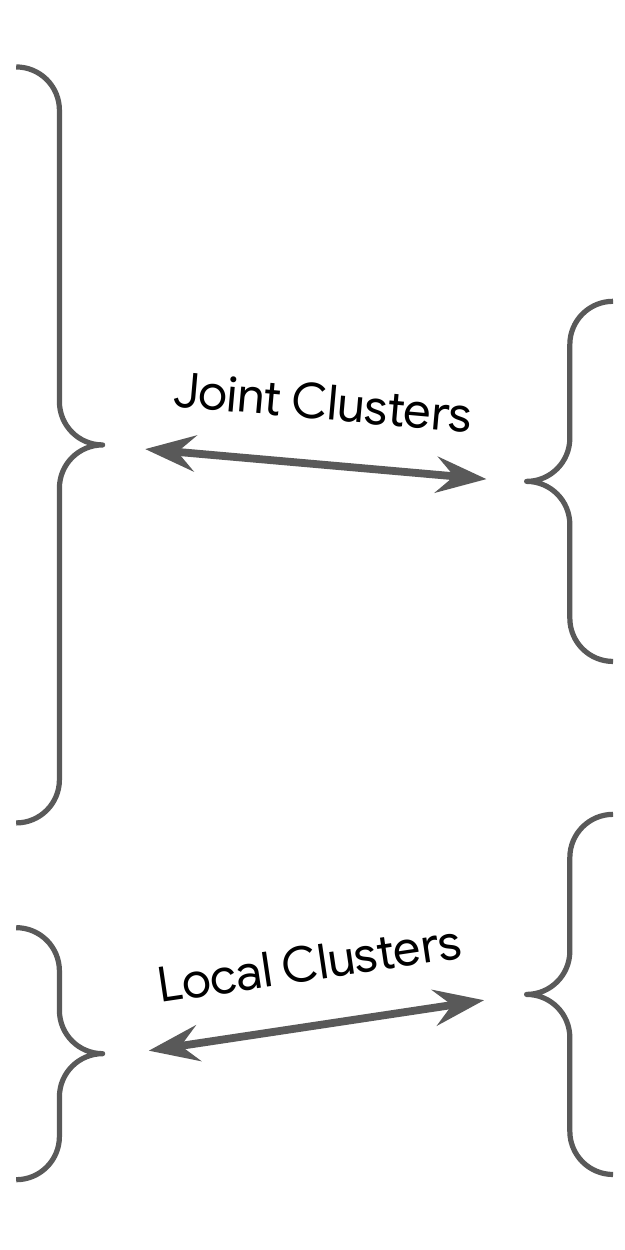}%
    \begin{subfigure}[b]{0.42\textwidth}
        \centering
        \includegraphics[width=70pt]{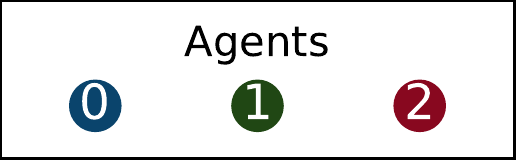}\\
        \includegraphics[width=\textwidth/3]{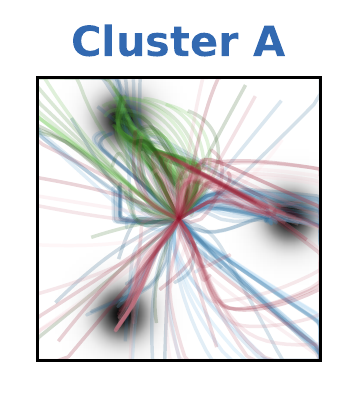}%
        \includegraphics[width=\textwidth/3]{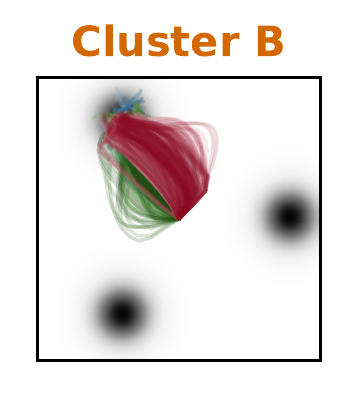}%
        \includegraphics[width=\textwidth/3]{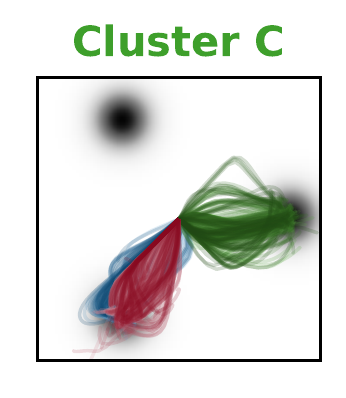}\\
        \includegraphics[width=\textwidth/3]{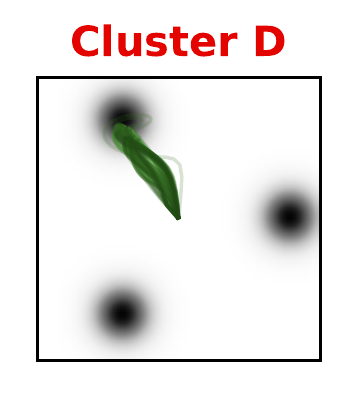}%
        \includegraphics[width=\textwidth/3]{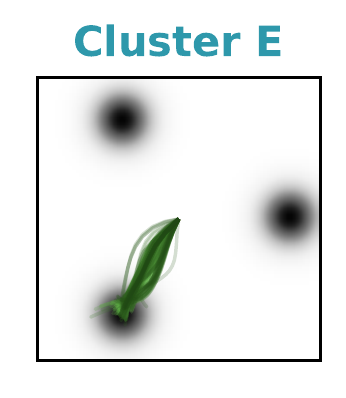}%
        \includegraphics[width=\textwidth/3]{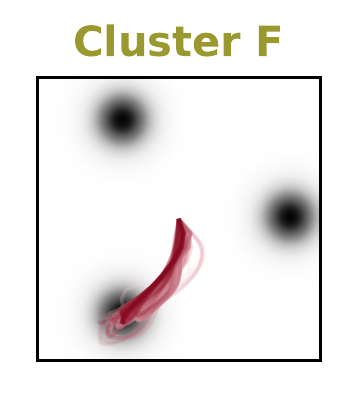}
        \caption{}
        \label{fig:multimodal_reward_env_clusters_trajs}
    \end{subfigure}
    \caption{
        Results for 3-agent hill climbing domain (see interactive version {\color{blue}\href{https://storage.googleapis.com/mohba-beyond-rewards-22/n22/3hill/interactive_final.html}{here}}).
        \subref{fig:multimodal_reward_env_clusters_latent} Example behavioral clusters discovered by \OurAlgAcronym.
        \subref{fig:multimodal_reward_env_clusters_trajs} Trajectories corresponding to each cluster, with reward-hills shown in grey.
        Clusters A to C show joint behaviors, whereas D to F separately show local agent behaviors (with other agents faded in local trajectory plots for readability).
        }
    \vspace{-18pt}
    \label{fig:multimodal_reward_env_clusters}
\end{figure}

We here analyze the hierarchical latent structure learned by \OurAlgAcronym.
Specifically, we highlight differences in agent behaviors as identified at the joint and local levels, respectively, by $\zhi$ and $\zlo$.
We first conduct a simple sanity check in a 3-agent hill-climbing domain (our earlier example in \cref{fig:multimodal_reward_env_trials}).
States and actions correspond, respectively, to 2D positions $(x,y)$ and forces $(\Delta x, \Delta y)$ imparted by each agent for movement.
Agents spawn at the origin and are each rewarded for climbing any of three hills (shown in grey in the domain figures) within the domain in episodes of $50$ timesteps.
We generate a dataset using $50$ independent MARL training trials, each conducted for $1e5$ environment steps; trajectories are saved every $200$ steps, yielding $1.25e6$ frames of data ($25000$ trajectories).

\Cref{fig:multimodal_reward_env_clusters_latent} highlights several example clusters of joint and local behaviors identified by \OurAlgAcronym, using samples of $\zhi$ and $\zlo$ from the joint and local encoders.
We use Euclidean distances in the original latent spaces to identify nearby vectors, then visualize their 2D projection using principal component analysis.
For each cluster, we visualize all associated trajectories in \cref{fig:multimodal_reward_env_clusters_trajs}, thus enabling analysis of behaviors captured in latent space.
Clusters A, B, and C correspond to examples of joint behaviors ($\zhi$).
Cluster A contains trajectories early in training, where agents have not learned to converge to a particular hill in the domain.
In Cluster B, all agents have learned to navigate towards the top-left hill.
In Cluster C, agents 0 and 2 prefer the bottom-left hill, whereas agent 1 prefers the right hill.
Overall, clusters at the joint level capture meaningful collective behaviors of the agents.

Next, we analyze individual agent clusters.
Note that for all agents, three prominent clusters are apparent in their respective $\zlo$ spaces, as each training trial in the data generation process can lead to various agent-wise hill preferences.
We compare two such clusters, D and E, for agent 1, observing in the trajectory plots that these corresponds to this agent preferring the top-left and bottom-left hills, respectively.
Similarly, Cluster F corresponds to trajectories where agent 2 prefers the bottom-left hill.
These results illustrate that the local latents reasonably disentangle each agent's observed behaviors.

\begin{figure}[t]
    \centering
    \newcommand{\myFigSep}{1pt}
    \newcommand{\myFigWidth}{0.195\textwidth}
    \newcommand{\myFigVSpace}{3pt}
    \captionsetup[subfigure]{aboveskip=-2pt,belowskip=-3pt}
    \begin{subfigure}[b]{\myFigWidth}
        \centering
        \includegraphics[width=\textwidth]{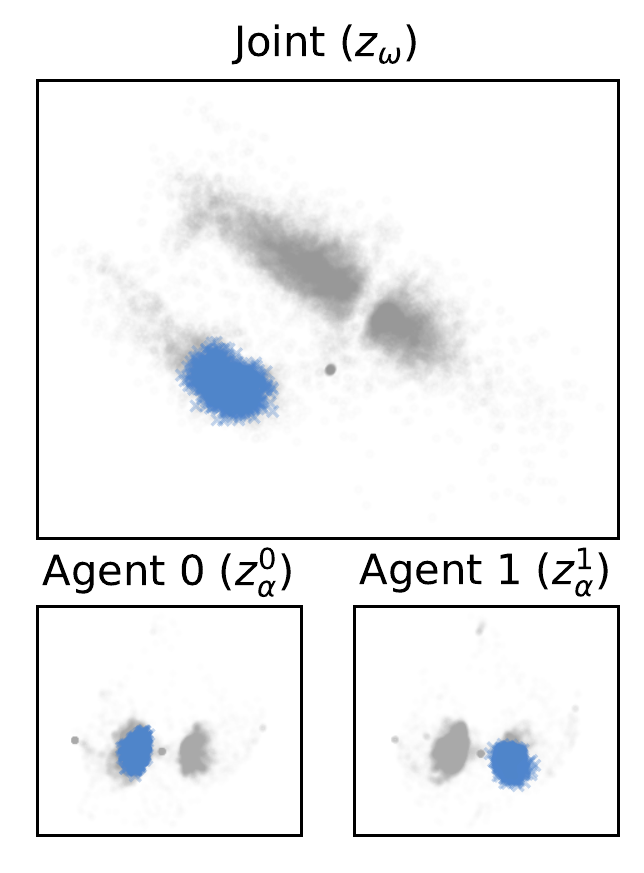}\\
        \includegraphics[width=\textwidth]{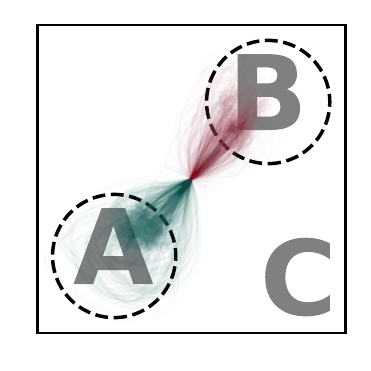}\\
        \phantomgraphics[width=40pt]{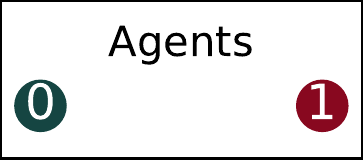}
        \vspace{3pt}
        \caption{}
        \label{fig:left_right_env_clusters_traj_hist_A}
    \end{subfigure}%
    \hspace{\myFigSep}
    \rulesep
    \hspace{\myFigSep}
    \begin{subfigure}[b]{\myFigWidth}
        \centering
        \includegraphics[width=\textwidth]{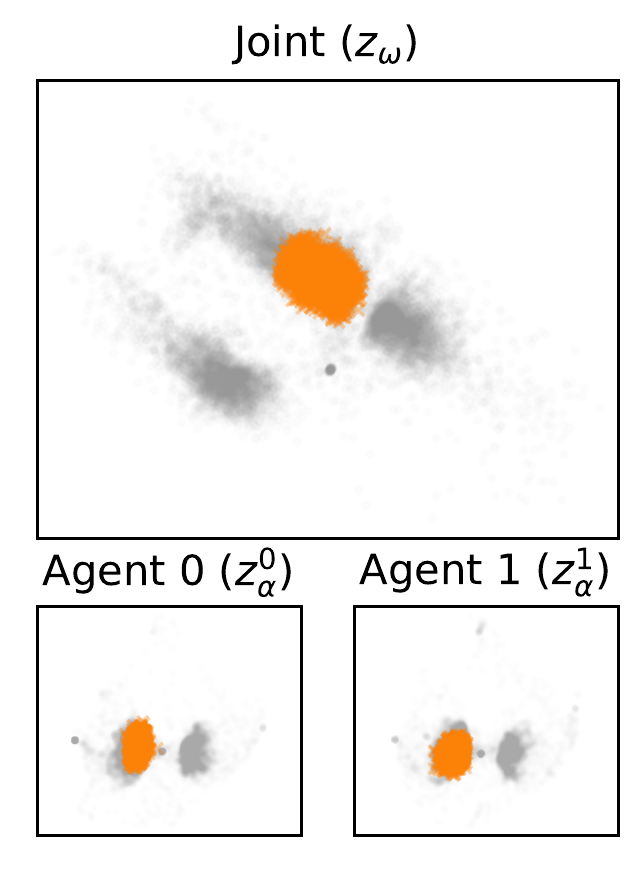}\\
        \includegraphics[width=\textwidth]{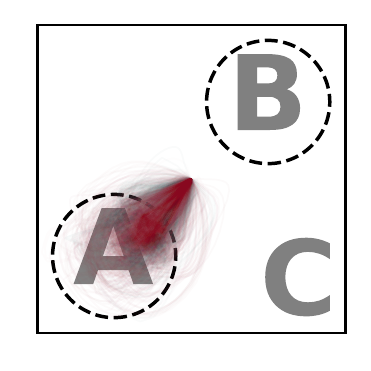}\\
        \includegraphics[width=40pt]{figs/left_right_env_agent_legend.pdf}
        \vspace{3pt}
        \caption{}
        \label{fig:left_right_env_clusters_traj_hist_B}
    \end{subfigure}%
    \hspace{\myFigSep}
    \rulesep
    \hspace{\myFigSep}
    \begin{subfigure}[b]{\myFigWidth}
        \centering
        \includegraphics[width=\textwidth]{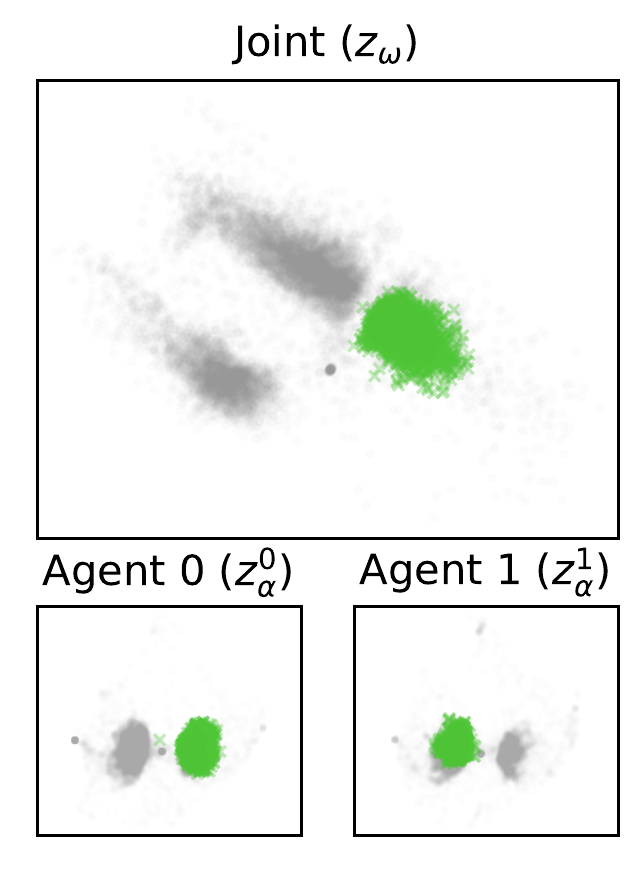}\\
        \includegraphics[width=\textwidth]{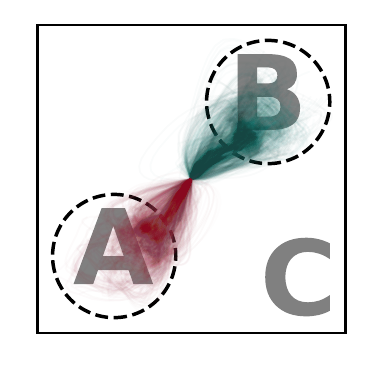}\\
        \phantomgraphics[width=40pt]{figs/left_right_env_agent_legend.pdf}
        \vspace{3pt}
        \caption{}
        \label{fig:left_right_env_clusters_traj_hist_C}
    \end{subfigure}
    \hspace{\myFigSep}
    \rulesep
    \hspace{\myFigSep}
    \begin{subfigure}[b]{0.17\textwidth}
        \centering
        \includegraphics[width=0.7\textwidth]{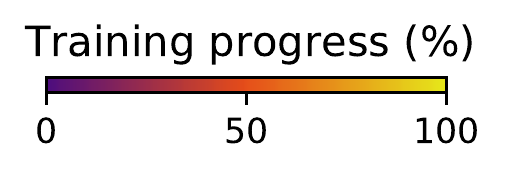}\\
        \includegraphics[width=\textwidth]{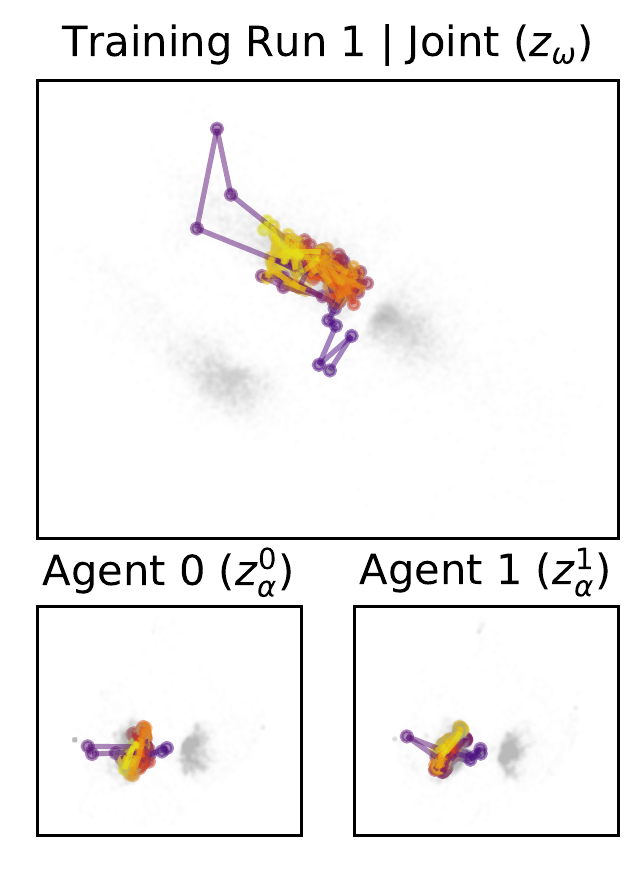}\\
        \hrule
        \includegraphics[width=\textwidth]{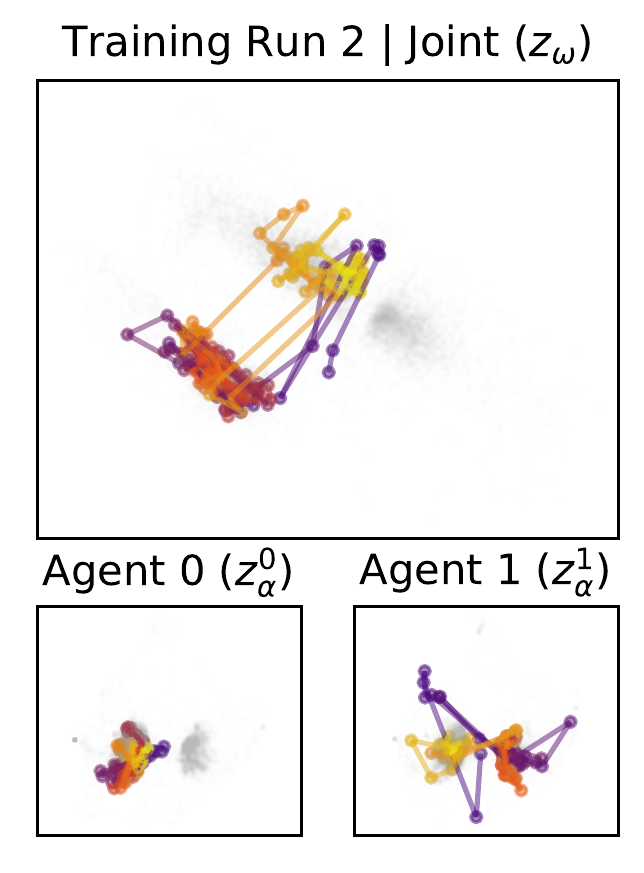}
        \caption{}
        \label{fig:left_right_env_rl_track}
    \end{subfigure}%
        \caption{
        Results for two-agent coordination game (see interactive version {\color{blue}\href{https://storage.googleapis.com/mohba-beyond-rewards-22/n22/2agents/interactive_final.html}{here}}).
        \subref{fig:left_right_env_clusters_traj_hist_A} to \subref{fig:left_right_env_clusters_traj_hist_C} show key joint clusters, the coupled local clusters for each agent, and associated trajectories in the domain (over all 50 MARL training runs in the dataset).
        \subref{fig:left_right_env_rl_track} visualizes the progression of agent behaviors throughout the original MARL training phase, for two example training runs (top and bottom panels).
    }
    \vspace{-14pt}
    \label{fig:left_right_env_clusters}
\end{figure}

\begin{wraptable}[7]{h}{5cm}
    \vspace{-\intextsep}
    \centering
    \setlength{\extrarowheight}{1pt}
    \caption{Coordination rewards.}
    \vspace{-8pt}
    \label{table:bimatrix_game}
    \begin{tabular}{cc|c|c|c|}
      & \multicolumn{1}{c}{} & \multicolumn{3}{c}{\emph{Agent 1}}\\
      & \multicolumn{1}{c}{} & \multicolumn{1}{c}{$A$}  & \multicolumn{1}{c}{$B$} & \multicolumn{1}{c}{$C$} \\\cline{3-5}
      \parbox[t]{1mm}{\multirow{3}{*}{\rotatebox[origin=c]{90}{\emph{Agent 0}}}}
      & $A$ & $(1,1)$ & $(1,1)$ & $(0,0)$\\\cline{3-5}
      & $B$ & $(1,1)$ & $(0,0)$ & $(0,0)$\\\cline{3-5}
      & $C$ & $(0,0)$ & $(0,0)$ & $(0,0)$\\\cline{3-5}
    \end{tabular}
\end{wraptable}
\paragraph{Coupled analysis of joint ($\zhi$) and local ($\zlo$) behaviors.} 
The above experiments independently analyzed the joint and local latent spaces.
We can also concurrently analyze them to better understand local agent contributions to joint behaviors.
Consider a two-agent domain with close inter-agent coordination (visualized in the top of \cref{fig:left_right_env_clusters_traj_hist_A}), with state and actions-spaces similar to the hill domain and episodes consisting of 50 timesteps.
Three regions are defined in this domain: 
$A$ and $B$ (circular regions), and $C$ (region exterior to the circles).
When an agent enters a given region, it `activates' the corresponding strategy in \cref{table:bimatrix_game}, with agents receiving rewards at each timestep according to the joint strategy they have activated.
For example, if both agents enter region $A$, they each receive a reward of $1$, whereas if one agent enters $A$ while the other is in exterior region $C$, neither receives a reward.
This domain involves a significant degree of coordination as agents must discover the rewarding regions, while receiving a sparse reward signal until a valid combination of strategies is discovered.
There is also potential for miscoordination:
navigating to region $B$ is rewarding assuming the other agent navigates to $A$, but yields 0 reward if the other agent instead navigates to $B$ (potentially destabilizing training).
We run 50 independent MARL sweeps in this domain, yielding $1e6$ data frames ($20000$ trajectories). 
As shown in \cref{fig:left_right_env_clusters}, \OurAlgAcronym discovers three dominant joint behavior clusters $\zhi$ here. 
In each of \cref{fig:left_right_env_clusters_traj_hist_A,fig:left_right_env_clusters_traj_hist_B,fig:left_right_env_clusters_traj_hist_C}, we highlight one of these clusters and its corresponding local behavior latents $\zlo$ for each agent.
\OurAlgAcronym reveals that across the dataset, the agents have learned to cover all 3 optimal joint behaviors in \cref{table:bimatrix_game}: $(A, A)$, $(A,B)$, and $(B,A)$.
Moreover, despite each agent discovering two local behaviors ($A$ and $B$), $\zhi$ highlights only the 3 observed joint behaviors (i.e., does not simply highlight 4 clusters consisting of the Cartesian product of individual agents' behavior spaces).

\paragraph{Behavior emergence throughout MARL training.}
We next use \OurAlgAcronym's latent space to inspect behavior emergence \emph{during} MARL training.
\Cref{fig:left_right_env_rl_track} visualizes the training progression of two MARL runs used for data generation, tracking behavior changes throughout.
In Training Run 1 (top panel of \cref{fig:left_right_env_rl_track}), the agents converge to and maintain a fixed joint behavior throughout training.
By contrast, Training Run 2 (bottom panel) has numerous behavior changepoints, where agents flip back and forth between preferring one of two clusters in $\zhi$.
Concurrently inspecting the $\zlo$ space for Training Run 2, we observe that agent 0 converges to and maintains a consistent behavior, whereas agent 1 changes its preference sporadically, also explaining the detected changes in joint behaviors.

\paragraph{Baseline comparisons.}
We next compare against baselines previously used for multiagent behavioral analysis in the literature~\citep{jaderberg2019human,liu2021motor}.
An LSTM baseline conducts next-action prediction at each step in a trajectory $\tau$, using an action-prediction loss (APL), defined as the $\Ltwo$ loss over predicted vs. ground truth actions. 
This baseline targets using the LSTM hidden states, rather than a learned distribution over latent variables, for understanding and clustering agent behaviors (akin to the analysis in~\citep{liu2021motor,jaderberg2019human}). 
A flat-VAE baseline provides a non-hierarchical ablation of \OurAlgAcronym that simply feeds the joint latent $\zhi$ (rather than $\zlo^i$) to reconstructed agent policies (i.e., $\pi_\theta^i(a^i| s, \zhi)$);
we use our usual loss (with the local KL term \cref{eq:loss_kl_local} removed) to train this VAE.
We conduct a hyperparameter sweep for the baselines (see \cref{sec:appendix_exp_details}), reporting the best results averaged over 3 random seeds.

\begin{wraptable}[8]{h}{9.7cm}
    \vspace{-\intextsep}
    \centering
    \setlength{\tabcolsep}{0.5em}
    \caption{Baseline comparisons. Action-prediction loss (APL) and intra-cluster trajectory distance (ICTD); lower is better for both.}
    \vspace{-8pt}
    \label{table:baseline_comparisons}
    \begin{tabular}{@{\extracolsep{2pt}}ccccc}
        \toprule
        & \multicolumn{2}{c}{Hill-climbing domain} & \multicolumn{2}{c}{Coordination game}\\
        \cline{2-3}  \cline{4-5}
        & APL & ICTD & APL & ICTD\\
        \cline{2-2} \cline{3-3}  \cline{4-4} \cline{5-5}
        LSTM & $4.8 \pm 0.1$ & $0.47 \pm 0.19$ &  $2.9 \pm 0.1$ & $0.34 \pm 0.15$\\
        VAE & $6.9 \pm 0.4$ & $\mathbf{0.18 \pm 0.13}$ & $5.0 \pm 0.3$ & $\mathbf{0.16 \pm 0.10}$\\
        \OurAlgAcronym & $\mathbf{1.3 \pm 0.1}$ & $\mathbf{0.17 \pm 0.13}$ & $\mathbf{2.4 \pm 0.3}$ & $\mathbf{0.16 \pm 0.10}$\\[-3pt]
        \bottomrule
    \end{tabular}
\end{wraptable}
Comparisons are conducted at the joint behavior level:
for the LSTM, we use the final hidden state as an encoding of each trajectory; 
for our method and the VAE, we use $\zhi$ directly.
\Cref{table:baseline_comparisons} compares the methods in the hill-climbing and coordination game domains.
Here we note the APL, which measures how well each method reconstructs ground truth policies.
Additionally, we use K-means (sweeping over the \# of clusters) to identify behavior clusters for each method, then report the intra-cluster trajectory distance (ICTD), defined as the average distance of all trajectories in a cluster from the mean trajectory in said cluster (akin to intra-cluster point scatter, a common cluster analysis statistic~\citep{hastie2009elements}) .
Combined, these measures provide a proxy for evaluating the latent representations in terms of enabling accurate reconstructions (APL) while clustering similar trajectories together (ICTD).
\OurAlgAcronym significantly outperforms the LSTM and VAE baselines in terms of APL, while also clustering similar trajectories at the $\zhi$ level in terms of ICTD.
We provide additional results including APL throughout training, full ICTD sweeps, and visualizations of the latent spaces for the baselines in \cref{appendix:additional_baseline_experiments}.

\paragraph{Behavior concept discovery.}
Next, we test the representation power of the latent spaces learned by \OurAlgAcronym, and illustrate a means of discovering ‘behavior concepts’ in the latent space.
We adopt the completeness-aware concept explanations framework of~\citet{yeh2020completeness}, with slight changes to make it amenable to this setting (see \cref{sec:appendix_concept_discovery_details}).
At a high-level, given a set of inputs, a set of concept vectors $C$ in the same space as inputs, and prediction targets (e.g., domain characteristics of interest), \citet{yeh2020completeness} define a framework to compute class-conditioned Shapley values, called ConceptSHAP.
To do so, every input is projected onto each of the concept vectors, yielding a vector of `concept scores', which is then passed to a simple prediction head $g$ (e.g., small MLP or linear model) to predict the targets and compute the ConceptSHAP.
ConceptSHAP provides numerical scores interpreted as the importance of each concept in $C$ for predicting a given target class, which is useful for identifying key concepts (and nearby inputs) associated with certain domain characteristics.

\begin{table}[t]
    \caption{Discovered concepts using $\zhi$ in hill-climbing domain. For each characteristic (agent dispersion and return), 5 classes are constructed using the ground truth information in trajectories $\tau$. The concept explanation framework of~\citet{yeh2020completeness} is then used to first predict the correct classes (using only $\zhi$, rather than the trajectory $\tau$, as input), then identify core concepts related to each class.
    }
    \label{table:user_concept_discovery_multimodal}
    \centering
    \setlength{\tabcolsep}{0.3em} 
    \begin{tabular}{cccccc}
    \toprule
    Concept & \multicolumn{5}{c}{Top discovered concept using $\zhi$ (per class, with average concept measure beneath)}\\
    \midrule
    \multirow[b]{3}{*}[-0.25cm]{\parbox{2cm}{\centering Agent\\ Dispersion\\[0.2cm](Classification accuracy: 54.27\%)}}
    & Class 0 & Class 1 & Class 2 & Class 3 & Class 4\\
    \cline{2-6}
    &
    \includegraphics[width=\textwidth/8]{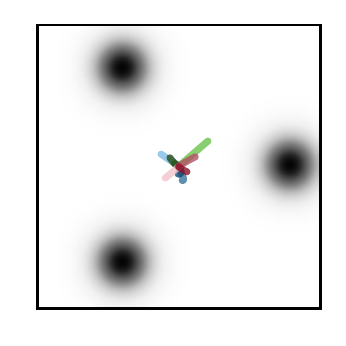}%
    &
    \includegraphics[width=\textwidth/8]{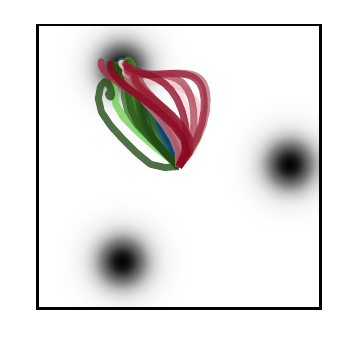}%
    &
    \includegraphics[width=\textwidth/8]{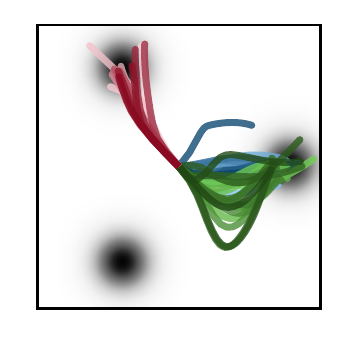}%
    &
    \includegraphics[width=\textwidth/8]{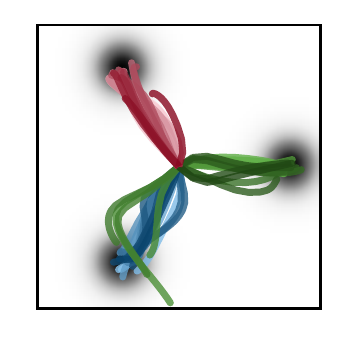}%
    &
    \includegraphics[width=\textwidth/8]{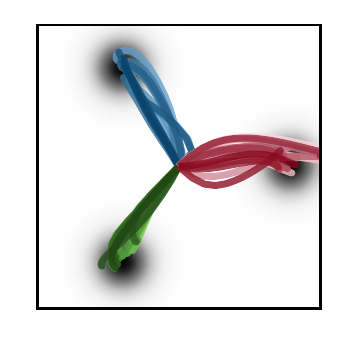}\\[-8pt]
     & $0.05 \pm 0.13$ & $0.15 \pm 0.06$ & $1.79 \pm 0.10$ & $2.18 \pm 0.19$ & $2.36 \pm 0.10$\\
    \midrule
    \multirow[b]{2}{*}[1.4cm]{\parbox{2cm}{\centering Total\\ Return\\[0.2cm](Classification accuracy: 60.98\%)}}
    &
    \includegraphics[width=\textwidth/8]{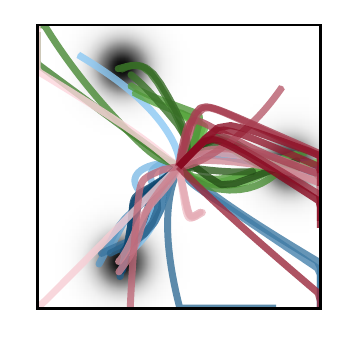}%
    &
    \includegraphics[width=\textwidth/8]{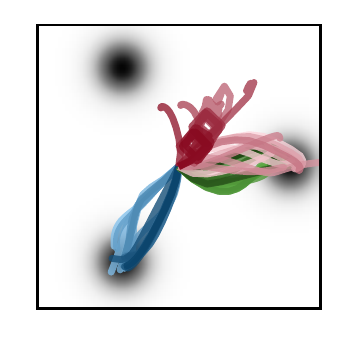}%
    &
    \includegraphics[width=\textwidth/8]{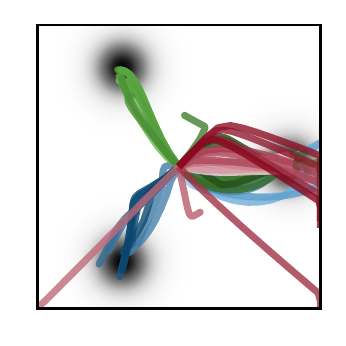}%
    &
    \includegraphics[width=\textwidth/8]{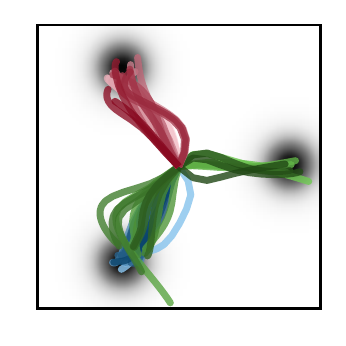}%
    &
    \includegraphics[width=\textwidth/8]{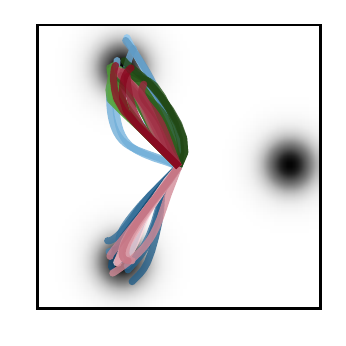}\\[-8pt]
     & $361.0 \pm 256.0$ & $671.4 \pm 115.1$ & $611.1 \pm 244.2$ & $764.9 \pm 72.4$ & $820.4 \pm 155.0$\\
    \bottomrule
    \end{tabular}
    \vspace{-15pt}
\end{table}

In~\cref{table:user_concept_discovery_multimodal}, we use this technique to identify concepts in $\zhi$ space associated with varying characteristics in the hill-climbing domain.
The concept set $C$ considered is generated using K-means in the $\zhi$ space (see \cref{sec:appendix_concept_discovery_details} for details).
The first row of \cref{table:user_concept_discovery_multimodal} shows classes corresponding to increasing levels of agent dispersion in the domain (with dispersion defined as the sum of $\Ltwo$-distances of agents from their centroid at the final timestep).
For each trajectory $\tau$, we compute the associated dispersion, creating $5$ classes of equal-sized bins (labels 0 through 4 mapping to bins of increasing dispersion).
We create an 80-20 train-validation split, then train a 2-layer (8 hidden units each) MLP $g$ via a softmax-cross entropy loss to predict the classes using only $\zhi$ as input (rather than the actual trajectory $\tau$).
We attain a validation accuracy of 54.27\%, signifying the predictive capabilities of $\zhi$.
Next, we compute the class-conditioned ConceptSHAP, thus identifying the top-scoring concept vector for each class.
In \cref{table:user_concept_discovery_multimodal}, we visualize the $20$ trajectories with the closest $\zhi$ to the top-scoring concept for each class (with the mean agent dispersion across these trajectories listed under each image).
Intuitively, the identified concepts involve agents becoming increasingly dispersed, with agents first nearly stationary at the center (class 0), then all converging to the same hill (class 1), spreading across two hills (class 2), and finally covering all hills (classes 3 and 4).

The second row of \cref{table:user_concept_discovery_multimodal} repeats this experiment, now using classes associated with agents' sum-of-returns.
We attain a validation set accuracy of 60.98\%, and observe behaviors associated with generally increasing reward by identifying top clusters in each class (with some overlaps, e.g., classes 1 and 2 with high standard deviation in returns). 
Overall, these experiments help quantify the representational capacity of $\zhi$ and identify clusters associated with distinct behavioral characteristics.

\paragraph{Scalability to high-dimensional domains.}
\begin{figure}[t]
    \centering
    \newcommand{\myFigHeight}{0.9cm}
    \newcommand{\myHspace}{0.25cm}
    \newcommand{\myVspace}{0.3cm}
    \captionsetup[subfigure]{aboveskip=-3pt,belowskip=-3pt}
    \definecolor{mujoco_example_A}{rgb}{0.133, 0.403, 0.168}
    \definecolor{mujoco_example_B}{rgb}{0.21, 0.32, 0.60}
    \definecolor{mujoco_example_C}{rgb}{0.7, 0.03, 0.03}
    \begin{subfigure}[b]{0.25\textwidth}
        \includegraphics[width=\textwidth]{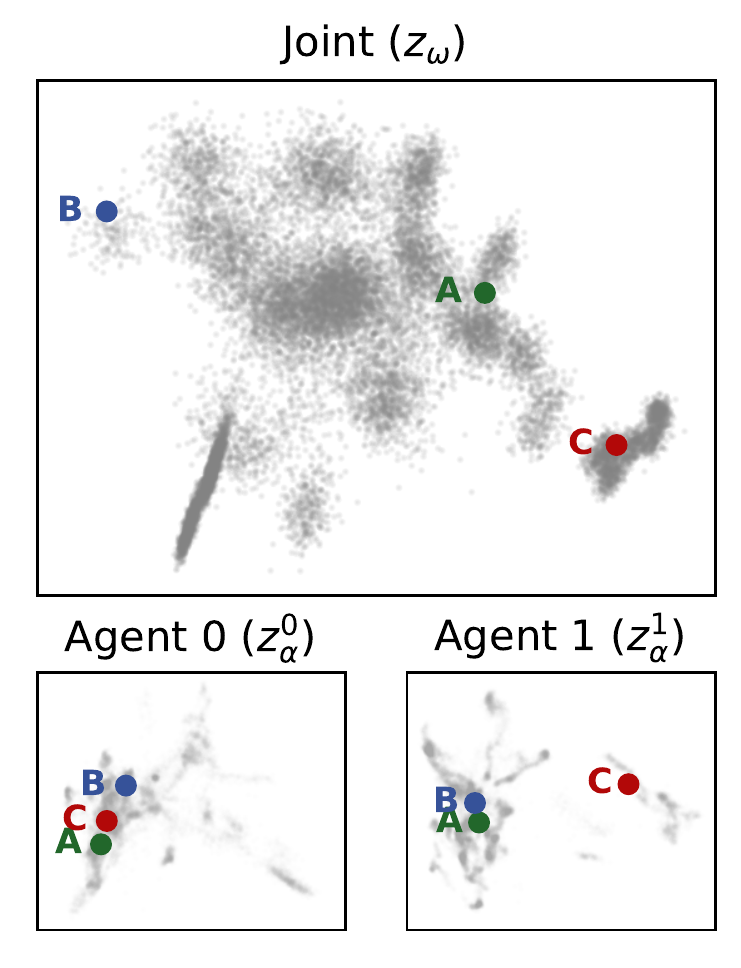}
        \caption{}
        \label{fig:mujoco_halfcheetah_latent_highlighted}
    \end{subfigure}%
    \hspace{10pt}
    \begin{subfigure}[b]{0.6\textwidth}
        \centering
        \textcolor{mujoco_example_A}{Example A (see animated version {\color{blue}\href{https://storage.googleapis.com/mohba-beyond-rewards-22/n22/multiagent_half_cheetah/trajectory_anims/im_w_4_e_1112.gif}{here}})}\\
            \includegraphics[height=\myFigHeight]{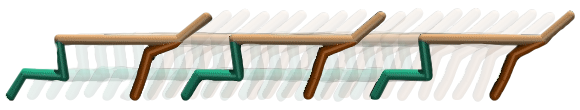}
            \hspace{\myHspace}
            \rulesep
            \hspace{\myHspace}
            \includegraphics[height=\myFigHeight]{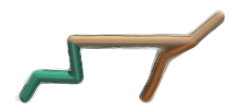}\\[\myVspace]
        \textcolor{mujoco_example_B}{Example B (see animated version {\color{blue}\href{https://storage.googleapis.com/mohba-beyond-rewards-22/n22/multiagent_half_cheetah/trajectory_anims/im_w_23_e_1084.gif}{here}})}\\
            \includegraphics[height=\myFigHeight]{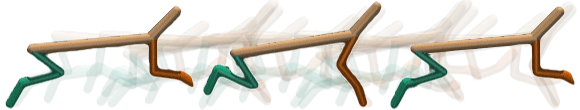}%
            \hspace{\myHspace}
            \rulesep
            \hspace{\myHspace}
            \includegraphics[height=\myFigHeight]{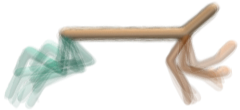}\\[\myVspace]
        \textcolor{mujoco_example_C}{Example C (see animated version {\color{blue}\href{https://storage.googleapis.com/mohba-beyond-rewards-22/n22/multiagent_half_cheetah/trajectory_anims/im_w_25_e_1562.gif}{here}})}\\
            \includegraphics[height=\myFigHeight]{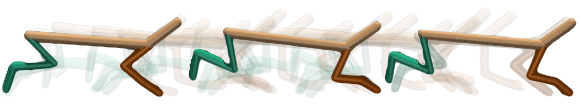}%
            \hspace{\myHspace}
            \rulesep
            \hspace{\myHspace}
            \includegraphics[height=\myFigHeight]{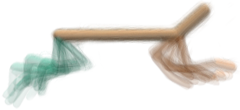}
        \caption{}
        \label{fig:mujoco_examples_zomega_zalpha}
    \end{subfigure}
    \caption{
        Multiagent MuJoCo HalfCheetah behavioral space (see interactive version {\color{blue}\href{https://storage.googleapis.com/mohba-beyond-rewards-22/n22/multiagent_half_cheetah/interactive_final.html}{here}}).
        Agent 0 and 1, respectively, control the back and front limbs, coordinating to move the cheetah. 
        \subref{fig:mujoco_halfcheetah_latent_highlighted} Detected behavioral space, with three example trajectories indicated.
        \subref{fig:mujoco_examples_zomega_zalpha} HalfCheetah behavior corresponding to the same three examples (left panels showing trajectory frames, and right panels showing the same frames with the cheetah torso aligned to disambiguate the back and front leg agent behaviors).
        }
    \vspace{-7pt}
    \label{fig:mujoco_halfcheetah}
\end{figure}

We next test \OurAlgAcronym's scalability to high-dimensional domains using the 2-Agent HalfCheetah multiagent MuJoCo domain~\citep{de2020deep}, where agents 0 and 1, respectively, control the back and front limbs of a cheetah to coordinate movement.
Each agent's state consists of a 6D vector summarizing velocity and position information for the 3 joints it controls, with its 3D action space corresponding to motor torques applied to these joints. 
The agents coordinate to maximize the forward-speed of the cheetah over episodes of 200 timesteps each.
We generate data using 30 independent trials of MARL training, collecting $1e5$ total trajectories to train \OurAlgAcronym.

\Cref{fig:mujoco_halfcheetah_latent_highlighted} visualizes the behavior space learned by \OurAlgAcronym, wherein we observe several clusters.
We highlight three example trajectories stemming from distinct clusters in the joint behavior space $\zhi$, showing the corresponding HalfCheetah behaviors in \cref{fig:mujoco_examples_zomega_zalpha}.
The left panel of each example provides a view of the cheetah's overall movement, where we observe key behavior differences: 
in Example A, the cheetah runs forward using subtle vibration of its limbs;
in Example B, it bounces forward with its torso arched up due to its front limb (agent 1) being more extended than its back limb (agent 0);
in Example C, it moves closer to the ground with its torso arched down.
In the latent space (\cref{fig:mujoco_halfcheetah_latent_highlighted}), we observe that for Examples B and C, the $\zlo^0$ (back limb) latents are close to one another, while the $\zlo^1$ (front limb) latents are far apart.
To investigate these local agent behavioral differences, the right panel of \cref{fig:mujoco_examples_zomega_zalpha} provides an overlaid view of the same frames, with the torso now aligned (making it easier to discern the behaviors of agent 0 vs. agent 1).
Here we observe that the back limb (agent 0) behaves similarly across both examples, while the front limb (agent 1) stays much closer to the head for Example C, in contrast to Example B, which coincides with the findings in the $\zlo$ space.
\Cref{sec:additional_large_scale_experiments} provides additional results for a 4-agent MuJoCo AntWalker environment.

\paragraph{Application to externally-trained policies: hide-and-seek game.}
\begin{figure}[t]
    \centering
    \includegraphics[width=0.19\textwidth]{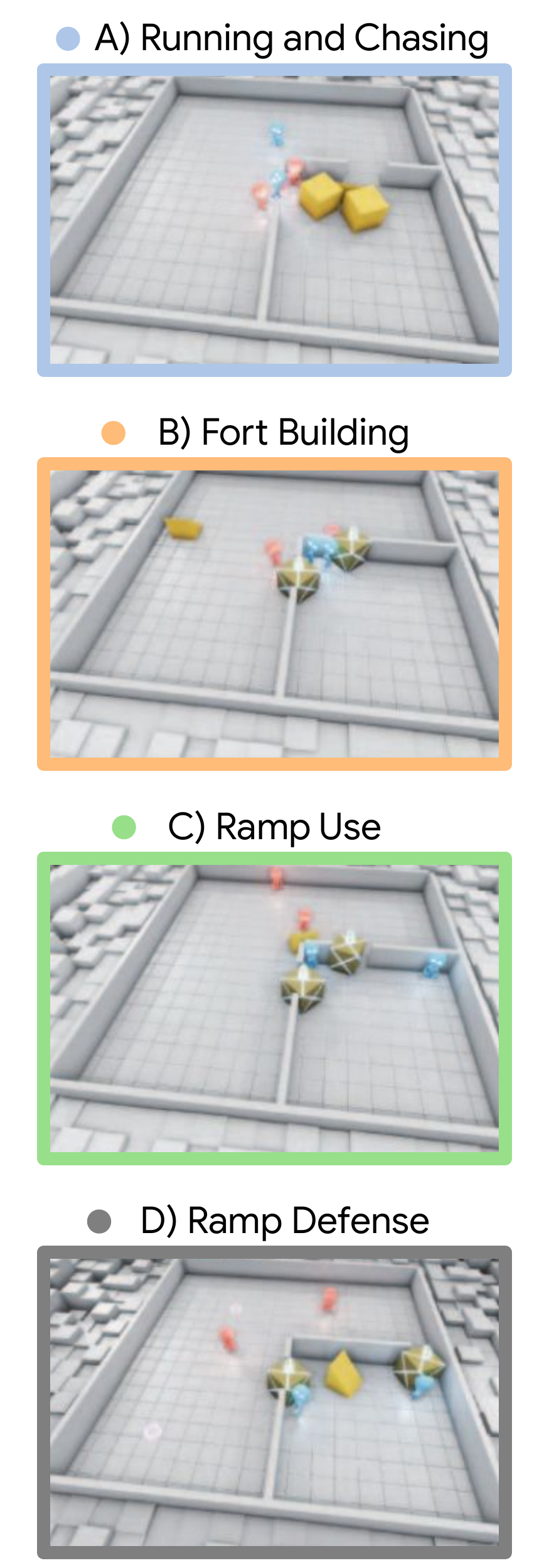}%
    \hspace{30pt}%
    \includegraphics[width=0.67\textwidth]{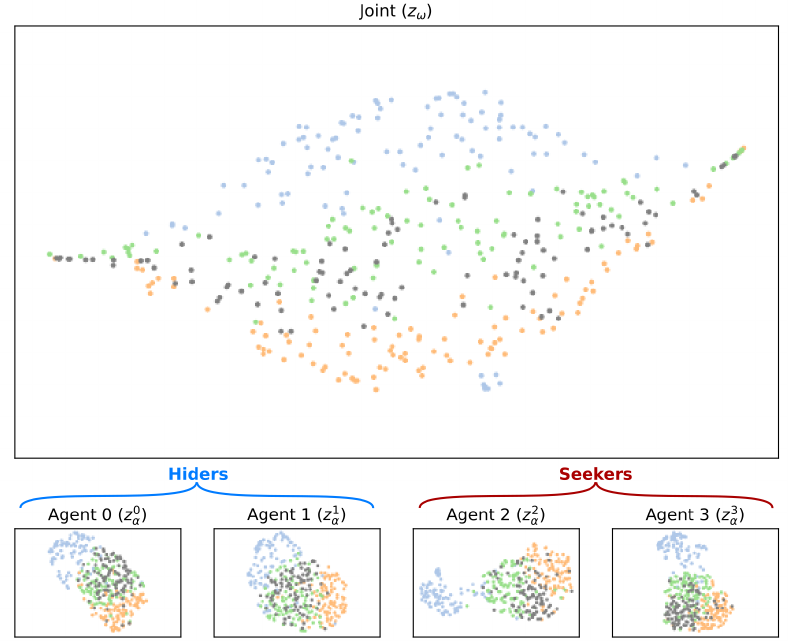}
    \caption{
        Results for the OpenAI hide-and-seek environment~\citep{Baker2020Emergent}, involving two teams of agents (2 hiders and 2 seekers). 
        We mix the trajectories collected from 4 OpenAI-annotated behavioral checkpoints (indicated on the left panel) together, then train \OurAlgAcronym on this shuffled dataset.
        Using \OurAlgAcronym, we observe the presence of behavior clusters that correspond well to human-annotated policy labels, both at the joint and local agent levels, despite our algorithm not having access to policy labels during training.
        Domain screenshots reproduced with permission from~\citet{Baker2020Emergent}.
    }
    \vspace{-12pt}
    \label{fig:hide_n_seek_results}
\end{figure}
We next apply \OurAlgAcronym to policies trained by external teams.
We consider the OpenAI hide-and-seek environment~\citep{Baker2020Emergent}, where 2 hiders and 2 seekers compete in a rich environment with various interactive objects (boxes and ramps).
OpenAI has open-sourced policies~\citep{OpenAIHideAndSeekGithub} annotated by humans as exhibiting distinctive behaviors at key stages of training.
We consider four policy checkpoints that correspond to the following human annotations:
A) `running and chasing', B) `fort building', C) `ramp use', and D) `ramp defense'.
The state-space used for each of the agents is 100-dimensional, consisting of the agent's own state (position, rotation, and velocity), states of the other 3 agents, and the states of 3 boxes and 1 ramp in the environment (position, velocity, and box size);
each agent's action space consists of a 3-dimensional force vector, and a `glue' and `lock' action for interacting with objects.
We collect $100$ trajectories per policy checkpoint, each being $200$ timesteps long.
These trajectories exhibit a wide distribution of behaviors, as agent and object initializations are random in each episode.
We then mix the trajectories collected from all policy checkpoints, then train \OurAlgAcronym on this shuffled dataset.

\Cref{fig:hide_n_seek_results} visualizes the behavior spaces discovered by \OurAlgAcronym in the hide-and-seek domain.
Agents 0 and 1 in this figure correspond to `hiders', whereas agents 2 and 3 are `seekers'.
We label each of the trajectories in this figure with the human-expert annotations provided by OpenAI.
In \cref{fig:hide_n_seek_results}, we observe the presence of behavior clusters that correspond well to the human-expert labels, both at the joint and local agent levels.
Interestingly, for the seekers (agents 2 and 3), policy A (`running and chasing') is highly distinctive and well-separated from the other behaviors. 
Moreover, despite the order of emergent behaviors in the original hide-and-seek MARL training being A $\to$ B $\to$ C $\to$ D, policies B (`fort building') and D (`ramp defense') appear to be behaviorally slightly closer to one another than C (‘ramp use’) and D, both in the joint space and also the local spaces of the seekers (agent 2 and 3).
This could perhaps be due to both the `fort building' and `ramp defense' policies being associated with situations where the seekers cannot easily find the hiders, due to the hiders using obstacles to block entrances (B) and moving ramps to prevent their effective use (D).
Overall, these experiments help to validate \OurAlgAcronym's learned latent spaces using policy labels manually annotated by human experts, and highlight its applicability to behaviorally-rich domains.

\section{Discussion}
Our proposed method, \OurAlgAcronym, leverages trajectory data to better understand multiagent behaviors during and after training. 
\OurAlgAcronym assumes no knowledge of agents' underlying training algorithms, does not require access to their hidden states or internal models, and applies even to reward-free settings.
Our experiments showcased a variety of applications of \OurAlgAcronym, including the analysis of joint and local agent behaviors, monitoring of behavior emergence throughout training, discovery of behavioral concepts associated with certain domain criteria, and disentanglement of third-party-labeled behaviors from open-source policies such as those for OpenAI hide-and-seek~\citep{Baker2020Emergent}. 

While we believe our approach is an important step in terms of increasing the understanding of multiagent systems, there are several limitations and potential societal impacts of note.
One limitation is related to the collection of a dataset of agent behaviors. 
Understanding emergent behaviors throughout MARL training using \OurAlgAcronym requires the storage of large-scale trajectory datasets, which could potentially take a lot of storage and compute power to generate. 
It would be interesting to consider follow-ups to our model that learn behavioral clusters in a streaming fashion, thus building knowledge of behaviors over time and permitting older data to be discarded throughout agent training. 
Moreover, in our datasets, we collected trajectories at uniform intervals throughout original MARL training. 
However, it might be interesting to consider a non-uniform collection scheme, e.g., collecting trajectories only when they are detected to diverge from behavioral clusters of a pre-trained \OurAlgAcronym model.
Finally, it might be interesting to enable the agents to first learn joint behaviors / `skills' using the proposed method, and leverage them to transfer to related downstream tasks (potentially more sample-efficiently than training from scratch).

The study of behavioral interactions in multiagent systems can potentially be used for both positive and negative societal applications.
For example, such an approach could be used to prevent certain undesirable behaviors by agents that interact with humans (e.g, self-driving cars), but potentially also used by adversaries to predict and exploit other behaviors (e.g., exploiting certain human preferences for harm),
or even inadvertently cause harm due to misinterpretation of certain behavioral modes.
As such, further research and evaluation will be required prior to deployment of this and related behavioral analysis approaches to human-facing domains.
Nonetheless, as agent capabilities continue to grow, our view is that behavioral analysis of multiagent systems will become increasingly important and should complement traditional reward-based performance monitoring.

\begin{ack}
We thank Asma Ghandeharioun, Meredith Morris, and Kathy Meier-Hellstern for their helpful feedback and support during the paper writing phase.
\end{ack}

\bibliographystyle{unsrtnat}
\bibliography{references}

\section*{Checklist}


\begin{enumerate}

\item For all authors...
\begin{enumerate}
  \item Do the main claims made in the abstract and introduction accurately reflect the paper's contributions and scope?
    \answerYes{}
  \item Did you describe the limitations of your work?
    \answerYes{We include these in the Discussions section.}
  \item Did you discuss any potential negative societal impacts of your work?
    \answerYes{We include these in the discussions section.}
  \item Have you read the ethics review guidelines and ensured that your paper conforms to them?
    \answerYes{}
\end{enumerate}

\item If you are including theoretical results...
\begin{enumerate}
  \item Did you state the full set of assumptions of all theoretical results?
    \answerYes{These are provided in the main paper and the proofs included in the appendix.}
  \item Did you include complete proofs of all theoretical results?
    \answerYes{Proofs are provided in the appendix.}
\end{enumerate}

\item If you ran experiments...
\begin{enumerate}
  \item Did you include the code, data, and instructions needed to reproduce the main experimental results (either in the supplemental material or as a URL)?
    \answerYes{We provide details for experiment reproducibility in \cref{sec:appendix_exp_details}. We also include model high-level code in \cref{sec:model_code}.}
  \item Did you specify all the training details (e.g., data splits, hyperparameters, how they were chosen)?
    \answerYes{We provide all training details in \cref{sec:appendix_exp_details}.}
    \item Did you report error bars (e.g., with respect to the random seed after running experiments multiple times)?
    \answerYes{}
    \item Did you include the total amount of compute and the type of resources used (e.g., type of GPUs, internal cluster, or cloud provider)?
    \answerYes{We provide all computational details in \cref{sec:appendix_exp_details}.}
\end{enumerate}

\item If you are using existing assets (e.g., code, data, models) or curating/releasing new assets...
\begin{enumerate}
  \item If your work uses existing assets, did you cite the creators?
    \answerNA{Our work does not use existing assets.}
  \item Did you mention the license of the assets?
    \answerNA{Our work does not use existing assets.}
  \item Did you include any new assets either in the supplemental material or as a URL?
    \answerNo{}
  \item Did you discuss whether and how consent was obtained from people whose data you're using/curating?
    \answerNA{Our work does not use existing assets.}
  \item Did you discuss whether the data you are using/curating contains personally identifiable information or offensive content?
    \answerNA{Our work does not use existing assets.}
\end{enumerate}

\item If you used crowdsourcing or conducted research with human subjects...
\begin{enumerate}
  \item Did you include the full text of instructions given to participants and screenshots, if applicable?
    \answerNA{Our work does not crowdsourced or human subject data.}
  \item Did you describe any potential participant risks, with links to Institutional Review Board (IRB) approvals, if applicable?
    \answerNA{Our work does not crowdsourced or human subject data.}
  \item Did you include the estimated hourly wage paid to participants and the total amount spent on participant compensation?
    \answerNA{Our work does not crowdsourced or human subject data.}
\end{enumerate}

\end{enumerate}

\newpage

\renewcommand{\thepage}{}

\appendix

\section{Appendix for `Beyond Rewards: a Hierarchical Perspective on Offline Multiagent Behavioral Analysis'}

\subsection{Derivations}\label{sec:appendix_elbo_derivation}
\subsubsection{On the Conditional Independence of $\pi^i$ from $\zhi$ given $\zlo^i$}\label{sec:appendix_conditional_indep}
In \cref{eq:policy_decomp_intermed}, we assumed that each agent's latent-conditioned policy is conditionally-independent of the high-level latent behavior $\zhi$ given its low-level latent $\zlo^i$.
This section provides justification of this assumption.

One of the prototypical paradigms in multiagent (MARL) training is that of ‘centralized training, decentralized execution’. 
In this regime, each agent’s decision-making policy is conditioned only on its available local information / local observations during execution; 
thus, the behaviors exhibited by each agent are ultimately informed by their local information, rather than global (joint) information. 

In our setting, $z_{\alpha}^i$ serves to provide this local behavior context for each agent. 
Thus, the assumption of the policy being conditionally-independent of $z_{\omega}$ given $z_{\alpha}^i$ corresponds well to the assumption of agents only using local information (rather than joint information) in MARL to inform their policy/decision-making.

Having said this, we also note that there is a strong relationship between local and joint behavioral distributions in our setting, which helps provide a coordination signal to the policy implicitly through $z_{\alpha}^i$. Specifically, our derivation results in a local behavior prior term, $p(z_{\alpha}^i | z_{\omega})$; using this term, the joint behavior latent $z_{\omega}$ is able to influence the learned space over local behavior latents $z_{\alpha}^i$. Thus, the information-flow in our framework can be more intuitively described as follows:
\be
    \item The joint behavior latent observes a trajectory and summarizes the joint behavior exhibited in it (e.g., team-wide cooperation, competition, etc.) in $z_{\omega}$.
    \item Subsequently, this joint latent $z_{\omega}$ affects the local behavior spaces $z_{\alpha}^i$, which are then sampled and used to inform each agent how to behave locally in order to achieve the joint behavior observed.
\ee

Overall, the above relationship between the two latent spaces implies makes the conditional-independence of the policy from $\zhi$ a reasonable simplifying assumption.

\subsubsection{Derivation of the Variational Lower Bound}

This section details the derivation of the variational lower bound described in \cref{sec:offline_analysis} of the main text, which is used for training \OurAlgAcronym.

Similar to traditional variational autoencoder approaches~\citep{kingma2014auto,rezende2014stochastic}, we approximate the maximization of the latent-conditioned trajectory probability \cref{eq:conditional_tau} in the main text using the evidence lower bound
\begin{align}
    J_{lb} &= \E_{\tau \sim \mathcal{D}, q_\phi(\zlo,\zhi|\tau)}\left[\log p^\pi(\tau|\zlo,\zhi)] - \E_{\tau \sim \mathcal{D}}[ \KL(q_\phi(\zlo,\zhi|\tau)||p_\theta(\zlo,\zhi))\right] \, ,
\end{align}
where $q_\phi(\cdot|\tau)$ and $p_\theta(\cdot)$ are, respectively, learned posterior and prior distributions.

Using \cref{eq:conditional_tau} from the main text to expand this expression yields
\begin{align}
    J_{lb} &= \E_{\tau \sim \mathcal{D}, q_\phi(\zlo,\zhi|\tau)}\left[\log p(s_0) 
    + \sum_t \log p(s_{t+1}|s_t,a_t) + \sum_{t,i} \log \pi^i_\theta(a_t^i|s_t,\zlo^i)\right] \nonumber\\
    &- \E_{\tau \sim \mathcal{D}}\left[ \KL(q_\phi(\zlo,\zhi|\tau)||p_\theta(\zlo,\zhi))\right]
\end{align}
Note that the terms $p(s_0)$ and $p(s_{t+1}|s_t,a_t)$ stem from the underlying MA-MDP environment, and thus cannot be optimized via parameters $\phi$ and $\theta$.
Dropping these extraneous terms yields,
\begin{align}
    J_{lb} &= \E_{\tau \sim \mathcal{D}, q_\phi(\zlo|\tau)}\left[\sum_{t,i} \log \pi^i_\theta(a_t^i|s_t,\zlo^i)\right] -\E_{\tau \sim \mathcal{D}}\left[ \KL(q_\phi(\zlo,\zhi|\tau)||p_\theta(\zlo,\zhi))\right] \, . \label{eq:loss_intermediate}
\end{align}
The first term simply relates to trajectory reconstruction, inducing our agent-wise policies $\pi_\theta^i$ to behave similarly to the observed trajectories in the dataset.
The second term is a regularization term involving our two latent parameters, which we further simplify using assumptions used in previous multi-level VAE models~\citep{shen2020towards}.
First, we assume that the posterior distribution is factorizable when conditioned on a particular trajectory, as follows,
\begin{align}
    q_\phi(\zlo,\zhi|\tau) &= q_\phi(\zlo|\tau)q_\phi(\zhi|\tau)\\
    &= \left[\prod_i q_\phi(\zlo^i|\tau)\right]q_\phi(\zhi|\tau) \, , \label{eq:factorize_posterior}
\end{align}
and similarly for the prior,
\begin{align}
    p_\theta(\zlo,\zhi) &= p_\theta(\zlo|\zhi)p_\theta(\zhi)\\
    &=\left[\prod_i p_\theta(\zlo^i|\zhi)\right]p_\theta(\zhi) \,. \label{eq:factorize_prior}
\end{align}

We next simplify the KL-divergence component of \cref{eq:loss_intermediate} as follows by combining it with \cref{eq:factorize_posterior,eq:factorize_prior}: 
\begin{align}
     \KL(q_\phi&(\zlo,\zhi|\tau)||p_\theta(\zlo,\zhi)) \\
    &= \int_{\zlo,\zhi} q_\phi(\zhi,\zlo|\tau) \log \frac{q_\phi(\zlo,\zhi|\tau)}{p_\theta(\zlo,\zhi)}d_{\zlo}d_{\zhi} \\
    &= \int_{\zlo,\zhi} q_\phi(\zlo|\tau) q_\phi(\zhi|\tau) \log \frac{q_\phi(\zlo|\tau)q_\phi(\zhi|\tau)}{p_\theta(\zlo|\zhi)p_\theta(\zhi)} d_{\zlo}d_{\zhi}\\
    &= \int_{\zlo,\zhi} \left[q_\phi(\zlo|\tau) q_\phi(\zhi|\tau) \log \frac{q_\phi(\zlo|\tau)}{p_\theta(\zlo|\zhi)}  + q_\phi(\zlo|\tau) q_\phi(\zhi|\tau) \log \frac{q_\phi(\zhi|\tau)}{p_\theta(\zhi)} \right] d_{\zlo}d_{\zhi}\\
    &= \int_{\zlo,\zhi} q_\phi(\zhi|\tau) \prod_j q_\phi(\zlo^j|\tau)  \sum_i \log \frac{q_\phi(\zlo^i|\tau)}{ p_\theta(\zlo^i|\zhi)}d_{\zlo}d_{\zhi}  \nonumber\\
        & \hspace{20pt} + \int_{\zhi}  q_\phi(\zhi|\tau) \log \frac{q_\phi(\zhi|\tau)}{p_\theta(\zhi)} d_{\zhi} \underbrace{\int_{\zlo}q_\phi(\zlo|\tau)d_{\zlo}}_{=1}\\
    &= \sum_i \int_{\zlo,\zhi} q_\phi(\zhi|\tau) \left(\prod_j q_\phi(\zlo^j|\tau)\right)   \log \frac{q_\phi(\zlo^i|\tau)}{ p_\theta(\zlo^i|\zhi)}d_{\zlo}d_{\zhi}  + \KL(q_\phi(\zhi|\tau)||p_\theta(\zhi))\\
    &= \sum_i \int_{\zlo^i,\zhi} q_\phi(\zhi|\tau)  q_\phi(\zlo^i|\tau)   \log \frac{q_\phi(\zlo^i|\tau)}{ p_\theta(\zlo^i|\zhi)}d_{\zlo^i}d_{\zhi} \left[\prod_{j \neq i} \underbrace{\int_{\zlo^{j}}q_\phi(\zlo^j|\tau) d_{\zlo^j}}_{=1}\right] \nonumber\\
    & \hspace{20pt} + \KL(q_\phi(\zhi|\tau)||p_\theta(\zhi))\\
    &= \E_{q_\phi(\zhi|\tau)} \left[\sum_i \KL(q_\phi(\zlo^i|\tau) || p_\theta(\zlo^i|\zhi))\right]  + \KL(q_\phi(\zhi|\tau)||p_\theta(\zhi))
\end{align}
Combining this result with \cref{eq:loss_intermediate} and additionally modulating the KL-terms in the lower bound as in $\beta$-VAEs~\citep{higgins2017beta}, our overall objective simplifies to
\begin{align}
    J_{lb} = &\E_{\tau \sim \mathcal{D}, \zlo \sim q_\phi(\zlo|\tau)}\left[\sum_{t,i} \log \pi^i_\theta(a_t^i|s_t,\zlo^i)\right]\\
    - \beta \Bigg[&\E_{\tau \sim \mathcal{D}, \zhi \sim q_\phi(\zhi|\tau)} \Big[\sum_i \KL(q_\phi(\zlo^i|\tau) || p_\theta(\zlo^i|\zhi))\Big]\\
    +  &\E_{\tau \sim \mathcal{D}}\left[ \KL(q_\phi(\zhi|\tau)||p_\theta(\zhi))\right] \Bigg] \,. 
\end{align}

\subsection{Experiment Details and Hyperparameters}\label{sec:appendix_exp_details}

This section provides an overview of the details for our experiments.

\subsubsection{Data Generation Details}

The multiagent trajectory data analyzed in this paper is generated using the Acme RL library~\citep{hoffman2020acme}, using the TD3 algorithm~\citep{fujimoto2018addressing} in a decentralized fashion, with dataset management handled using RLDS~\citep{ramos2021rlds}.
Each agent uses a 2-layer MLP (256 hidden units for each layer) for its TD3 network (using the vanilla TD3 network specifications in Acme), with the hyperparameters used for data generation summarized for each domain in \cref{table:appendix_data_generation_hparams}.
Note that the large number of training trials/seeds used in this data generation pipeline serve to produce a wide variety of agent behaviors in the dataset subsequently analyzed by \OurAlgAcronym.

\begin{table}[h]
    \centering
    \caption{Hyperparameters used for data generation. Values in braces indicate hyperparameter sweeps.}
    \label{table:appendix_data_generation_hparams}
    \begin{tabular}{lcccc}
    \toprule
    Parameter                            & Hill-climbing & Coordination Game & HalfCheetah & AntWalker \\
    \midrule
    MARL algorithms used         & TD3                        & TD3                & TD3 & \{TD3, SAC\}                  \\
    \# agents               & 3                        & 2                & 2   &   4           \\
    Episode steps               & 50                        & 50                & 200   &   300           \\
    \# steps per training trial & 1e5                       & 2e5               & 2e5   &   2e5          \\
    TD3 / SAC batch size              & 256                       & 256               & 256   &  \{32, 256\}           \\
    TD3 / SAC learning rates    & 5e-5                      & 5e-5              & 5e-4  &  \{1e-3 to 9e-3\}          \\
    TD3 $\sigma$                   & 0.1                       & 0.1               & 0.1 &   0.1            \\
    TD3 target $\sigma$            & 0.1                       & 0.1               & 0.1 &     0.1          \\
    TD3 $\tau$                     & 0.005                     & 0.005             & 0.005 &   0.005          \\
    TD3 delay                   & 2                         & 2                 & 2        & 2          \\
    \# seeds                    & 50                        & 50                & 30       & 50         \\
    \bottomrule
    \end{tabular}
\end{table}

\subsubsection{\OurAlgAcronym Hyperparameters}

We conduct a wide hyperparameter sweep for training \OurAlgAcronym itself, summarized in \cref{table:appendix_ouralg_hparams}.
Our joint and local encoder models each consist of a bidirectional LSTM, followed by a 2-layer MLP head for mapping to latent distribution parameters; 
a GMM head is used for the joint encoder, and a Gaussian for the local encoder.
Our local prior model consists of a 2-layer MLP with Gaussian head, with our joint prior model simply being the learned parameters of a GMM.
All models use ReLU for intermediate layer activations.
Note that we found that cyclically-annealing~\citep{fu_cyclical_2019_naacl} the $\beta$ term in our variational lower bound from $0$ to the values specified in \cref{table:appendix_ouralg_hparams} to help avoid KL-vanishing.
Sweeps are conducted over the \OurAlgAcronym latent space dimensionality ($4$ and $8$-dimensional), MLP hidden sizes ($64$ and $128$ units), LSTM hidden sizes ($64$ and $128$ units), Adam optimizer~\citep{kingma2014adam} learning rates ($0.001$ and $0.0001$), with 3 seeds per parameter set (standard deviations are reported over all seeds).
All \OurAlgAcronym and baseline training is conducted using the Adam optimizer~\citep{kingma2014adam}, with Adam parameters $\beta_1=0.9$, $\beta_2=0.999$, gradient clipping using a max global norm threshold of $10.0$, and learning rates swept over as indicated in the tables.

\begin{table}[h]
    \centering
    \caption{Hyperparameter sweeps for training \OurAlgAcronym. Values swept over are indicated via braces.}
    \label{table:appendix_ouralg_hparams}
    \begin{tabular}{lc}
    \toprule
    Parameter                                                   & Value          \\
    \midrule
    Training steps                                              & 1e5            \\
    Batch size (\# trajectories)                                & 128            \\
    Latent $\zhi$ and $\zlo^i$ dimensionality                                       & \{4, 8\}       \\
    Hidden units (joint encoder/local prior/local encoder MLPs) & \{64, 128\}    \\
    Hidden units (joint encoder /local encoder LSTMs)           & \{64, 128\}    \\
    Hidden units (reconstructed policy MLP)                     & 32             \\
    GMM mixture size (joint prior/joint encoder)                & 8              \\
    Adam optimizer learning rates                               & \{1e-3, 1e-4\} \\
    KL-loss $\beta$ weighing term                                             & \{1e-4, 1e-2\} \\
    KL-loss cyclical annealing period                           & \{5e3, 1e4\}   \\
    \# seeds                                & 3              \\
    \bottomrule
    \end{tabular}
\end{table}

\subsubsection{Baseline Model Hyperparameters}

We similarly conduct a wide sweep over hyperparameters for the considered baselines, summarized in \cref{table:appendix_baseline_hparams}.

\begin{table}[h]
    \centering
    \caption{Hyperparameter sweeps for baselines. Values swept over are indicated via braces.}
    \label{table:appendix_baseline_hparams}
    \begin{tabular}{lcc}
    \toprule
    Parameter                                                   & LSTM Baseline & VAE Baseline          \\
    \midrule
    Training steps                                              & 1e5   & 1e5         \\
    Batch size (\# trajectories)                                & 128   & 128            \\
    Latent $\zhi$ dimensionality                                       & N/A   & \{4, 8\}       \\
    Hidden units (joint encoder/local prior/local encoder MLPs) & N/A   & \{64, 128\}    \\
    Hidden units (joint encoder /local encoder LSTMs)           & \{64, 128\} & \{64, 128\}   \\
    Hidden units (reconstructed policy MLP)                     & 32 & 32             \\
    GMM mixture size (joint prior/joint encoder)                & N/A & 8 \\
    Adam optimizer learning rates                               & \{1e-3, 1e-4\} & \{1e-3, 1e-4\}\\
    KL-loss $\beta$                                             & N/A & \{1e-4, 1e-2\} \\
    KL-loss cyclical annealing period                           & N/A & \{5e3, 1e4\}  \\
    \# seeds                                & 3      & 3        \\
    \bottomrule
    \end{tabular}
\end{table}

\subsubsection{Computational Details}
For MARL trajectory data generation, we used an internal CPU cluster for both the 3-agent hill-climbing and 2-agent coordination domains, using TPUs for only the multiagent MuJoCo data generation.
For training \OurAlgAcronym itself, we used TPUs for the 3-agent hill-climbing and 2-agent coordination domains, and a Tesla V100 GPU cluster for the MuJoCo environment. 
For training the LSTM and VAE baselines, we used Tesla P100 GPU clusters.

\subsubsection{Concept Discovery Framework}\label{sec:appendix_concept_discovery_details}
This section describes details of how we apply the completeness-aware concept-based explanation framework of \citet{yeh2020completeness} to discover interesting concepts in our setting.

Given a characteristic of interest (e.g., the level of dispersion of agents), we define a training set consisting of joint latents $\zhi$ and class labels $y$ (e.g., classes corresponding to different intervals of team returns).
\citet{yeh2020completeness} seek to identify a set of concept vectors that are sufficient for predicting the labels given the inputs.
Specifically, they let $C = \{c_j\}_{j=1}^{m}$ denote the set of concepts, which are unique vectors where $c_j \in \R^{D_\omega} \, \forall j$.
Normalizing $\zhi$, we can then use the inner product $\langle \zhi, c_j\rangle$ as a similarity measure between $\zhi$ and concept $c_j$.
The \emph{concept product} is defined $\nu_\vc(\zhi) = \text{TH}(\langle \zhi, c_j\rangle, \kappa) \in \R^m$, where $\text{TH}(\cdot, \kappa)$ clips values less than $\kappa$ to $0$.
Normalizing the concept product yields the \emph{concept score} $\hat{\nu}_{\vc}(\zhi) = \nicefrac{\nu_{\vc}(\zhi)}{||\nu_{\vc}(\zhi)||_2} \in \R^m$, where elements provide a measure of similarity of the input $\zhi$ to each of the $m$ concept vectors.

Using these definitions, we can gauge the representational power of $\zhi$ by learning a mapping $g: \hat{\nu}_{\vc}(\zhi) \to y$.
In practice, $g$ is a simple model (e.g., shallow network or linear projection) so as to gauge the expressivity of the latent space.
Given a mapping $g$, we use the classification accuracy as a `completeness score'~\citep{yeh2020completeness} for the set of concepts $C$, defined $\eta = \sup_g P_{\zhi,y \sim V} \left[y = \argmax_{y'}g(\nu_\vc(\zhi)) \right]$, where $V$ is a validation set.
Importantly, this approach permits us to compute class-conditioned Shapley values, called ConceptSHAP in the framework of \citep{yeh2020completeness}.
Specifically, for a given class $k$, the class-conditioned ConceptSHAP value for each concept $c_j$ is defined, 
\begin{align}
   \lambda_j(\eta_k) = \sum_{S \subseteq C \setminus c_j} = \frac{(m - |S| - 1)!|S|!}{m!}\left[\eta_k(S \cup \{c_j\}) - \eta_k(S) \right] \, , \label{eq:class_concept_shap}
\end{align}
where $\eta_k$ is the completeness score for class $k$ (computed simply as the classification accuracy of the model for the subset of validation points with ground truth label $k$).
ConceptSHAP provides a measure of importance of each concept $c_j$ for predicting the outcomes associated with a given class, which we can then use to identify the trajectories.
The experiments in our main paper use the above setup to both quantitatively evaluate $\zhi$ in terms of representation power, and also reveal interesting concepts associated with relevant characteristics in the domain.

For each of the prediction problems considered in the main paper (agent dispersion and agent returns), we create an 80-20 training-validation data split (over 5 classes). 
For concept generation we use K-means with 16 and 24 clusters, a 2-layer MLP (with 8 hidden units per layer) for the prediction head $g$, training with a batch size of $64$, and a concept threshold $\kappa$ of 0.0 and 0.3 for the dispersion and agent return experiments, respectively, and train the prediction head for $1e4$ steps.
As in~\citet{yeh2020completeness}, we use KernelSHAP~\citep{lundberg2017unified} to approximate ConceptSHAP efficiently.

\subsection{Additional Results}\label{sec:appendix_additional_results}

\subsubsection{Additional Large-scale Experiments}\label{sec:additional_large_scale_experiments}

\paragraph{MultiAgent MuJoCo AntWalker Domain Results.}
\begin{figure}[t]
    \centering
    \begin{subfigure}{\linewidth}
        \centering
        \includegraphics[width=0.3\textwidth]{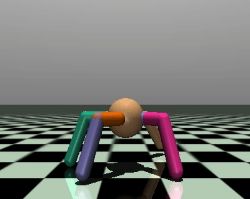}
        \caption{}
        \label{fig:antwalker_domain}
    \end{subfigure}\\
    \begin{subfigure}{\linewidth}
        \centering
        \includegraphics[width=\textwidth]{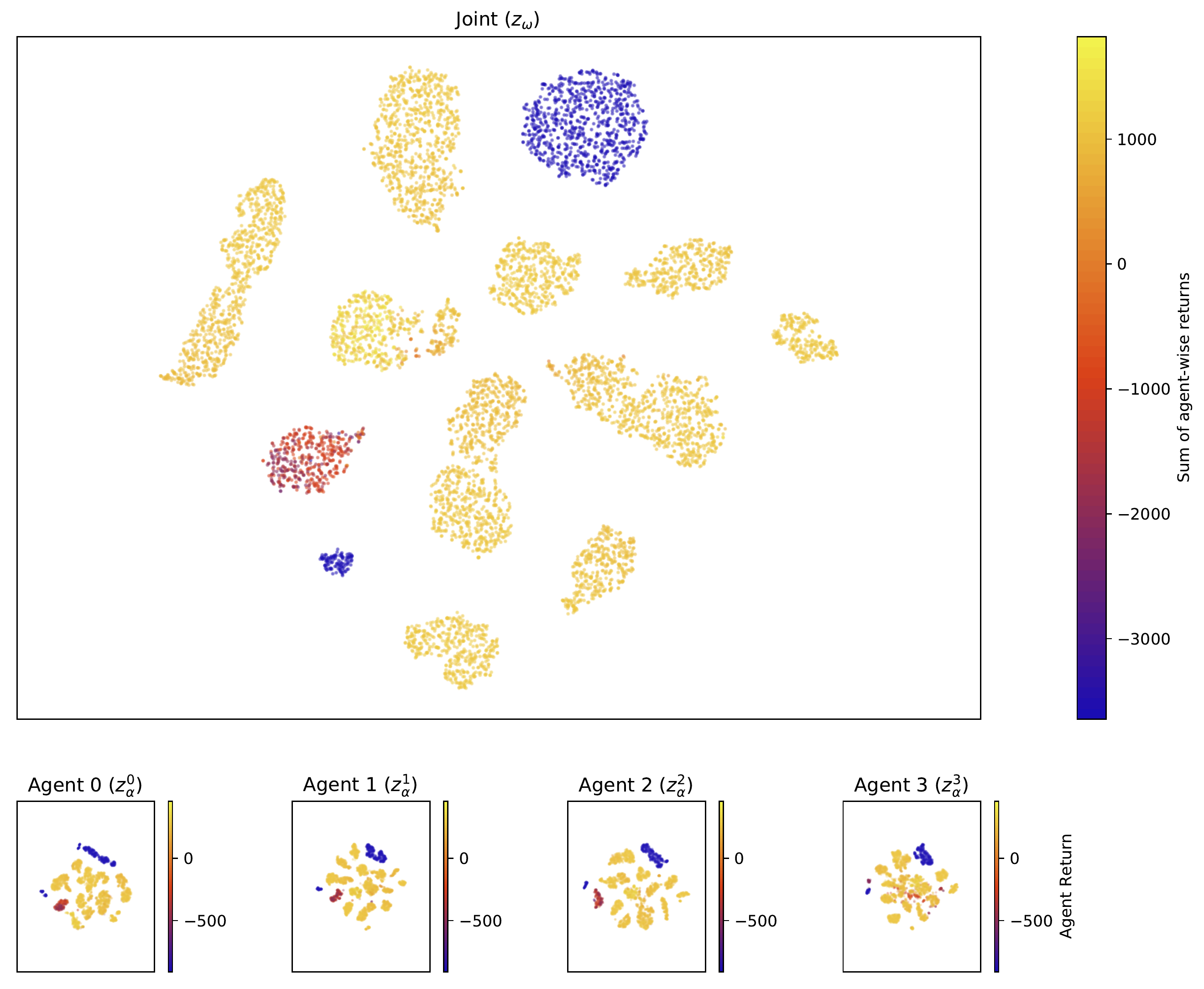}
        \caption{}
        \label{fig:mujoco_antwalker_env_latent_rewards_main_sec}
    \end{subfigure}%
    \caption{Results for Multiagent MuJoCo AntWalker domain. (see interactive version {\color{blue}\href{ https://storage.googleapis.com/mohba-beyond-rewards-22/n22/multiagent_ant_walker/interactive_final.html}{here}}).
    \subref{fig:antwalker_domain} This domain involves 4 agents, each controlling one of the ant legs to coordinate movement towards the $+x$ direction.
    \subref{fig:mujoco_antwalker_env_latent_rewards_main_sec} Behavior space learned by \OurAlgAcronym in the AntWalker domain.
    }
\end{figure}
We also consider the MultiAgent MuJoCo AntWalker domain, wherein 4 agents each control one of 4 ant legs to coordinate movement towards the $+x$ direction (\cref{fig:antwalker_domain}).
To collect data for this domain, we conduct a wider MARL parameter sweep using both the TD3 and SAC algorithms, with varying training batch sizes and learning rates to gather a widely-varying dataset of behaviors.
We subsequently train \OurAlgAcronym on this data, visualizing the learned behavior clusters in \cref{fig:mujoco_antwalker_env_latent_rewards_main_sec}.
We observe several behavior clusters of interest;
notably, a large joint cluster exists in the top-right region of the joint and local returns, which, upon inspection of underlying trajectory videos, corresponds to cases where agents attain very low return.
Similarly, there exists a smaller cluster in the lower-left region of the joint returns that also attains very low performance.
The remaining clusters correspond to the agents displaying various ant-poses, and moving only incrementally.
One exceptional cluster also exists in the left region of the joint behavior space, which attains medium-level return (points that are primarily red in color).
On closer inspection, the AntWalker behavior in this cluster corresponds to one of the agents learning a reasonably good walking gait, while the remaining three agents remain stationary.

\subsubsection{Additional Baseline Results}\label{appendix:additional_baseline_experiments}

\begin{figure}[h]
    \centering
    \begin{subtable}[b]{0.7\textwidth}
        \centering
        \begin{tabular}{@{\extracolsep{4pt}}ccccc}
            \toprule
            & APL & \multicolumn{3}{c}{ICTD} \\
            \cline{2-2}  \cline{3-5} 
            & & K=4 & K=8 & K=16 \\
             \cline{3-3} \cline{4-4} \cline{5-5}
             LSTM & $4.8 \pm 0.1$ & $0.60 \pm 0.15$ & $0.51 \pm 0.17$ & $0.47 \pm 0.19$ \\
             VAE & $6.9 \pm 0.4$ & $0.44 \pm 0.11$ & $0.30 \pm 0.14$ & $0.18 \pm 0.13$ \\
             \OurAlgAcronym & $1.3 \pm 0.1$ & $0.46 \pm 0.13$ & $0.31 \pm 0.13$ & $0.17 \pm 0.13$ \\
            \bottomrule
        \end{tabular}
        \caption{Action-prediction loss (APL) and intra-cluster trajectory distance (ICTD).}
    \end{subtable}\\
    \begin{subfigure}[b]{0.4\textwidth}
        \includegraphics[width=\textwidth]{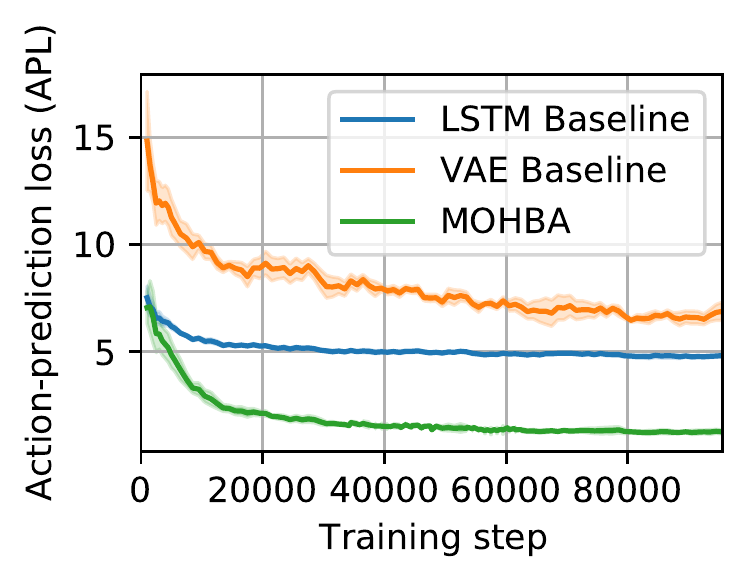}
        \caption{APL convergence throughout training.}
    \end{subfigure}\\
    \begin{subfigure}[b]{0.45\textwidth}
        \centering
        \includegraphics[width=0.6\textwidth]{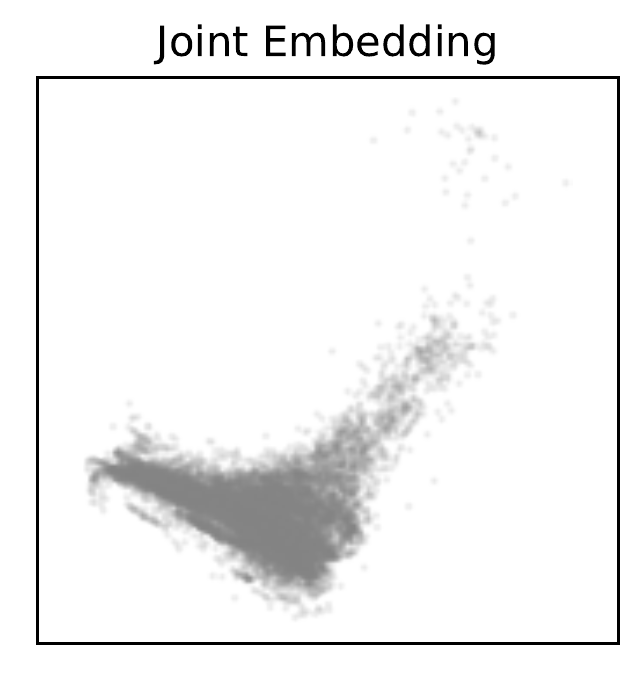}
        \caption{LSTM Baseline (PCA projection)}
    \end{subfigure}%
    \hspace{10pt}
    \begin{subfigure}[b]{0.45\textwidth}
        \centering
        \includegraphics[width=0.6\textwidth]{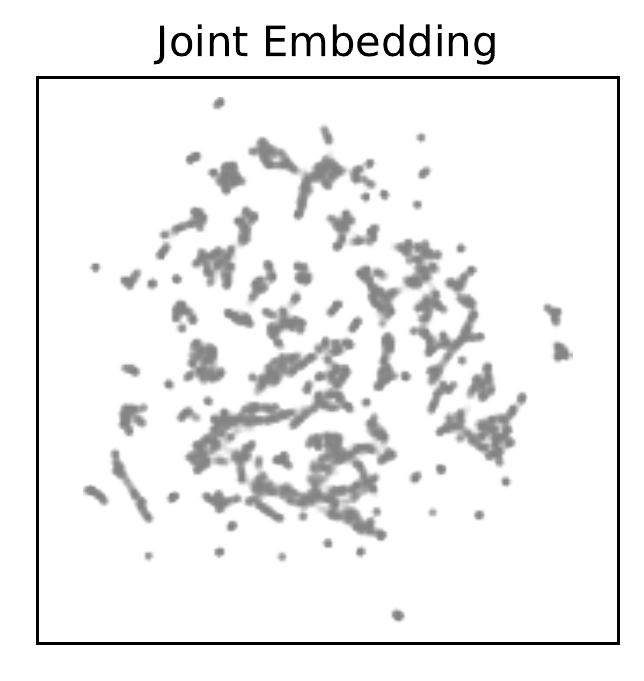}
        \caption{LSTM Baseline (UMAP projection)}
    \end{subfigure}\\
    \begin{subfigure}[b]{0.45\textwidth}
        \centering
        \includegraphics[width=0.6\textwidth]{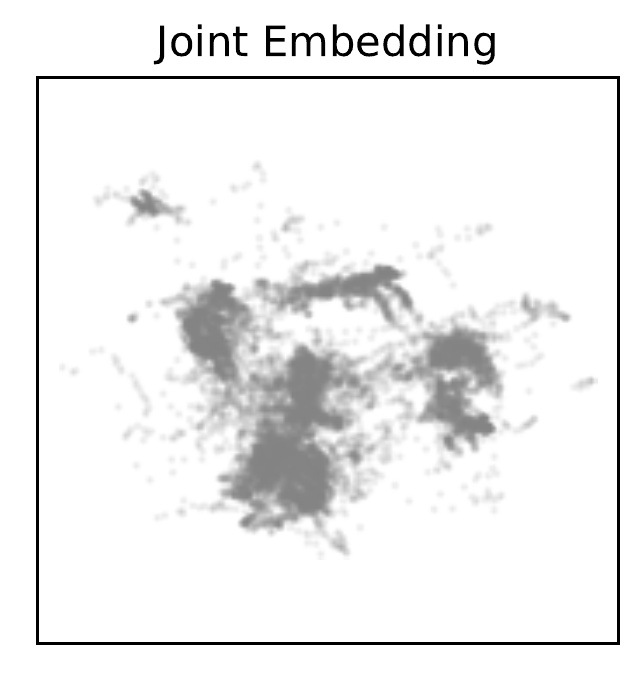}
        \caption{VAE Baseline (PCA projection)}
    \end{subfigure}%
    \hspace{10pt}
    \begin{subfigure}[b]{0.45\textwidth}
        \centering
        \includegraphics[width=0.6\textwidth]{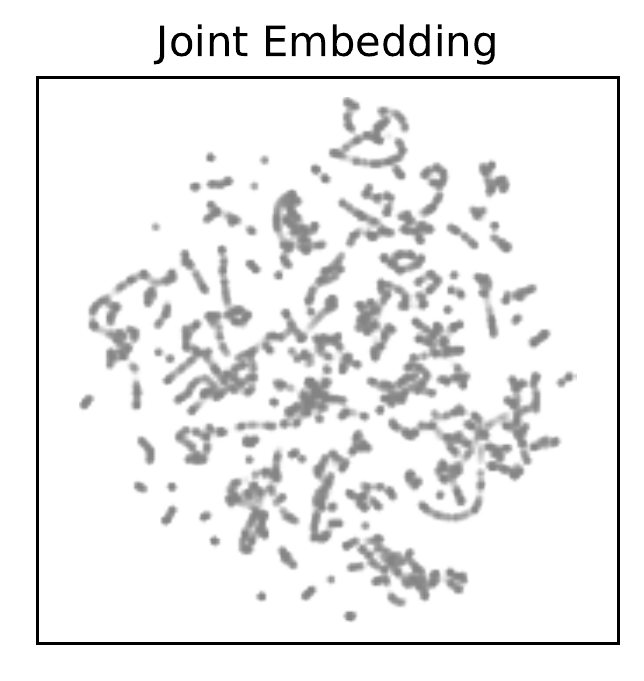}
        \caption{VAE Baseline (UMAP projection)}
    \end{subfigure}%
    \caption{Baseline comparisons (hill-climbing domain).}
    \label{fig:appendix_baseline_comparisons_multimodal}
\end{figure}

\begin{figure}[h]
    \centering
    \begin{subtable}[b]{0.7\textwidth}
        \centering
        \begin{tabular}{@{\extracolsep{2pt}}ccccc}
            \toprule
            & APL & \multicolumn{3}{c}{ICTD} \\
            \cline{2-2}  \cline{3-5} 
            & & K=4 & K=8 & K=16 \\
             \cline{3-3} \cline{4-4} \cline{5-5}
             LSTM & $2.9 \pm 0.1$ & $0.41 \pm 0.15$ & $0.40 \pm 0.14$ & $0.34 \pm 0.15$\\
             VAE & $5.0 \pm 0.3$ & $0.18 \pm 0.11$ & $0.17 \pm 0.11$ & $0.16 \pm 0.10$\\
             MOHBA & $ 2.4 \pm 0.3$ & $0.18 \pm 0.12$ & $0.17 \pm 0.10$ & $0.16 \pm 0.10$\\
            \bottomrule
        \end{tabular}
        \caption{Action-prediction loss (APL) and intra-cluster trajectory distance (ICTD).}
    \end{subtable}\\
    \begin{subfigure}[b]{0.4\textwidth}
        \centering
        \includegraphics[width=\textwidth]{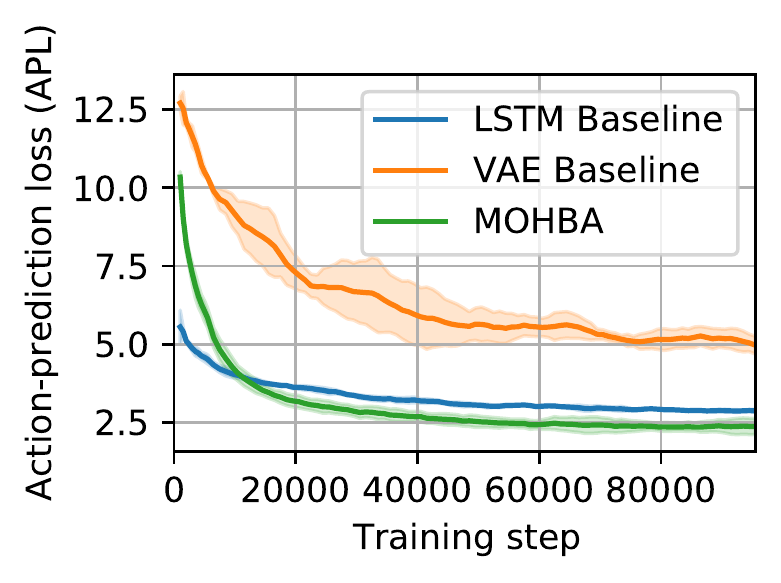}
        \caption{APL convergence throughout training.}
    \end{subfigure}\\
    \begin{subfigure}[b]{0.45\textwidth}
        \centering
        \includegraphics[width=0.6\textwidth]{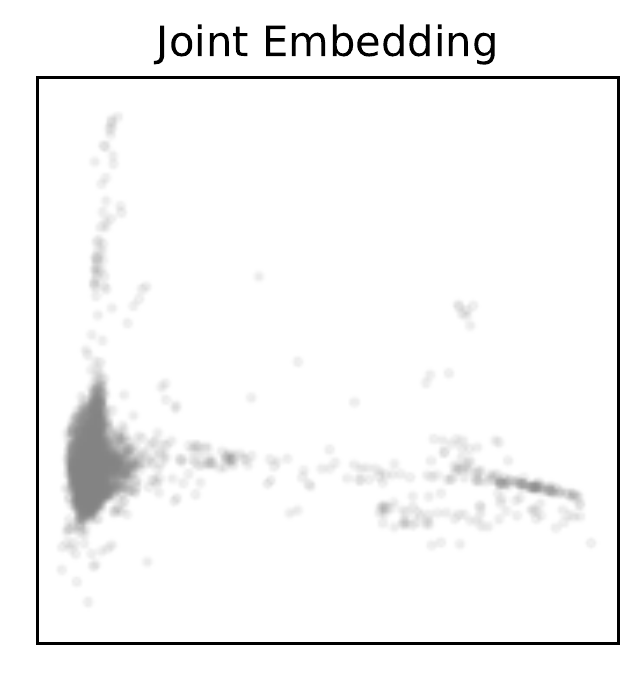}
        \caption{LSTM Baseline (PCA projection)}
        \label{fig:lstm_baseline_left_right_latent_space_pca}
    \end{subfigure}%
    \hspace{10pt}
    \begin{subfigure}[b]{0.45\textwidth}
        \centering
        \includegraphics[width=0.6\textwidth]{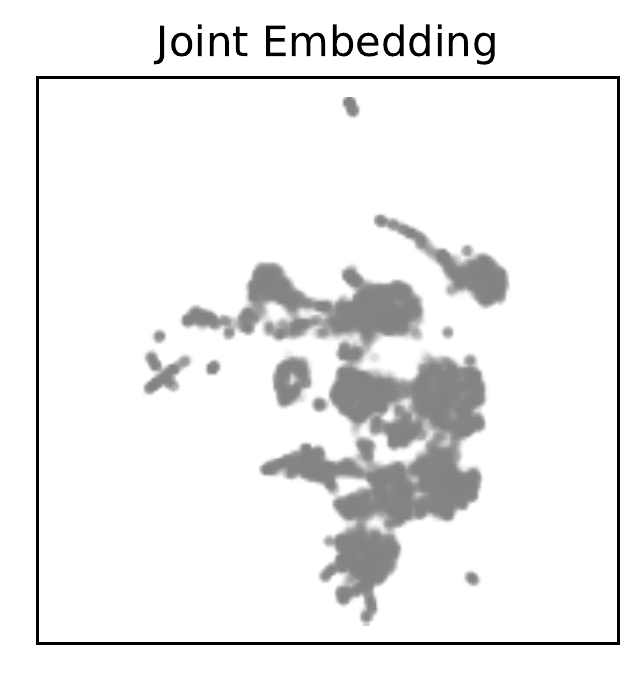}
        \caption{LSTM Baseline (UMAP projection)}
        \label{fig:lstm_baseline_left_right_latent_space_umap}
    \end{subfigure}\\
    \begin{subfigure}[b]{0.45\textwidth}
        \centering
        \includegraphics[width=0.6\textwidth]{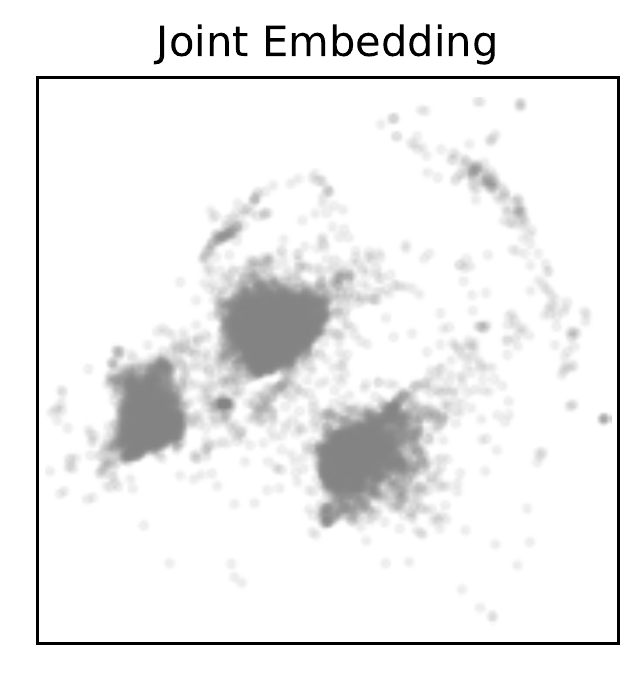}
        \caption{VAE Baseline (PCA projection)}
        \label{fig:vae_baseline_left_right_latent_space_pca}
    \end{subfigure}%
    \hspace{10pt}
    \begin{subfigure}[b]{0.45\textwidth}
        \centering
        \includegraphics[width=0.6\textwidth]{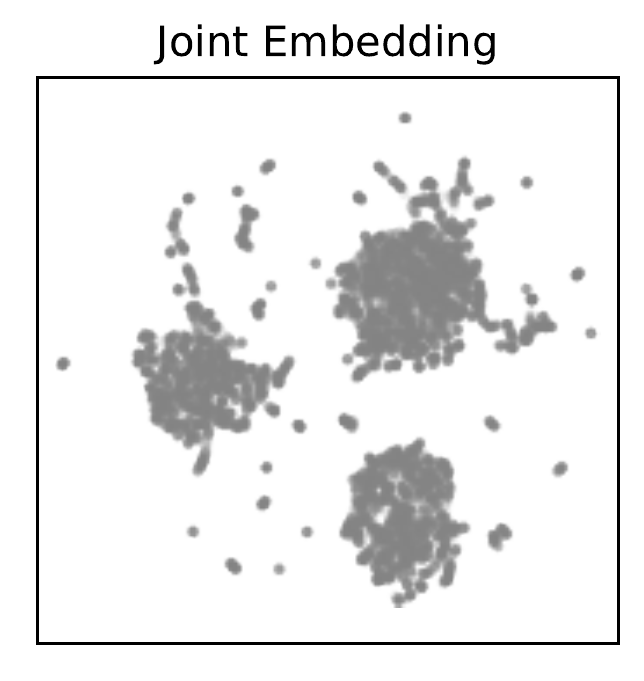}
        \caption{VAE Baseline (UMAP projection)}
        \label{fig:vae_baseline_left_right_latent_space_umap}
    \end{subfigure}%
    \caption{Baseline comparisons (coordination game).}
    \label{fig:appendix_baseline_comparisons_left_right}
\end{figure}

\Cref{fig:appendix_baseline_comparisons_multimodal,fig:appendix_baseline_comparisons_left_right} provide additional results for the baseline comparisons conducted in the main paper.
Specifically, (a) in each figure provides an expanded comparison of ICTD (with a sweep over the \# of clusters in K-means).
In (b) of each figure, we visualize the convergence of the APL throughout training for \OurAlgAcronym and the baselines.
Subfigures (c) and (d), respectively, show a PCA and UMAP projection of the latent space discovered by the LSTM baseline (recall from the main paper that `joint embeddings' used for the LSTM are simply the final hidden state).
Likewise, subfigures (e) and (f) show the same projections for the VAE baseline.

As especially evident in the coordination domain (\cref{fig:appendix_baseline_comparisons_left_right}), the joint LSTM baseline embeddings (whether using PCA or UMAP projections, \cref{fig:lstm_baseline_left_right_latent_space_pca,fig:lstm_baseline_left_right_latent_space_umap}, respectively) do not reveal to the fairly interpretable 3 clusters of coordinated behaviors discussed in the main paper.
By contrast, the VAE baseline (in both the PCA and UMAP case, \cref{fig:vae_baseline_left_right_latent_space_pca,fig:vae_baseline_left_right_latent_space_umap}, respectively) does produce these clusters;
despite this, note that the key limitation of this baseline is its non-hierarchical nature, which makes identification of the local-to-joint behavioral correspondences noted in the main paper significantly more difficult. 

\subsection{Hyperparameter Ablations and Effects on Latent Spaces}\label{sec:appendix_ablations}
This section conducts an analysis of the effect of key hyperparameters on the latent spaces learned by \OurAlgAcronym.
We focus our analysis on the KL-loss $\beta$ term in the training objective, alongside the latent $\zhi$ and $\zlo$ dimensionality.
We both compare the policy reconstruction loss \cref{eq:loss_reconstruction} across ablations to attain a quantitative comparison of the quality of the latent encodings, and additionally compare the structure of the learned latent spaces themselves as a function of these parameters for a qualitative understanding.

\begin{figure}[h]
    \centering
    \begin{subfigure}[b]{\textwidth}
        \centering
        \includegraphics[width=0.6\textwidth]{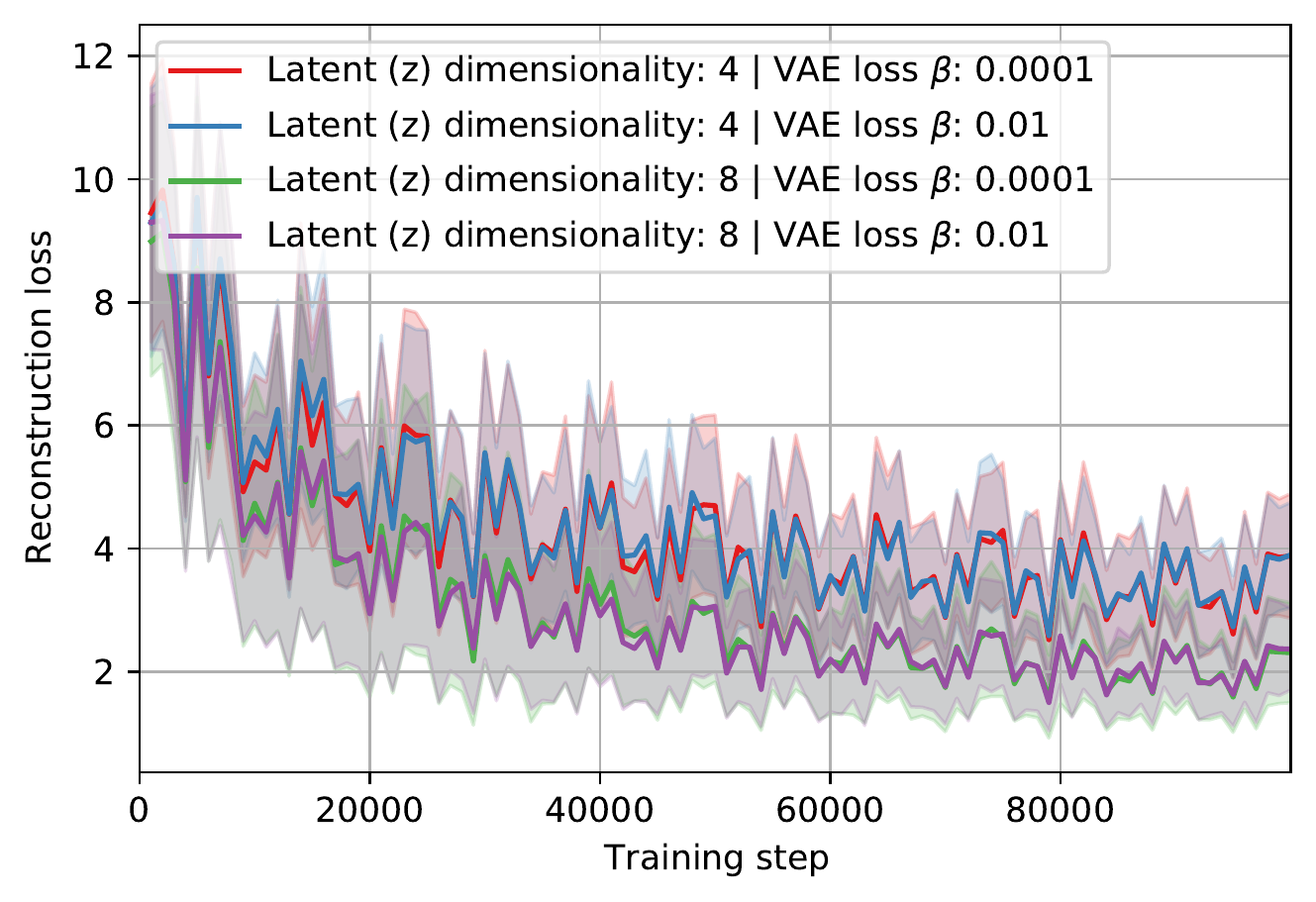}
        \caption{}
        \label{fig:ablative_loss_multimodal_reward_env}
    \end{subfigure}\\
    \begin{subfigure}[b]{0.48\textwidth}
        \centering
        \includegraphics[width=\textwidth]{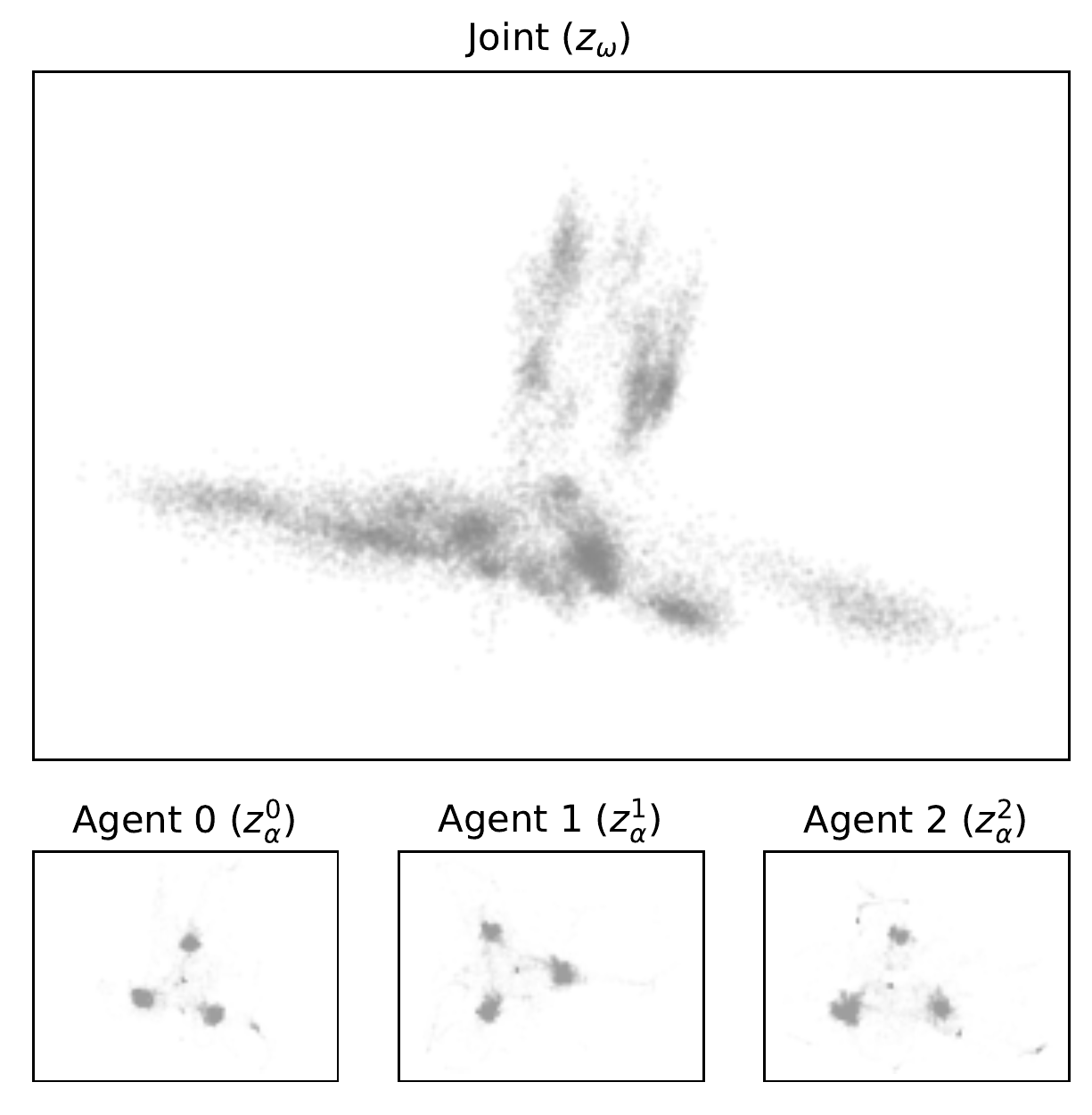}
        \caption{Latent $z$ dimensionality: 4, VAE KL-loss $\beta$: 1e-4}
        \label{fig:ablative_loss_multimodal_reward_env_a}
    \end{subfigure}%
    \hfill
    \begin{subfigure}[b]{0.48\textwidth}
        \centering
        \includegraphics[width=\textwidth]{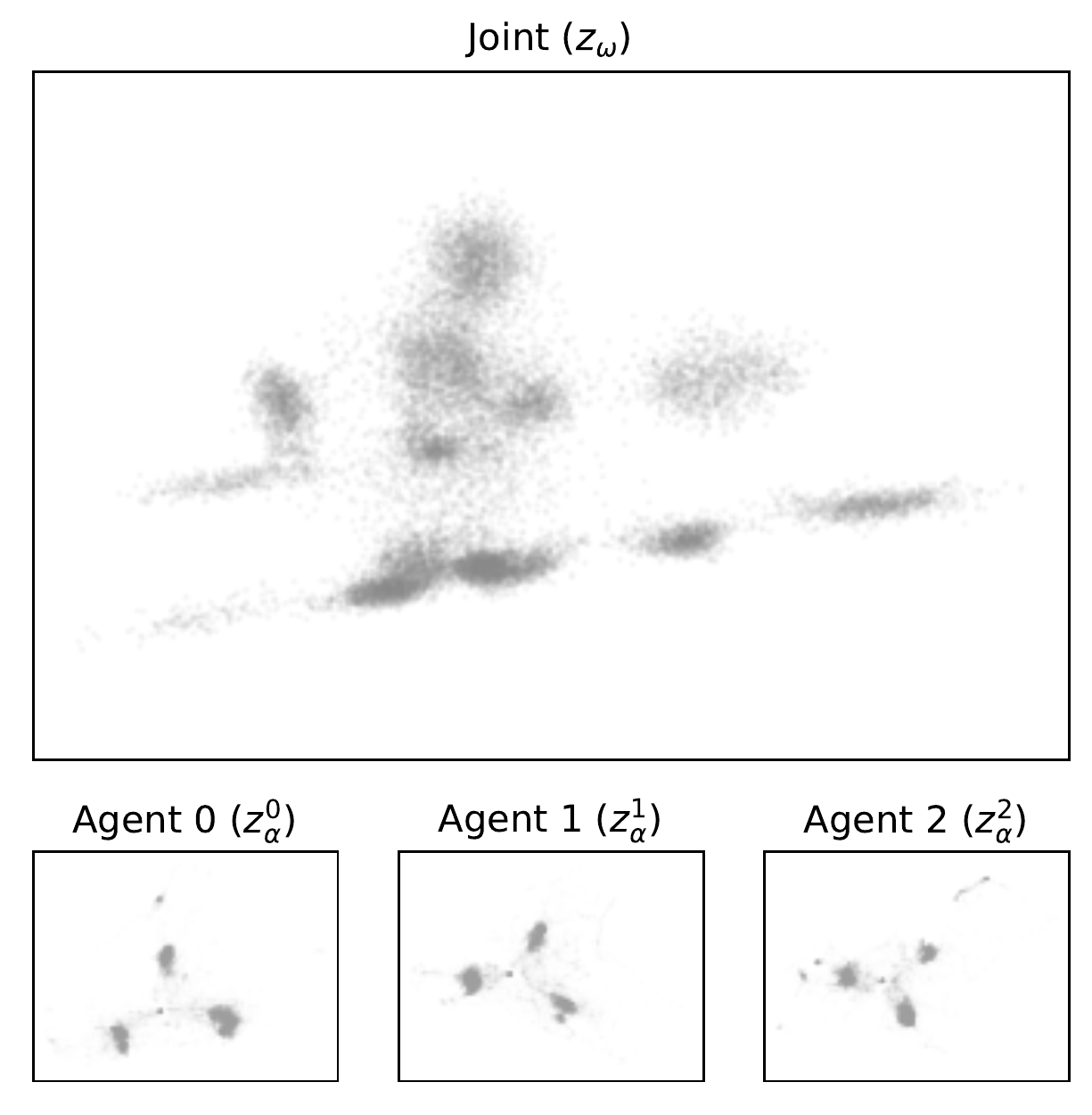}
        \caption{Latent $z$ dimensionality: 4, VAE KL-loss $\beta$: 1e-2}
        \label{fig:ablative_loss_multimodal_reward_env_b}
    \end{subfigure}\\
    \begin{subfigure}[b]{0.48\textwidth}
        \centering
        \includegraphics[width=\textwidth]{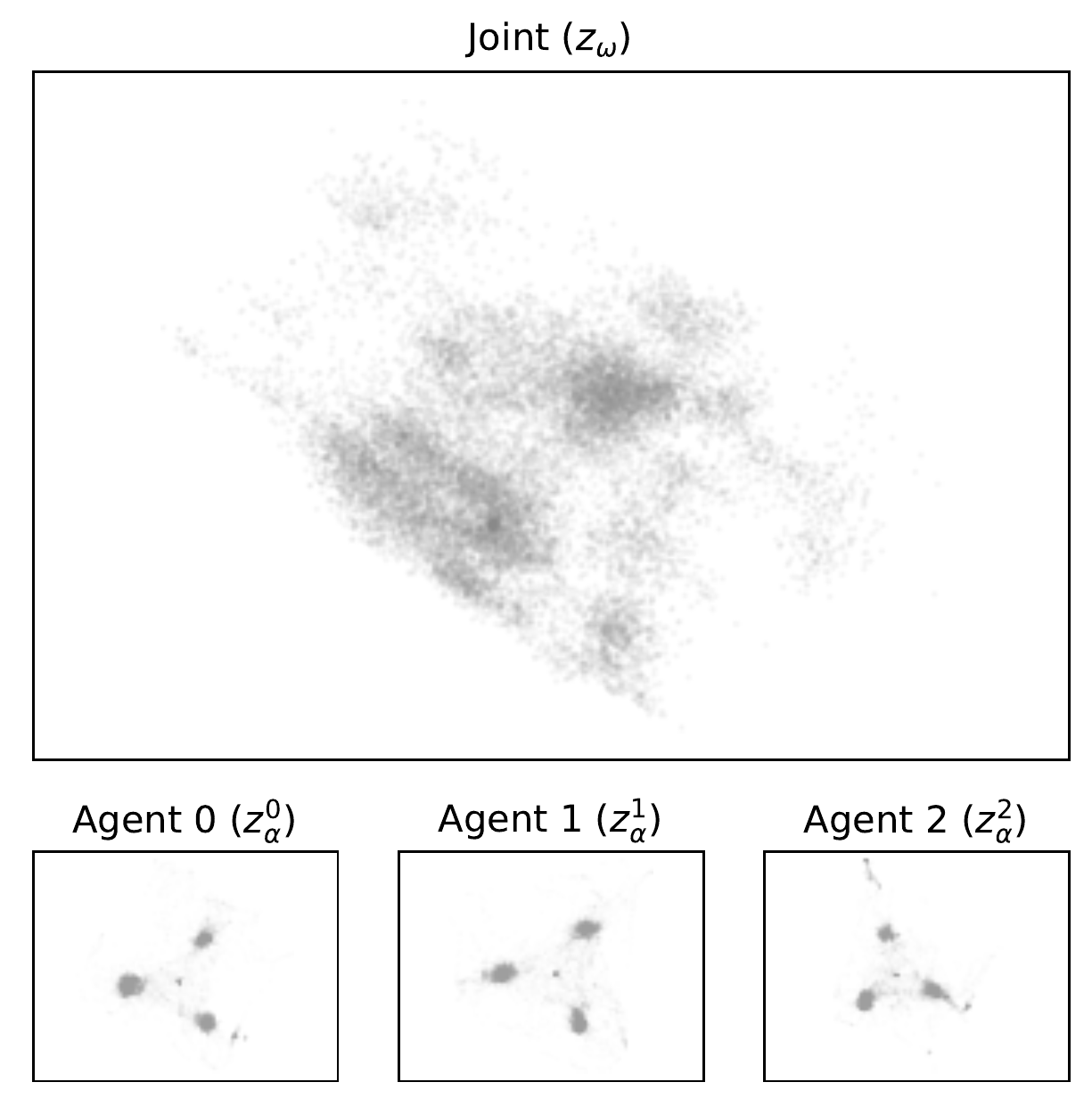}
        \caption{Latent $z$ dimensionality: 8, VAE KL-loss $\beta$: 1e-4}
        \label{fig:ablative_loss_multimodal_reward_env_c}
    \end{subfigure}%
    \hfill
    \begin{subfigure}[b]{0.48\textwidth}
        \centering
        \includegraphics[width=\textwidth]{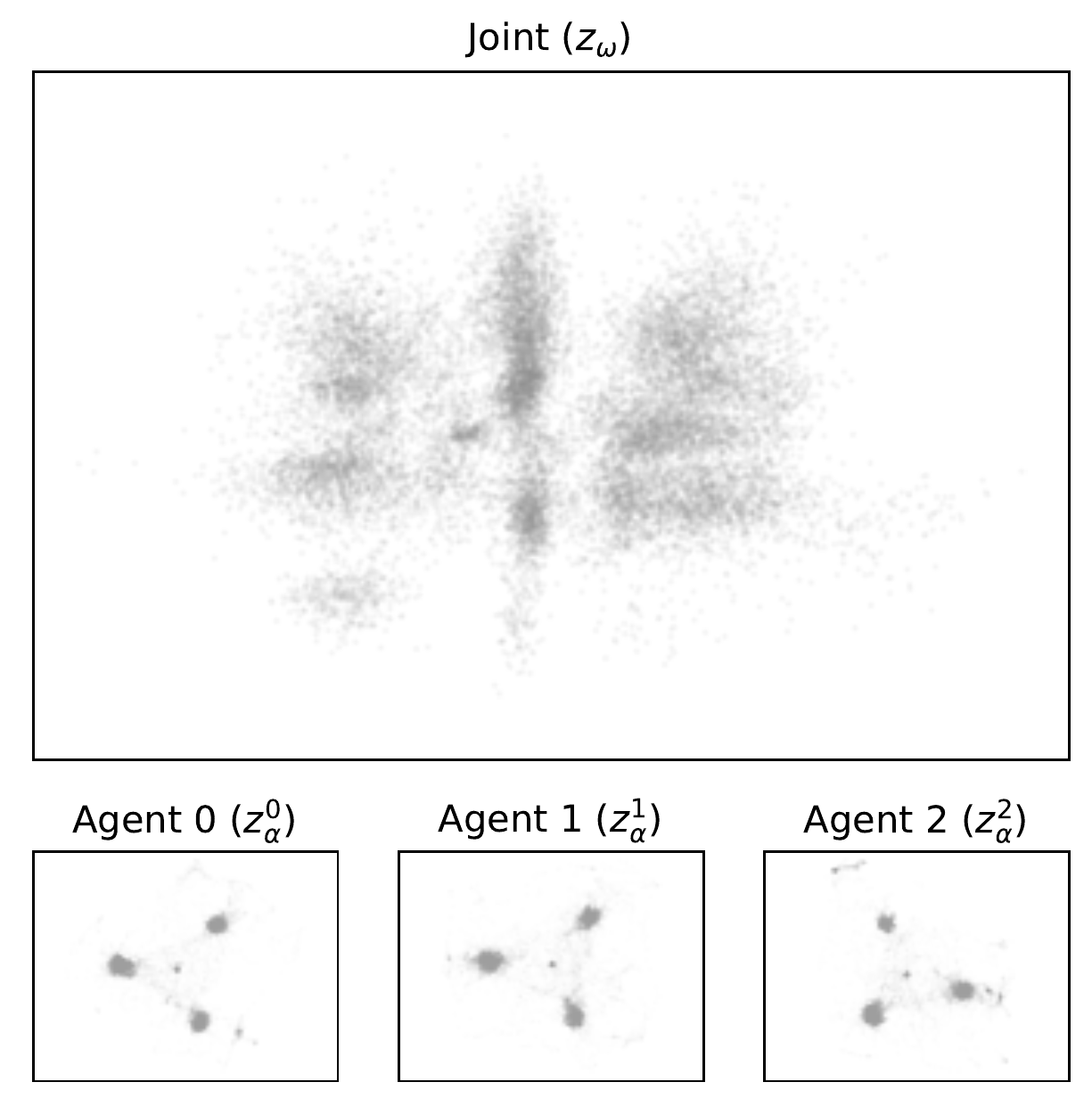}
        \caption{Latent $z$ dimensionality: 8, VAE KL-loss $\beta$: 1e-2}
        \label{fig:ablative_loss_multimodal_reward_env_d}
    \end{subfigure}\\
    \caption{Ablations over hyperparameters for 3-agent hill climbing environment. \subref{fig:ablative_loss_multimodal_reward_env} visualizes the reconstruction policy loss throughout \OurAlgAcronym training.
    \subref{fig:ablative_loss_multimodal_reward_env_a}-\subref{fig:ablative_loss_multimodal_reward_env_d} visualize the change in latent space for each of the final MOHBA models learned over these ablations.
    }
    \label{fig:ablative_loss_multimodal_reward_env_all_figs}
\end{figure}

\begin{figure}[h]
    \centering
    \begin{subfigure}[b]{\textwidth}
        \centering
        \includegraphics[width=0.6\textwidth]{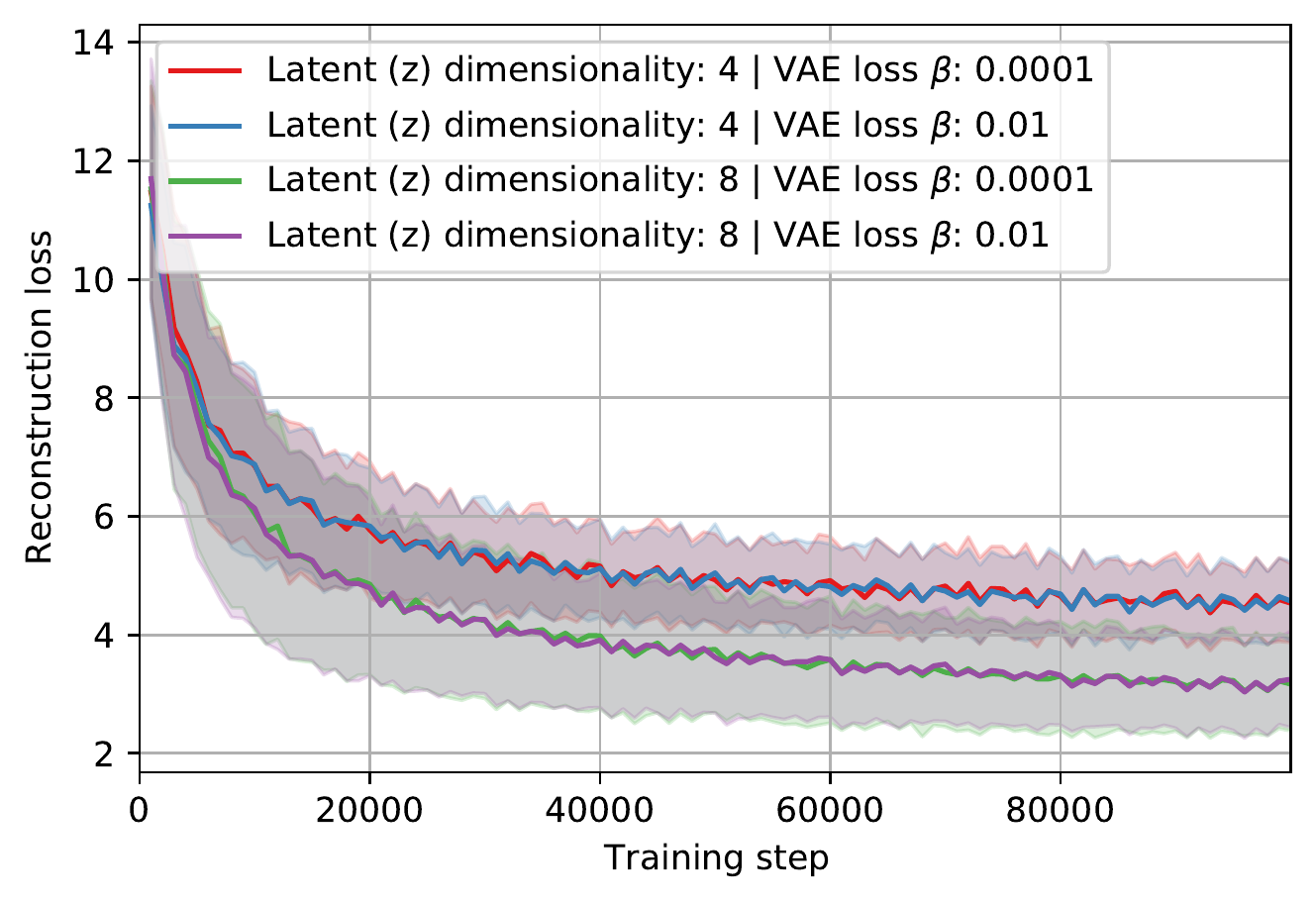}
        \caption{}
        \label{fig:ablative_loss_left_right_env}
    \end{subfigure}\\
    \begin{subfigure}[b]{0.48\textwidth}
        \centering
        \includegraphics[width=\textwidth]{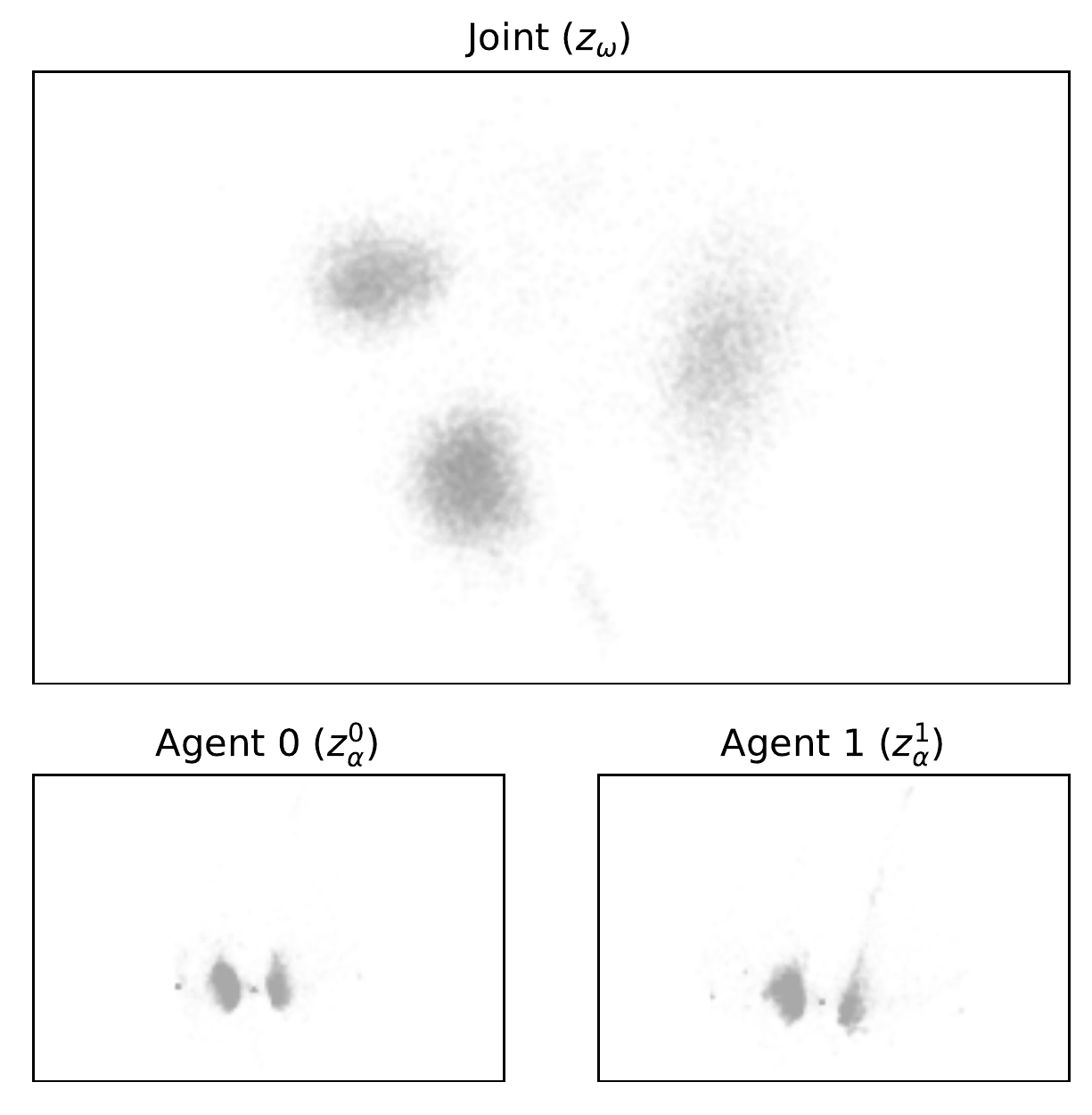}
        \caption{Latent $z$ dimensionality: 4, VAE KL-loss $\beta$: 1e-4}
        \label{fig:ablative_loss_left_right_env_a}
    \end{subfigure}%
    \hfill
    \begin{subfigure}[b]{0.48\textwidth}
        \centering
        \includegraphics[width=\textwidth]{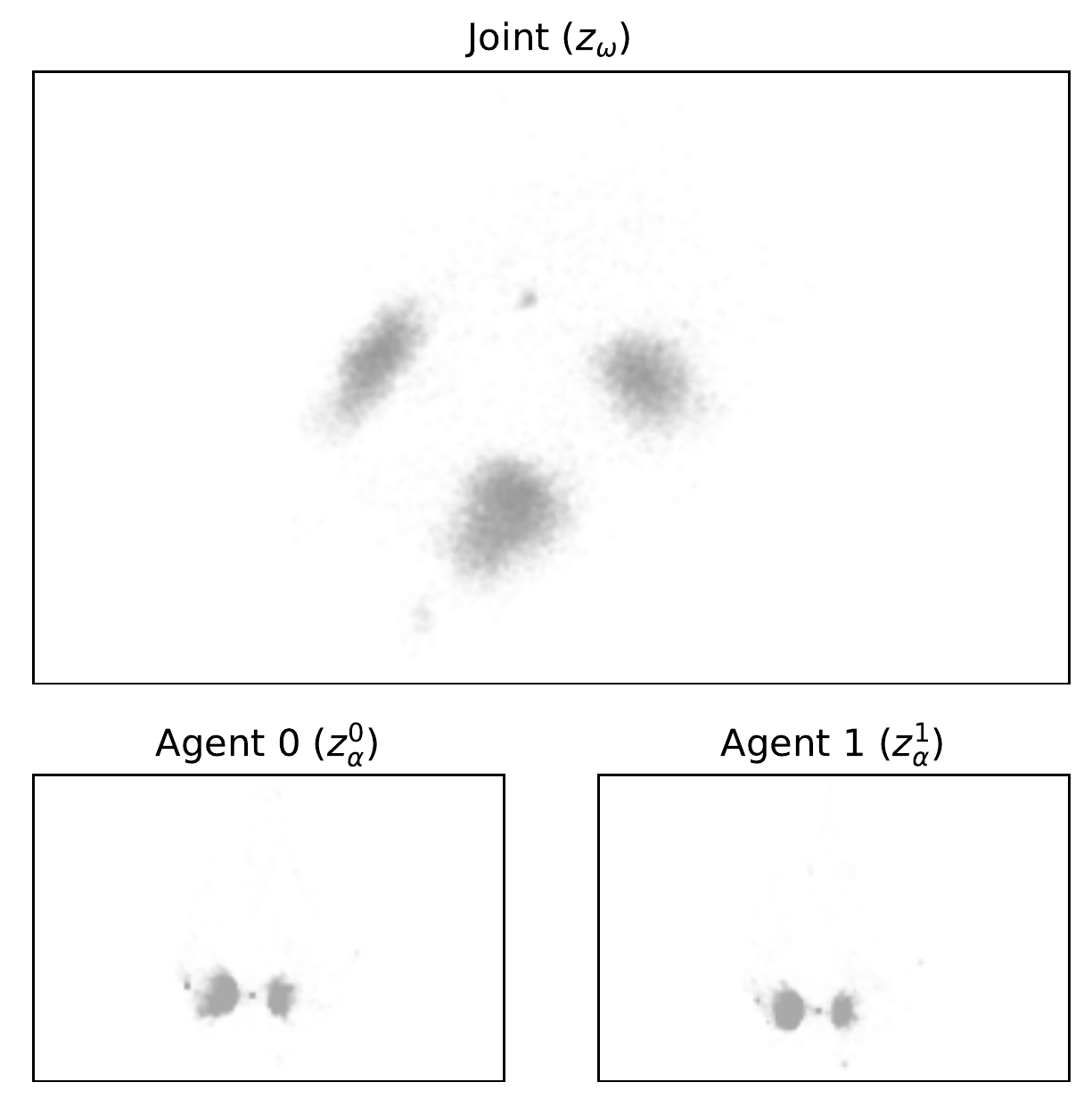}
        \caption{Latent $z$ dimensionality: 4, VAE KL-loss $\beta$: 1e-2}
        \label{fig:ablative_loss_left_right_env_b}
    \end{subfigure}\\
    \begin{subfigure}[b]{0.48\textwidth}
        \centering
        \includegraphics[width=\textwidth]{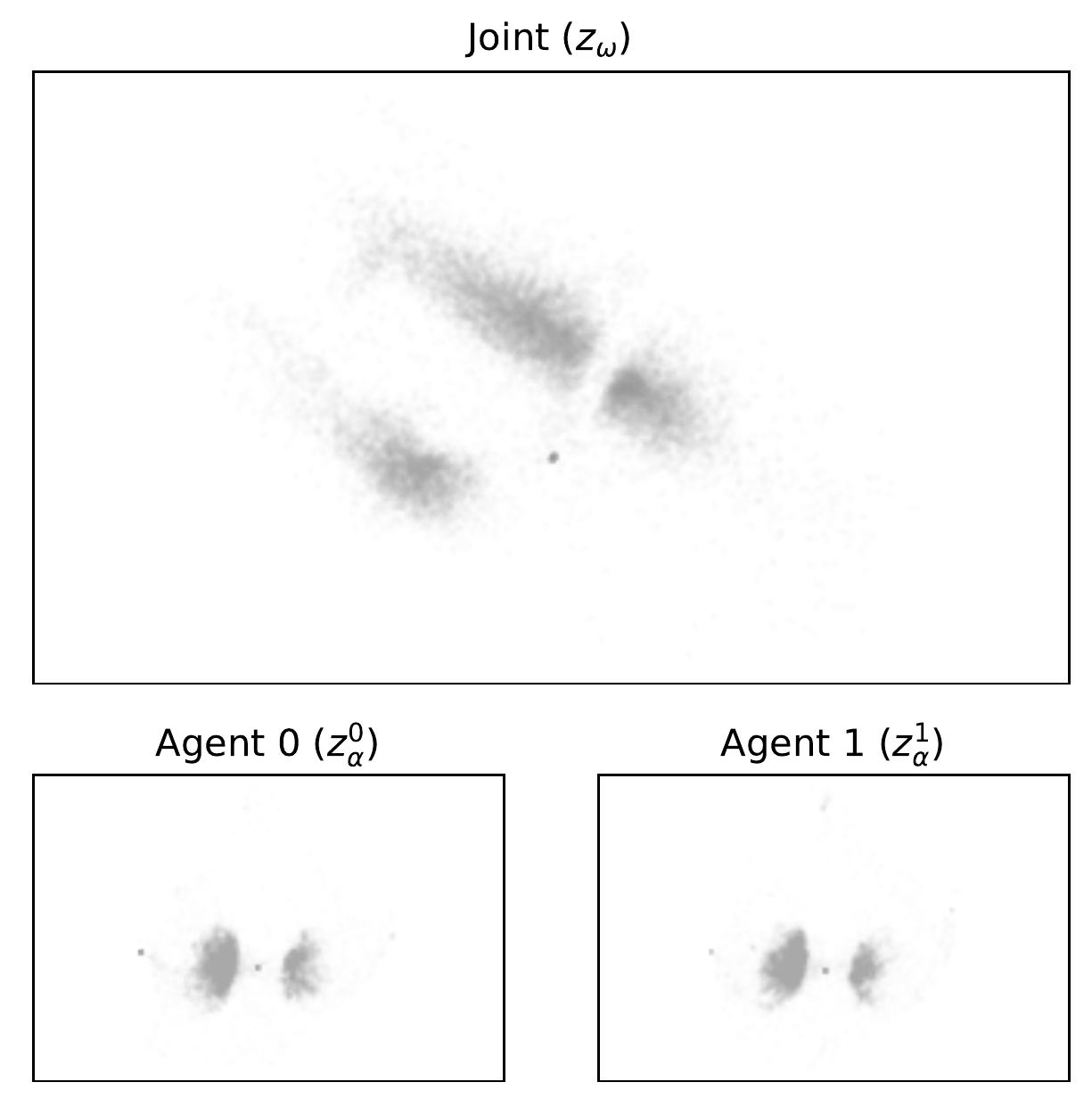}
        \caption{Latent $z$ dimensionality: 8, VAE KL-loss $\beta$: 1e-4}
        \label{fig:ablative_loss_left_right_env_c}
    \end{subfigure}%
    \hfill
    \begin{subfigure}[b]{0.48\textwidth}
        \centering
        \includegraphics[width=\textwidth]{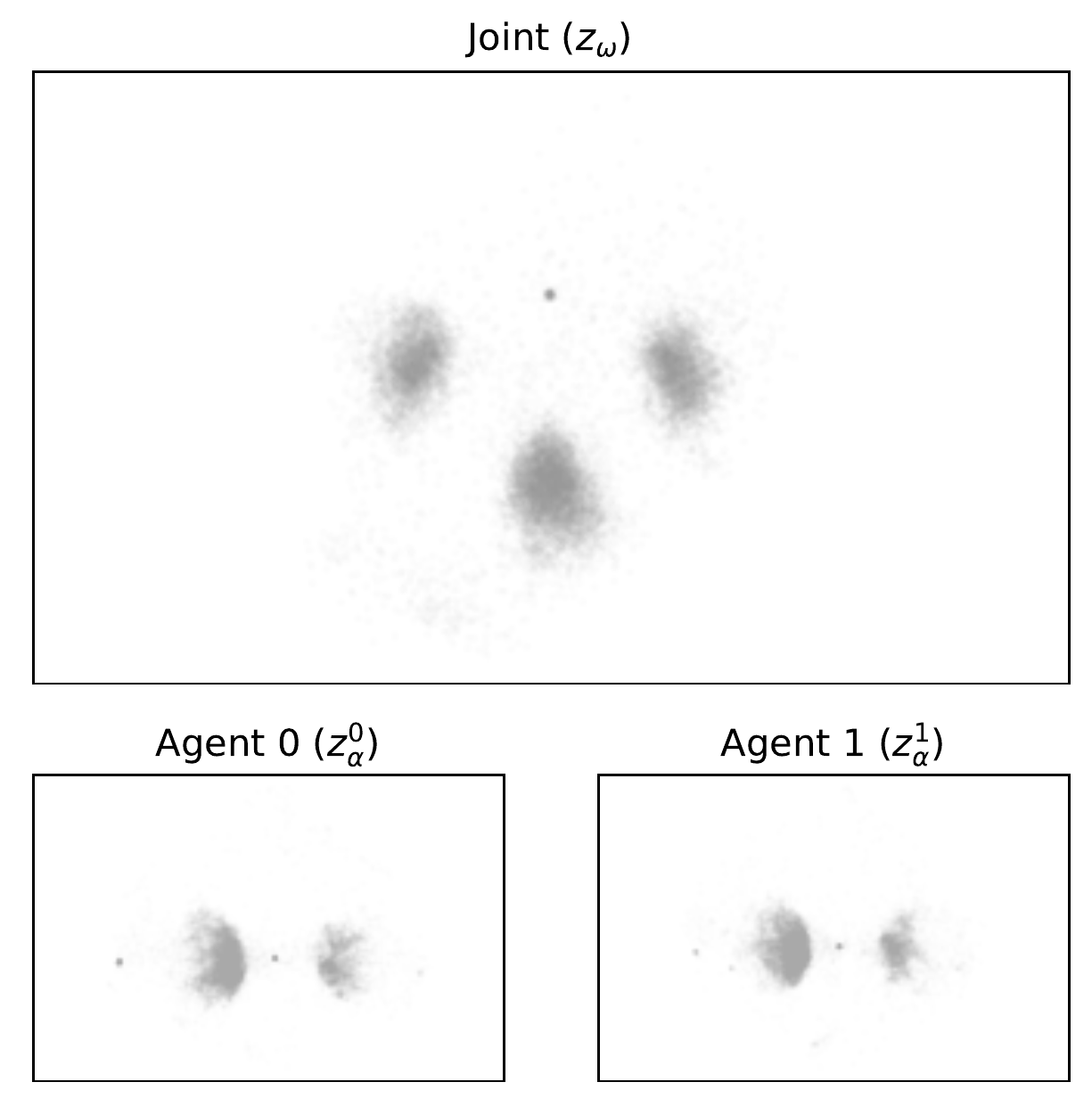}
        \caption{Latent $z$ dimensionality: 8, VAE KL-loss $\beta$: 1e-2}
        \label{fig:ablative_loss_left_right_env_d}
    \end{subfigure}\\
    \caption{Ablations over hyperparameters for 2-agent coordination game. 
    \subref{fig:ablative_loss_left_right_env} visualizes the reconstruction policy loss throughout \OurAlgAcronym training.
    \subref{fig:ablative_loss_left_right_env_a}-\subref{fig:ablative_loss_left_right_env_d} visualize the change in latent space for each of the final MOHBA models learned over these ablations.
    }
    \label{fig:ablative_loss_left_right_env_all_figs}
\end{figure}

\begin{figure}[h]
    \centering
    \begin{subfigure}[b]{\textwidth}
        \centering
        \includegraphics[width=0.6\textwidth]{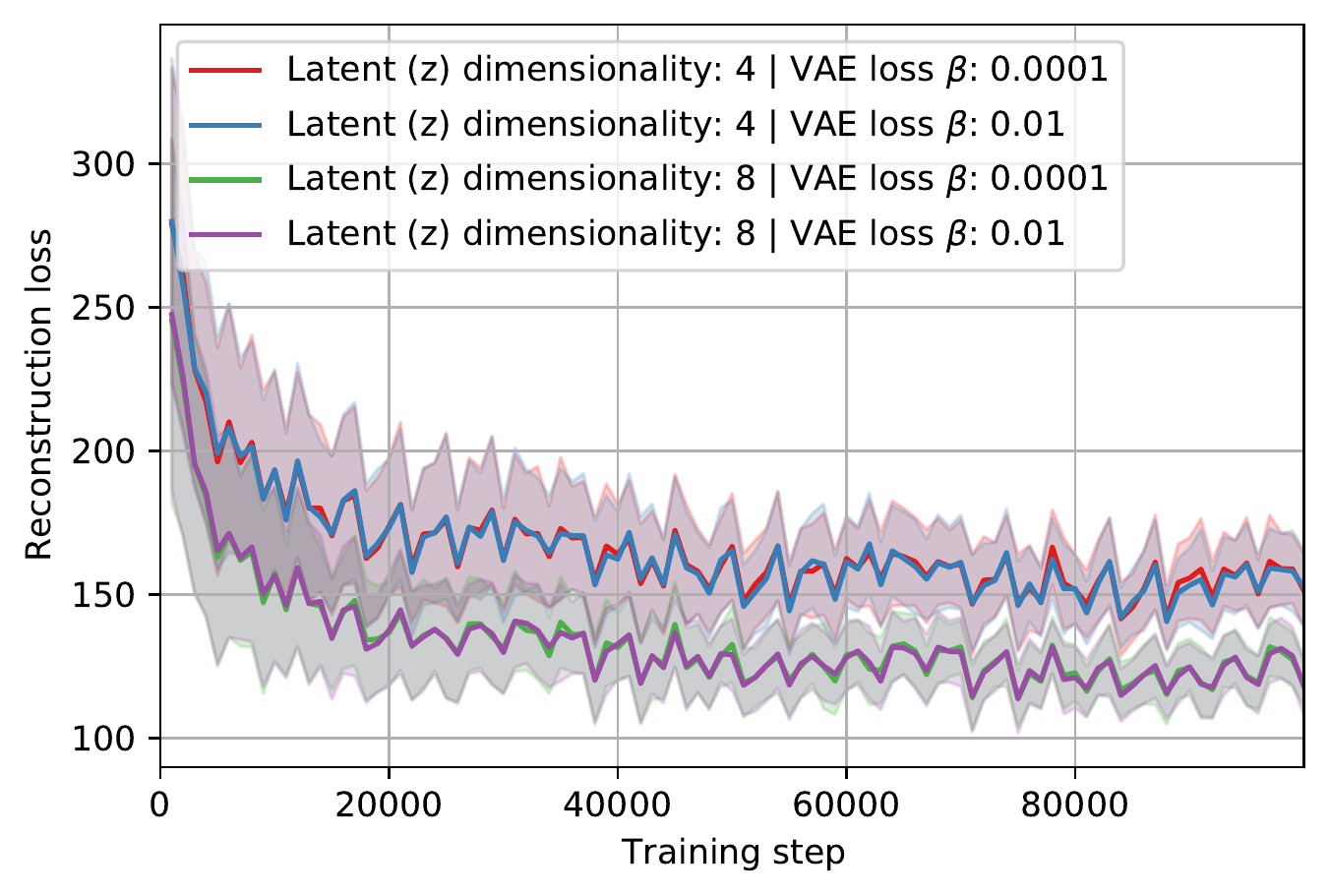}
        \caption{}
        \label{fig:ablative_loss_mujoco_halfcheetah}
    \end{subfigure}\\
    \begin{subfigure}[b]{0.48\textwidth}
        \centering
        \includegraphics[width=\textwidth]{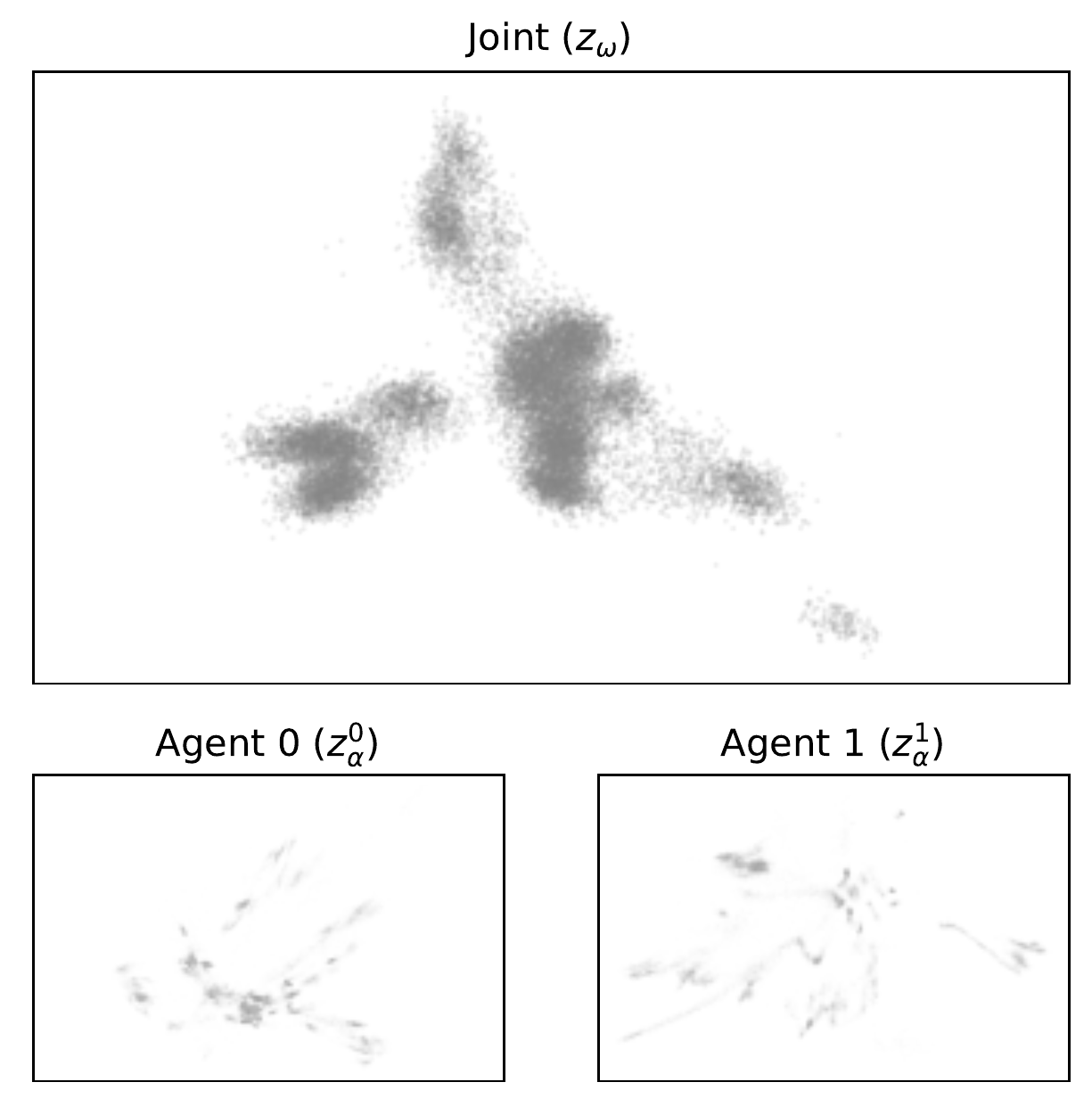}
        \caption{Latent $z$ dimensionality: 4, VAE KL-loss $\beta$: 1e-4}
        \label{fig:ablative_loss_mujoco_halfcheetah_a}
    \end{subfigure}%
    \hfill
    \begin{subfigure}[b]{0.48\textwidth}
        \centering
        \includegraphics[width=\textwidth]{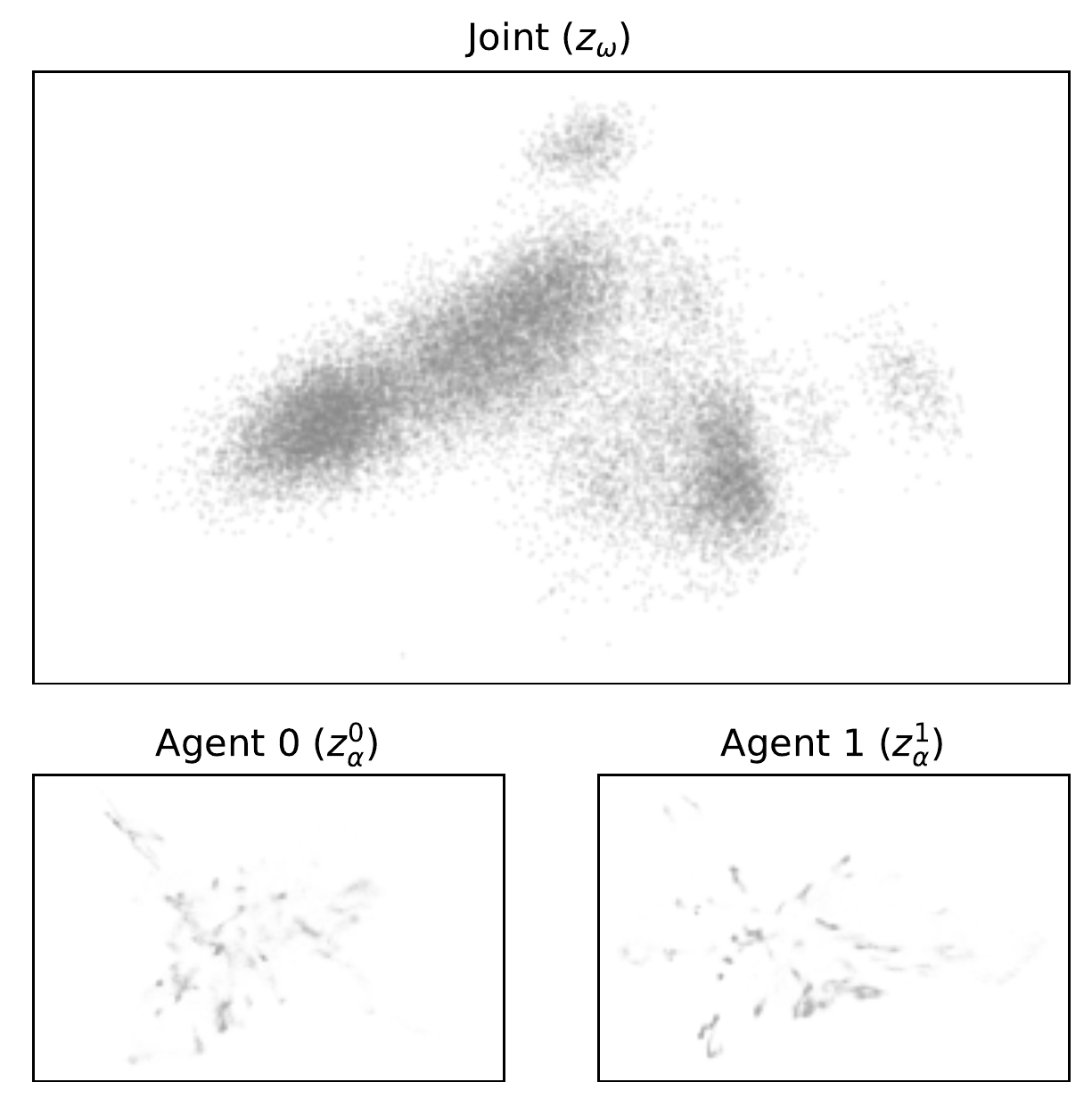}
        \caption{Latent $z$ dimensionality: 4, VAE KL-loss $\beta$: 1e-2}
        \label{fig:ablative_loss_mujoco_halfcheetah_b}
    \end{subfigure}\\
    \begin{subfigure}[b]{0.48\textwidth}
        \centering
        \includegraphics[width=\textwidth]{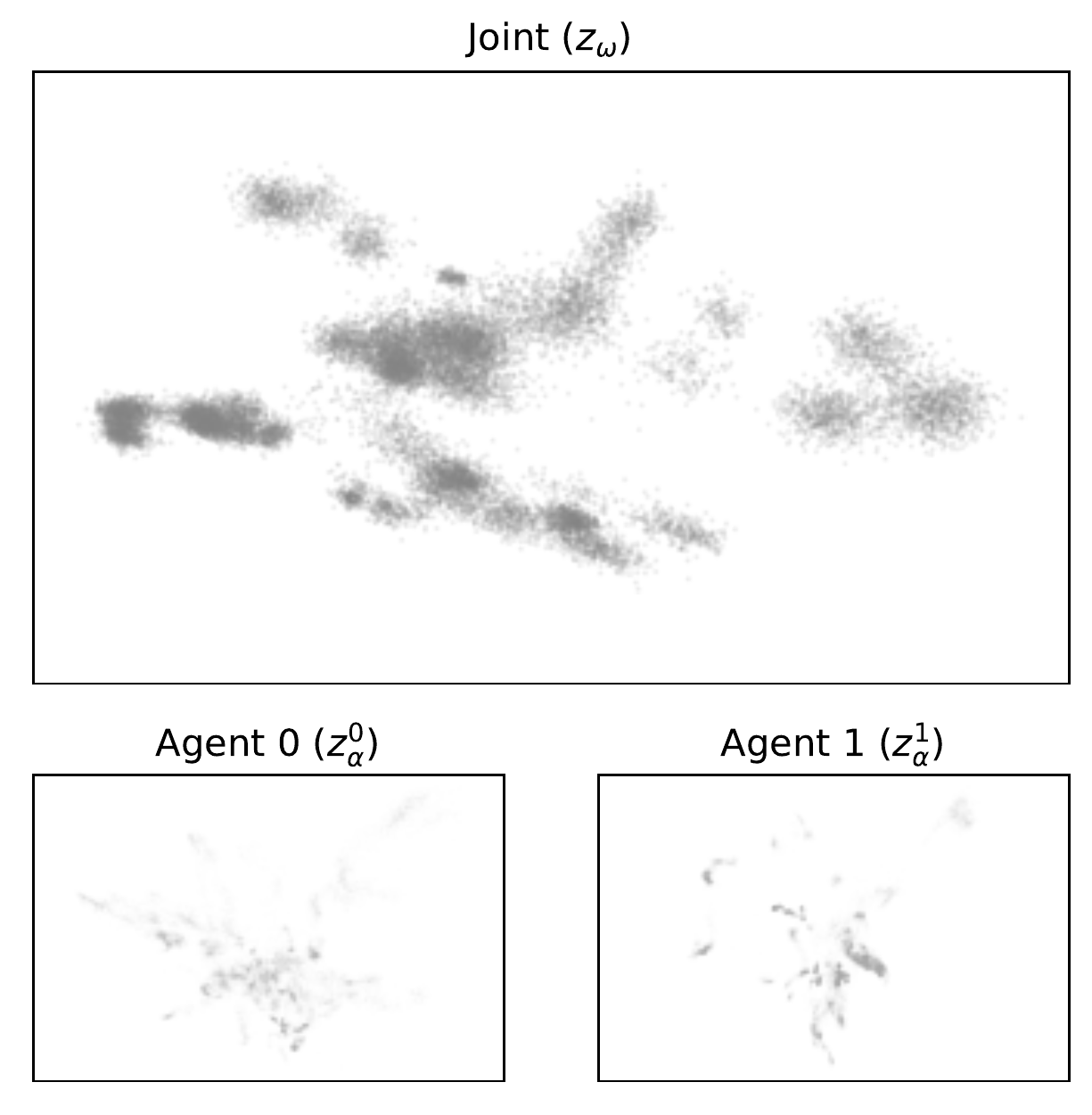}
        \caption{Latent $z$ dimensionality: 8, VAE KL-loss $\beta$: 1e-4}
        \label{fig:ablative_loss_mujoco_halfcheetah_c}
    \end{subfigure}%
    \hfill
    \begin{subfigure}[b]{0.48\textwidth}
        \centering
        \includegraphics[width=\textwidth]{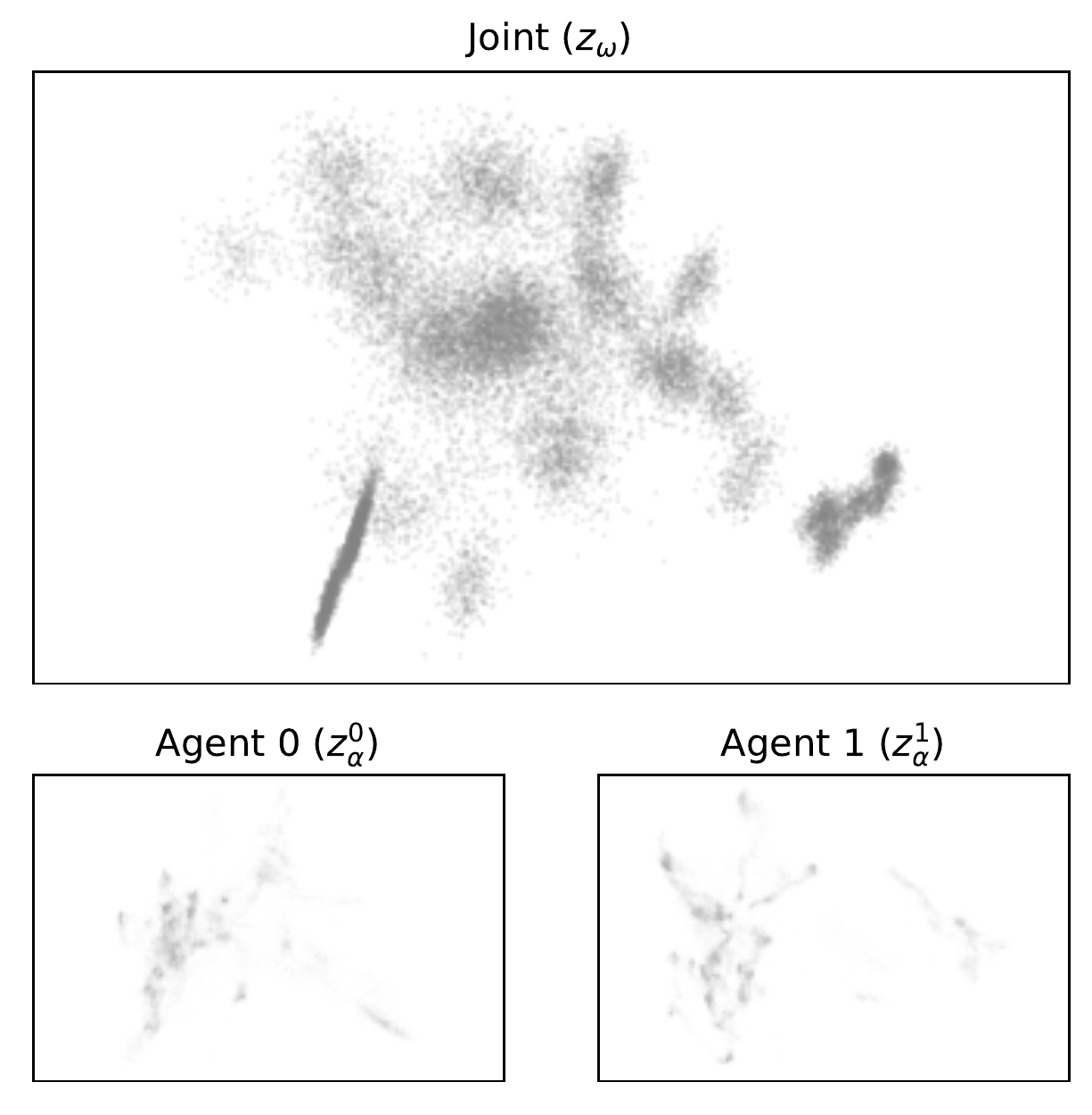}
        \caption{Latent $z$ dimensionality: 8, VAE KL-loss $\beta$: 1e-2}
        \label{fig:ablative_loss_mujoco_halfcheetah_d}
    \end{subfigure}\\
    \caption{Ablations over hyperparameters for Multiagent MuJoCo HalfCheetah.
    \subref{fig:ablative_loss_mujoco_halfcheetah} visualizes the reconstruction policy loss throughout \OurAlgAcronym training.
    \subref{fig:ablative_loss_mujoco_halfcheetah_a}-\subref{fig:ablative_loss_mujoco_halfcheetah_d} visualize the change in latent space for each of the final MOHBA models learned over these ablations.
    Increasing the dimensionality of $z$ tends to increase the number of joint clusters identified (e.g., \subref{fig:ablative_loss_mujoco_halfcheetah_b} versus \subref{fig:ablative_loss_mujoco_halfcheetah_d}).
    Increasing the KL $\beta$ term from $1e-4$ to $1e-2$ tends to result in clusters that overlap more / are `softer' and slightly less distinguishable (e.g., comparing $\zhi$ in \subref{fig:ablative_loss_mujoco_halfcheetah_b} versus \subref{fig:ablative_loss_mujoco_halfcheetah_c}).
    }
    \label{fig:ablative_loss_mujoco_halfcheetah_all_figs}
\end{figure}

First, in terms of the raw policy reconstruction loss, we find in \cref{fig:ablative_loss_multimodal_reward_env,fig:ablative_loss_left_right_env,fig:ablative_loss_mujoco_halfcheetah} that the latent dimensionality has a higher impact on policy reconstruction performance than the weighing term $\beta$; intuitively, increasing latent dimensionality leads to better reconstruction (lower loss) due to the latent space being able to encode more behavioral information about the agent trajectories.
By contrast, the effects of the KL weighing term are more negligible in terms of reconstruction loss.

Second, we inspect the effects of these hyperparameter sweeps on the learned latent distributions themselves.
Specifically, the respective panels (b)-(e) of each of \cref{fig:ablative_loss_multimodal_reward_env_all_figs}, \cref{fig:ablative_loss_left_right_env_all_figs}, and \cref{fig:ablative_loss_mujoco_halfcheetah_all_figs} visualize the change in latent space structure as a function of the latent dimensionality and $\beta$.
At a high-level, prominent behavioral clusters mentioned in the main text are re-discovered throughout these parameter sweeps (e.g., the three local clusters for the hill climbing environment, the joint and local clusters for the coordination environment, and the various walking gait clusters for the HalfCheetah environment).
For the more complex latent spaces (e.g., HalfCheetah), increasing the dimensionality of $z$ tends to increase the number of joint clusters identified (e.g., compare \cref{fig:ablative_loss_mujoco_halfcheetah_b} versus \cref{fig:ablative_loss_mujoco_halfcheetah_d});
this is intuitive as a larger latent space results in a richer encoding, capable of distinguishing more nuanced behaviors.
Increasing the KL $\beta$ term from $1e-4$ to $1e-2$ tends to result in clusters that overlap more / are `softer' and slightly less distinguishable (e.g., comparing $\zhi$ in \cref{fig:ablative_loss_mujoco_halfcheetah_b} versus \cref{fig:ablative_loss_mujoco_halfcheetah_c});
this also aligns well with intuition, as increasing the $\beta$ term prioritizes the KL divergence between the posterior and prior, which deprioritizes disentangling the behaviors for the policy reconstruction term $\cref{eq:loss_reconstruction}$.

Overall, these results provide us intuition in terms of the role of these hyperparameters in the behavior clusters learned. 
At a high level, it appears that the sensitivity of the results to these hyperparameters is fairly low (in the range of values explored in these experiments), thus allowing the high level behaviors discovered to remain reasonably distinctive across the various sweeps.



\subsection{Additional Dataset Details / Reward Distributions}\label{appendix:additional_dataset_details}
This section provides additional details and statistics regarding the trajectory datasets used for evaluation.
Specifically, one of the core properties of datasets typically used for offline RL is the inclusion of a large number of sub-optimal and random trajectories~\citep{fu2020d4rl}.
As noted earlier, the trajectory data for the agents in our experiments is collected from numerous underlying MARL training runs for each domain.
To help ensure the collection and analysis of a diverse set of behaviors including sub-optimal ones, we not only collect trajectories at the end of MARL training, but rather from policy checkpoints throughout all of training.
All such trajectories, including those stemming from randomly-initialized agent policies at the beginning of training, are shuffled together and used to simultaneously train \OurAlgAcronym.

\begin{table}[h]
    \centering
    \setlength{\extrarowheight}{1pt}
    \caption{Trajectory return statistics for each of the datasets analyzed.}
    \label{table:reward_stats_percent_of_expert}
    \begin{tabular}{cc}
    \toprule
    Domain                                   & \% of trajectories with total return $ < 0.5 \times$ (maximum observed return)\\
    \midrule
    HalfCheetah (2 agents) & 92.69\%\\
    AntWalker (4 agents)   & 22.47\%\\
    Hill-climbing                            & 23.91\% \\
    Coordination game                        & 11.83 \%\\
    \bottomrule
    \end{tabular}
\end{table}

Here we more closely inspect our datasets to better understand the distribution of agent behaviors.
At a high level, we first note that our datasets consist of a significant proportion of sub-optimal trajectories.
Specifically, in \cref{table:reward_stats_percent_of_expert}, we note the percent of trajectories in each of the considered datasets that attain less than 50\% of maximum observed return.
Note that in the high-dimensional MultiAgent MuJoCo environments especially, a large proportion of low-reward trajectories exist in the dataset, with the HalfCheetah dataset being a particularly notable one where 92.69\% of datasets attain less than 50\% of max observed return. 

\begin{figure}[h]
    \centering
    \begin{subfigure}[b]{0.7\textwidth}
        \centering
        \includegraphics[width=\textwidth]{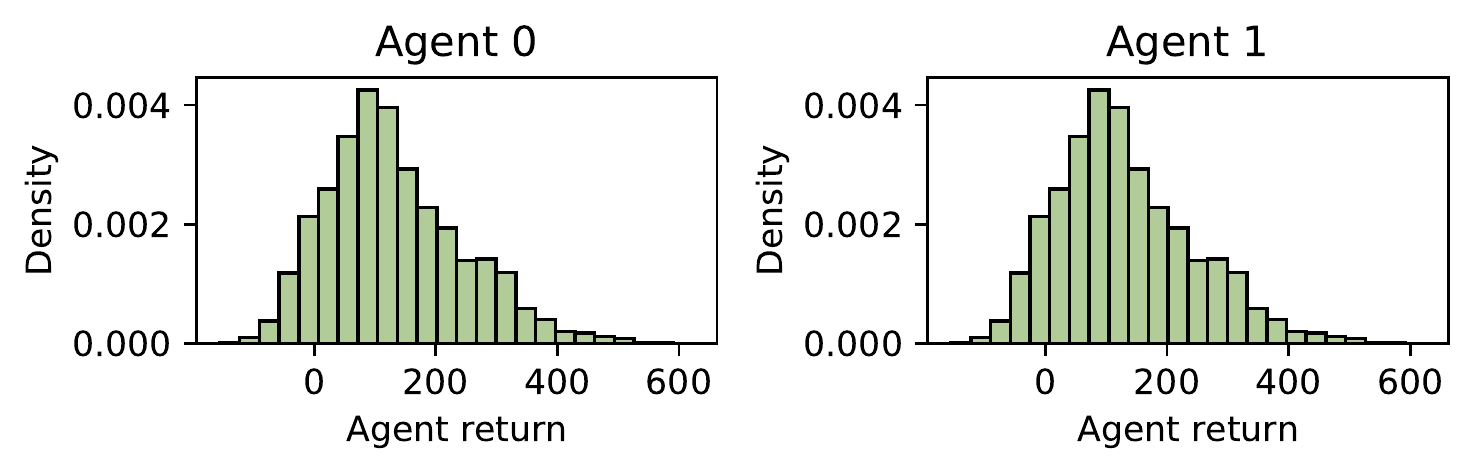}
        \caption{Agent-wise return distribution.}
        \label{fig:mujoco_halfcheetah_reward_statistics}
    \end{subfigure}\\
    \begin{subfigure}[b]{\textwidth}
        \centering
        \includegraphics[width=\textwidth]{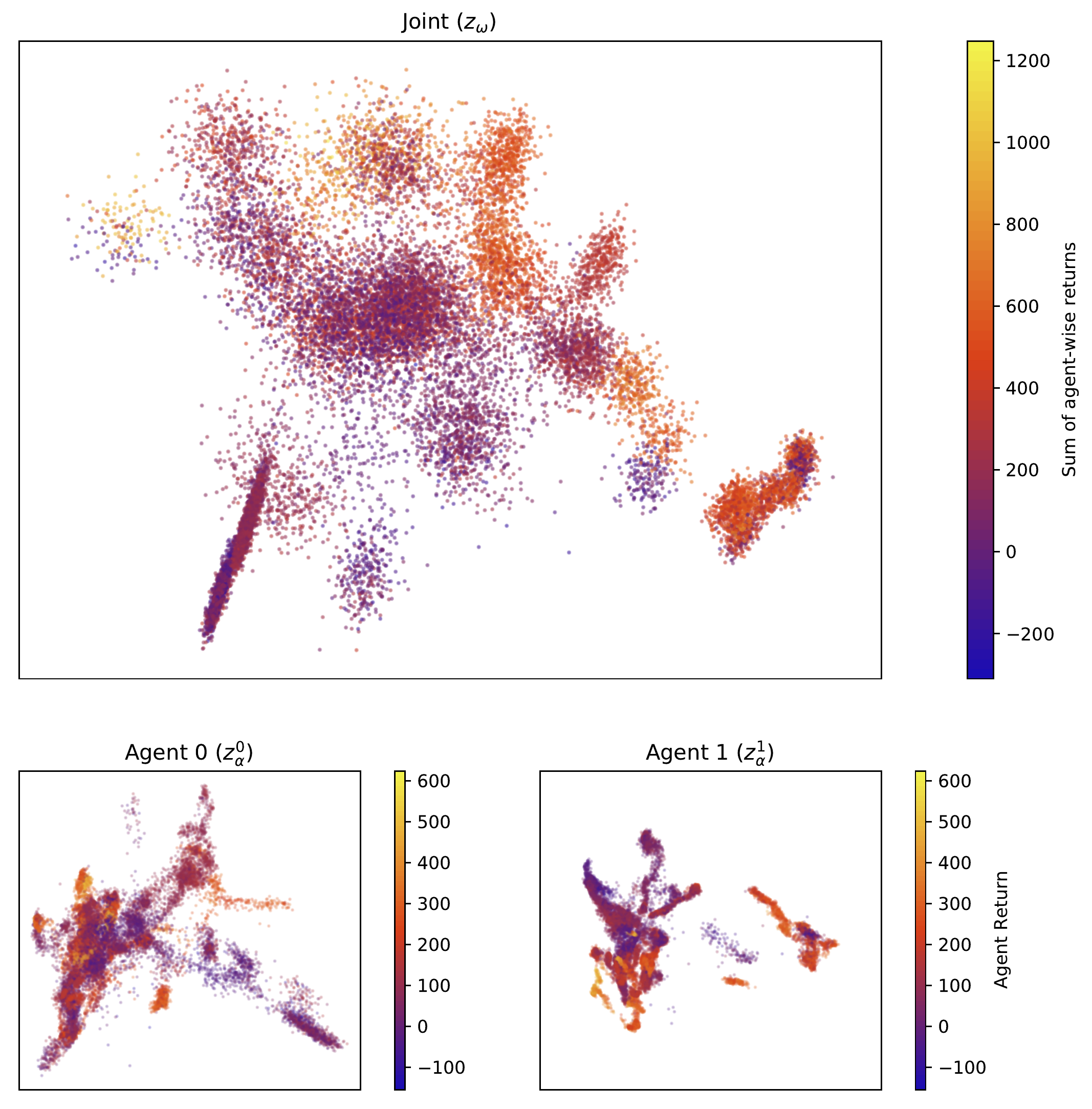}
        \caption{Latent spaces, labeled by trajectory return.}
        \label{fig:mujoco_halfcheetah_env_latent_rewards}
    \end{subfigure}
    \caption{
        Reward statistics for the Multiagent MuJoCo 2-agent HalfCheetah environment.
        \subref{fig:mujoco_halfcheetah_reward_statistics} visualizes the distribution of agent returns in the dataset.
        \subref{fig:mujoco_halfcheetah_env_latent_rewards} visualizes the agent behavior spaces, with trajectories labeled by the return attained.
    }
    \label{fig:mujoco_halfcheetah_env_reward_stats_all}
\end{figure}

\begin{figure}[h]
    \centering
    \begin{subfigure}[b]{\textwidth}
        \centering
        \includegraphics[width=\textwidth]{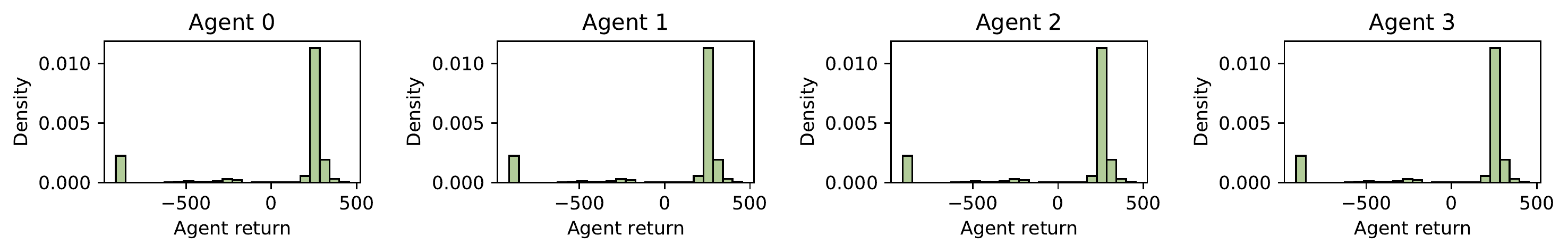}
        \caption{Agent-wise return distribution.}
        \label{fig:mujoco_antwalker_reward_statistics}
    \end{subfigure}\\
    \begin{subfigure}[b]{\textwidth}
        \centering
        \includegraphics[width=\textwidth]{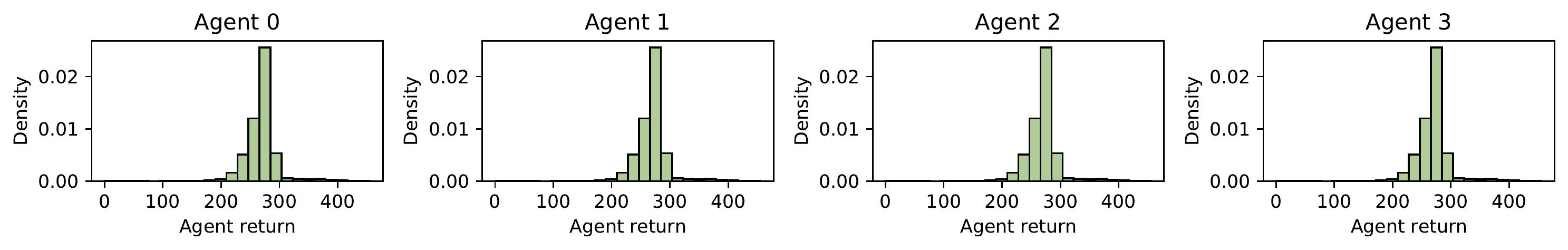}
        \caption{Agent-wise return distribution (anomalous trajectories with returns less than 0 removed).}
        \label{fig:mujoco_antwalker_reward_statistics_thresh}
    \end{subfigure}\\
    \begin{subfigure}[b]{\textwidth}
        \centering
        \includegraphics[width=\textwidth]{figs/mujoco_antwalker_env_latent_rewards.pdf}
        \caption{Latent spaces, labeled by trajectory return.}
        \label{fig:mujoco_antwalker_env_latent_rewards}
    \end{subfigure}
    \caption{
        Reward statistics for the Multiagent MuJoCo 4-agent AntWalker environment.
        \subref{fig:mujoco_antwalker_reward_statistics} visualizes the distribution of agent returns in the dataset.
        Interestingly, we observe a bimodal return distribution (seemingly consisting of low return and higher return trajectories.
        Closer inspection reveals that a large number of MARL training runs in this domain result in highly sub-optimal behaviors with very low returns;
        in \subref{fig:mujoco_antwalker_reward_statistics_thresh}, we visualize the distribution of returns with these low-return trajectories excluded, which better illustrates the distribution of medium-high reward trajectories in this dataset.
        \subref{fig:mujoco_antwalker_env_latent_rewards} visualizes the agent behavior spaces, with trajectories labeled by the return attained.
    }
    \label{fig:mujoco_antwalker_env_reward_stats_all}
\end{figure}

\begin{figure}[h]
    \centering
    \begin{subfigure}[b]{0.9\textwidth}
        \centering
        \includegraphics[width=\textwidth]{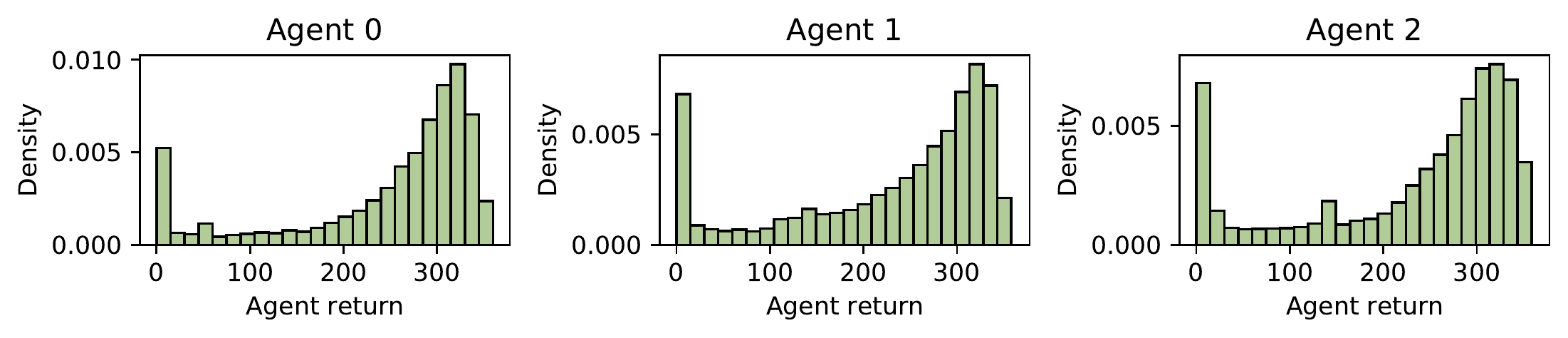}
        \caption{Agent-wise return distribution.}
        \label{fig:multimodal_reward_env_reward_statistics}
    \end{subfigure}\\
    \begin{subfigure}[b]{\textwidth}
        \centering
        \includegraphics[width=\textwidth]{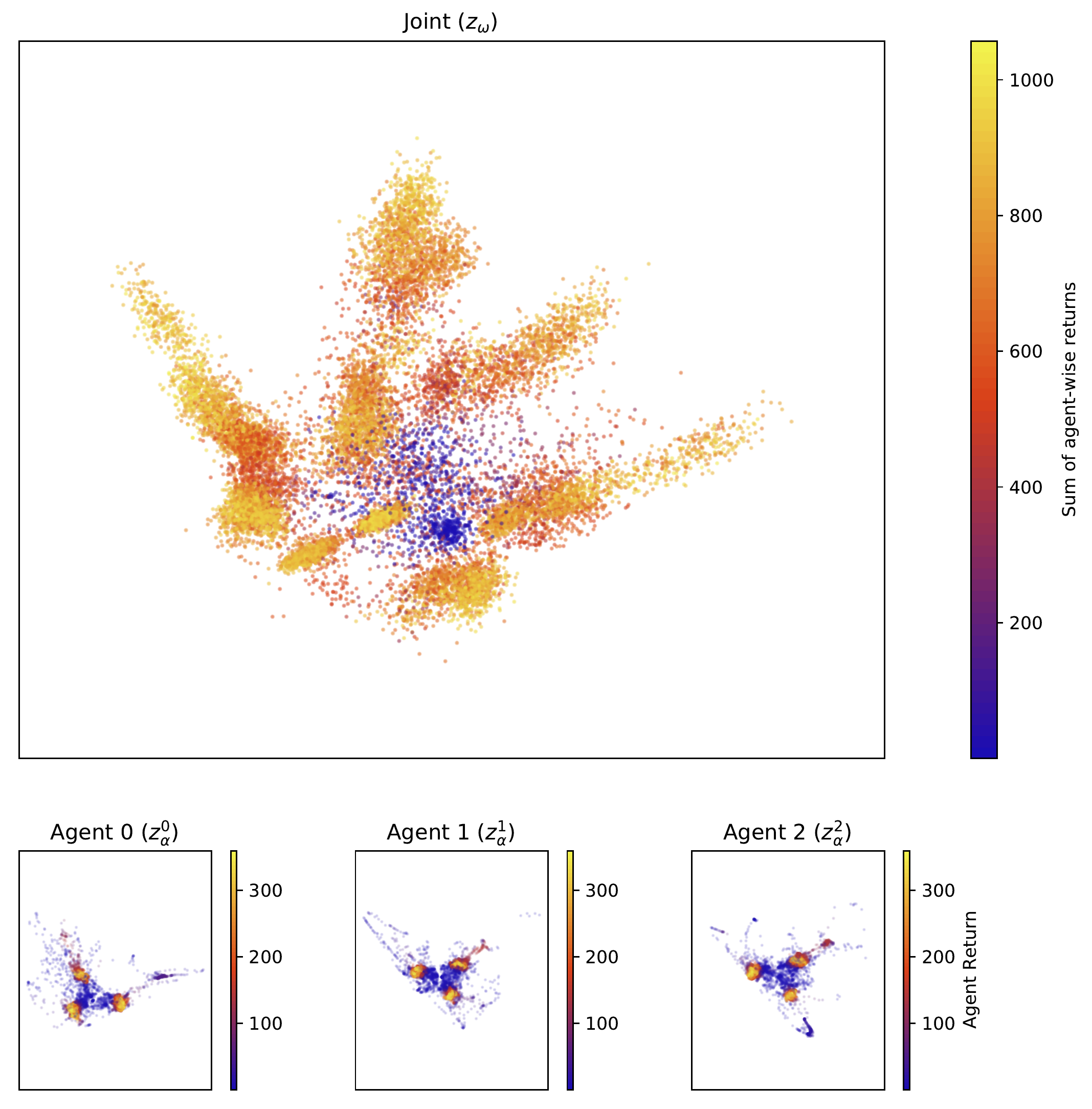}
        \caption{Latent spaces, labeled by trajectory return.}
        \label{fig:multimodal_reward_env_latent_rewards}
    \end{subfigure}
    \caption{
        Reward statistics for the 3-agent hill climbing environment.
        \subref{fig:multimodal_reward_env_reward_statistics} visualizes the distribution of agent returns in the dataset.
        \subref{fig:multimodal_reward_env_latent_rewards} visualizes the agent behavior spaces, with trajectories labeled by the return attained.
    }
    \label{fig:multimodal_reward_env_reward_stats_all}
\end{figure}

\begin{figure}[h]
    \centering
    \begin{subfigure}[b]{0.7\textwidth}
        \centering
        \includegraphics[width=\textwidth]{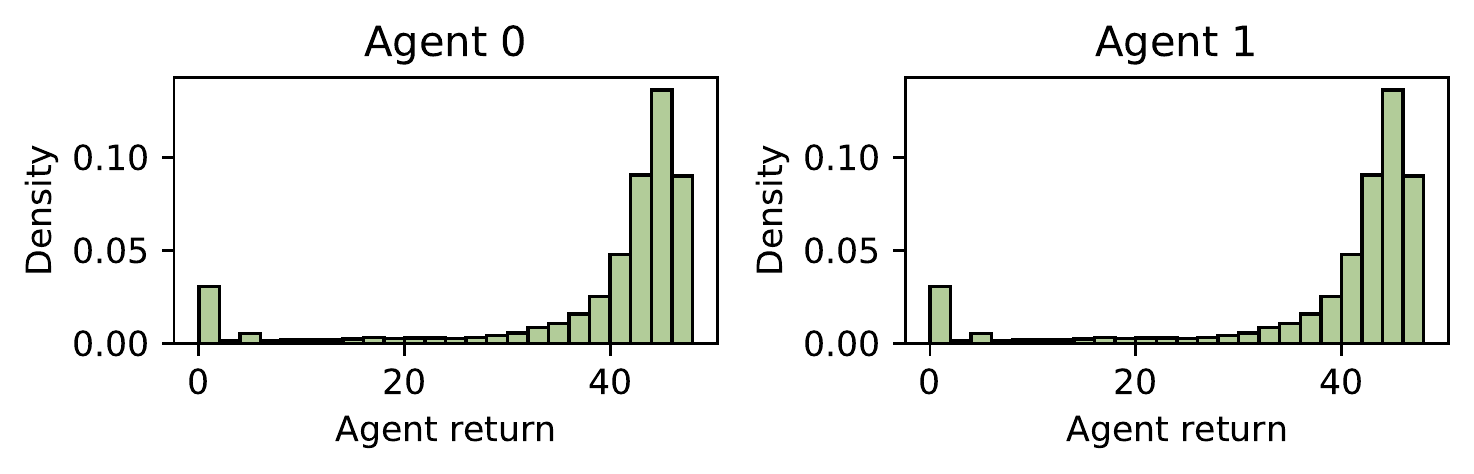}
        \caption{Agent-wise return distribution.}
        \label{fig:left_right_env_latent_reward_statistics}
    \end{subfigure}\\
    \begin{subfigure}[b]{\textwidth}
        \centering
        \includegraphics[width=\textwidth]{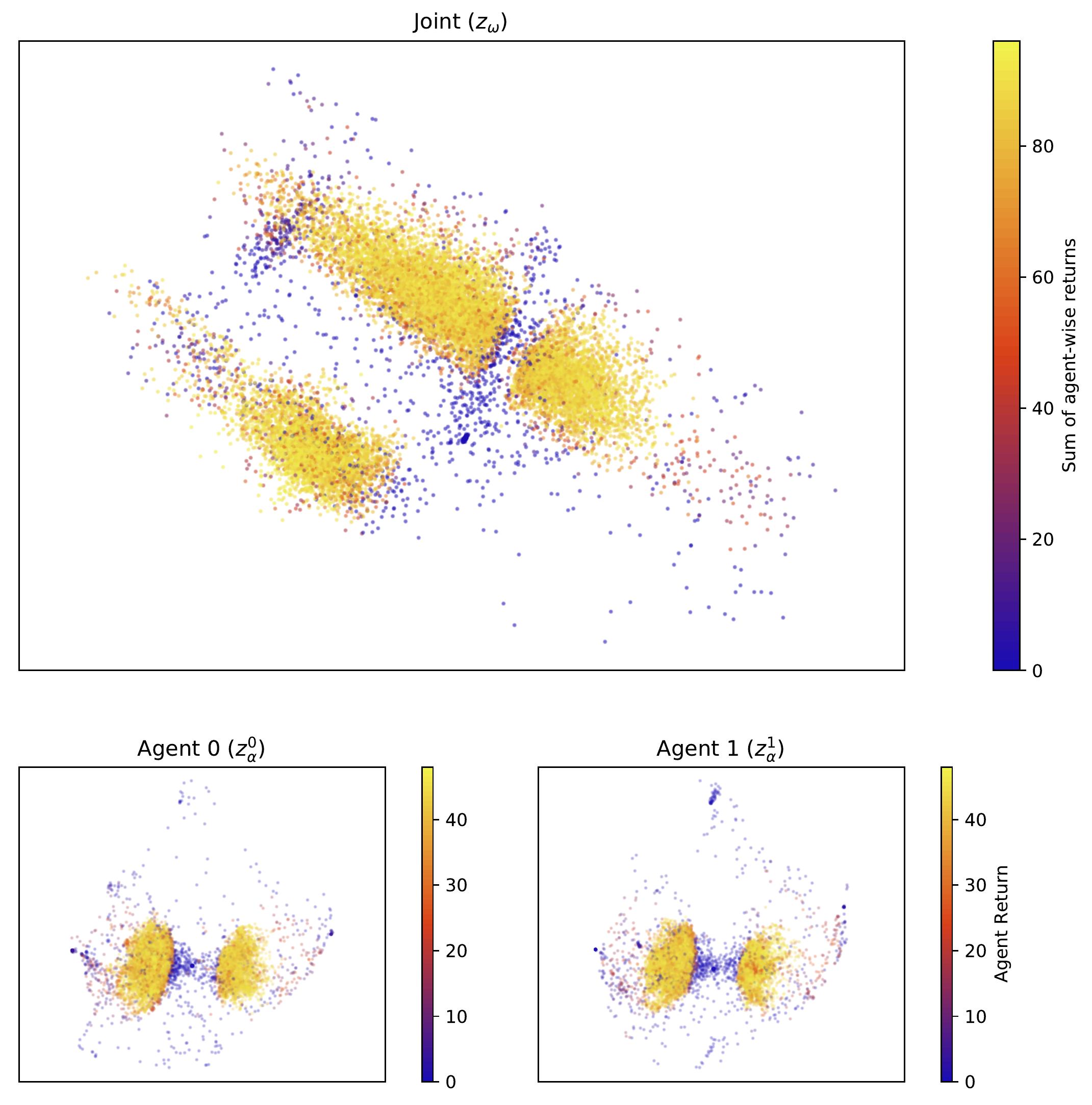}
        \caption{Latent spaces, labeled by trajectory return.}
        \label{fig:left_right_env_latent_rewards}
    \end{subfigure}
    \caption{
        Reward statistics for the 2-agent coordination environment.
        \subref{fig:left_right_env_latent_reward_statistics} visualizes the distribution of agent returns in the dataset.
        \subref{fig:left_right_env_latent_rewards} visualizes the agent behavior spaces, with trajectories labeled by the return attained.
    }
    \label{fig:left_right_env_reward_stats_all}
\end{figure}

We additionally provide a more detailed overview of trajectory returns in each of \cref{fig:multimodal_reward_env_reward_statistics,fig:left_right_env_latent_reward_statistics,fig:mujoco_halfcheetah_reward_statistics,fig:mujoco_antwalker_reward_statistics}.
These figures visualize the distribution of individual agent returns in each of the considered environments (respectively, hill-climbing, coordination game, 2-agent HalfCheetah, and 4-agent AntWalker).
Notably, many of the trajectories in the HalfCheetah domain are skewed towards medium and low-return behaviors, with very few high-return trajectories present (\cref{fig:mujoco_halfcheetah_reward_statistics}). 
Interestingly, in the AntWalker dataset we observe a bimodal return distribution (seemingly consisting of very low return and very high return trajectories, as seen in \cref{fig:mujoco_antwalker_reward_statistics}).
Closer inspection of this domain reveals that a large number of training runs result in highly sub-optimal behaviors with very low returns;
in \cref{fig:mujoco_antwalker_reward_statistics_thresh}, we visualize the distribution of returns with these low-return trajectories excluded, observing a similar distribution to the HalfCheetah dataset (consisting primarily of medium-return trajectories).
Similarly, a large proportion of trajectories with extremely low returns exist in the hill-climbing and coordination game domains (\cref{fig:multimodal_reward_env_reward_statistics,fig:left_right_env_latent_reward_statistics});
we verified that these low-return trajectories consist primarily of random agent behaviors collected early in training.
Overall, this analysis is helpful in determining that the analyzed datasets exhibit a wide range of trajectories (in terms of agent returns), covering random behaviors, sub-optimal behaviors, and high-return behaviors, similar to typical offline RL datasets~\citep{fu2020d4rl}. 

As part of this investigation, we also analyze the discovered behavior spaces (both at joint and local agent levels) with respect to the return distributions.
Specifically, \cref{fig:multimodal_reward_env_latent_rewards,fig:left_right_env_latent_rewards,fig:mujoco_halfcheetah_env_latent_rewards,fig:mujoco_antwalker_env_latent_rewards} visualize the behavior spaces learned by \OurAlgAcronym, which are then labeled by the ground truth return corresponding to each trajectory.
Interestingly, despite \OurAlgAcronym not using \emph{any} reward or return information in learning agent behavior spaces, clear clusters corresponding to low-return (random/early-training trajectories), medium-return trajectories, and high-return trajectories are automatically discovered in all of the considered domains.
For example, in the hill-climbing environment, many of the trajectories corresponding to low returns are clustered in the center of the joint latent space in \cref{fig:multimodal_reward_env_latent_rewards};
in the HalfCheetah environment, a prominent cluster of high-return trajectories is visible in the top-left and bottom-left local latent spaces of agents 0 and 1, respectively (\cref{fig:mujoco_antwalker_env_latent_rewards});
in the AntWalker environment, the lowest-return trajectories are clustered in distinctive regions both at the joint and local level, with a medium-return trajectory set also visible in the latent spaces (\cref{fig:mujoco_antwalker_env_latent_rewards}).
Overall, these results indicate the richness of the learned latent spaces in terms of not only raw trajectories but also their capacity to cluster trajectories that exhibit similar performance, even without observing the reward function.

\subsection{Visualizing $\zhi$ and $\zlo$}\label{appendix:visualizing_latents}
This section summarizes the procedure used to generate the behavior space figures associated with $\zhi$ and $\zlo$. The visualization procedure we use is as follows:
\be
    \item We train \OurAlgAcronym using the offline behavior datasets of interest.
    \item Following training of \OurAlgAcronym, we pass each trajectory $\tau$ through both the joint encoder $q_{\phi}(\zhi|\tau)$ and the agent-wise local encoders $q_{\phi}(\zlo^i|\tau)$, sampling a joint latent $\zhi$ and a set of local latents $\zlo^i$ for all $N$ agents accordingly. Thus, for $K$ such trajectories we obtain a set of latent parameters $\{(\zhi, \zlo^1,\cdots,\zlo^N)_k\}_{k=1}^{K}$. 
    \item As the latent parameters are high-dimensional in nature, we visualize their 2D projection (e.g., using Principal Component Analysis), thus yielding the behavior space visualizations such as those in \cref{fig:multimodal_reward_env_clusters_latent}.
\ee

\clearpage
\subsection{Model high-level code}\label{sec:model_code}
\Cref{code:main_alg} provides the high-level code of the \OurAlgAcronym model.
Note that the components of our model are standard network modules (e.g., MLPs, bidirectional LSTMs, etc.), which are indicated below and can be implemented using any desired neural network library.

\begin{lstlisting}[language=Python, caption=\OurAlgAcronym model high-level code., label={code:main_alg}]
"""Multiagent Offline Hierarchical Behavior Analyzer (MOHBA)."""

from typing import Dict, Tuple, Union
import distrax
from flax import linen as nn
import jax
import jax.numpy as jnp

Array = Union[np.ndarray, jnp.ndarray]


class MOHBA(nn.Module):
  config: Dict
  num_agents: int
  num_actions_per_agent: int

  def setup(self):
    """Setup model components."""
    # Joint models
    self.joint_prior = GaussianMixture(self.config.joint_prior)   
    self.joint_encoder = GaussianMixtureBidirLSTM(self.config.joint_encoder)

    # Local models (shared parameters across agents)
    self.local_prior = MLP(**self.config.local_prior)
    self.local_encoder = BidirLSTM(**self.config.local_encoder)
    self.local_policy = MLP(**self.config.policy)

  def __call__(self, rng: Array, states: Array, actions: Array):
    """Returns model outputs and KL divergences for loss computation.

    Args:
      rng: jax rng state.
      states: joint states, shape [B, T, N, S].
      actions: joint actions, shape [B, T, N, A].
    """
    B, T = states.shape[:2]
    agents_onehot = get_agents_onehot(
        front_dims=(B, T), num_agents=self.num_agents)

    # Append agent-onehots to tau.
    tau = jnp.concatenate((states, actions), axis=-1)
    tau_with_onehots = jnp.concatenate((tau, agents_onehot), axis=-1)

    ### Joint prior (z_omegas) ###
    dist_joint_prior = self.joint_prior(B)
    rng, z_omegas_rng = jax.random.split(rng)

    # self.joint_encoder internally reshapes tau to consume joint agent information
    dist_joint_encoder = self.joint_encoder(tau)
    z_omegas = dist_joint_encoder.sample(seed=z_omegas_rng)
    # Append agent one-hots to z_omegas
    z_omegas_with_onehots = utils.expand_and_concat_agent_ids(
        z_omegas, self.num_agents)  # [B, L] to [B, N, L+N]

    ### Local priors (z_alphas) ###
    # vmap over dim 1 of z_omegas_with_onehots (shape [B, N, L+N])
    dist_params = jax.vmap(
        self.local_prior.dist_params, in_axes=1, out_axes=1)(z_omegas_with_onehots)
    dist_local_prior = distrax.MultivariateNormalDiag(
        loc=dist_params[0], scale_diag=jnp.exp(dist_params[1]))

    rng, z_alphas_rng = jax.random.split(rng)
    dist_params = jax.vmap(
        self.local_encoder.dist_params, in_axes=2, out_axes=1)(tau_with_onehots)
    dist_local_encoder = distrax.MultivariateNormalDiag(
        loc=dist_params[0], scale_diag=jnp.exp(dist_params[1]))
    ### Agent policies ###
    z_alphas = dist_local_encoder.sample(seed=z_alphas_rng)
    z_alphas_policy = jnp.repeat(z_alphas[:, jnp.newaxis], T, axis=1)
    # Parameter-shared policies are run over each agent's own states
    pred_actions = jax.vmap(self.local_policy, in_axes=2, out_axes=2)(
        states, agents_onehot, z_alphas_policy)

    # KL divergence for joint latents.
    rng, kl_rng = jax.random.split(rng)
    kl_joint_latents = distrax.kl_divergence(
        distribution_a=dist_joint_encoder, distribution_b=dist_joint_prior)

    # KL divergence for local latents.
    rng, kl_rng = jax.random.split(rng)
    kl_local_latents = distrax.kl_divergence(
        distribution_a=dist_local_encoder, distribution_b=dist_local_prior)
    return pred_actions, kl_joint_latents, kl_local_latents
    
    
def get_agents_onehot(front_dims: Tuple[int, ...], num_agents: int):
  """Returns one-hot representation of agents for batched data.

  E.g., get_agents_onehot(front_dims=(2,3), num_agents=5) returns an
  array of shape [2,3,5,5], where the last two dims specify one-hots for each of
  the 5 agents, repeated over the first two `front_dims`.

  Args:
    front_dims: tuple specifying the sizes of the first K dims.
    num_agents: the number of agents.
  """
  agents_onehot = np.arange(num_agents)
  agents_onehot = jax.nn.one_hot(agents_onehot, num_agents, axis=-1)
  agents_onehot = jnp.broadcast_to(
    agents_onehot, front_dims+(num_agents, num_agents))
  return agents_onehot


def expand_and_concat_agent_ids(x: Array, num_agents: int):
  """Expand input x dims and concatenate agent ID one-hots to its features.

  Specifically, given input x of shape [B, features], adds agents dim,
  then concatenates agent ID one-hots to the features dim, resulting in output
  of shape [B, num_agents, features+num_agents].

  Args:
    x: input, shape [B, features].
    num_agents: the number of agents.

  Returns:
    x: output, shape [B, num_agents, features+num_agents].
  """
  # Create [B, num_agents, num_agents] agent IDs one-hot tensor
  agents_onehot = get_agents_onehot(
    front_dims=(x.shape[0],), num_agents=num_agents)

  # Expand x from [B, features] to [B, num_agents, features]
  x = jnp.expand_dims(x, axis=1)
  x = jnp.repeat(x, num_agents, axis=1)

  # Concat x and onehots
  return jnp.concatenate((x, agents_onehot), axis=-1)
\end{lstlisting}

\end{document}